\documentclass[11pt]{article}
\pdfoutput=1
\usepackage[
  margin=0.83in,
  headheight=12pt,
  headsep=25pt,
  includefoot,
  footskip=30pt,
]{geometry}
\usepackage{tikz}
\usepackage{graphicx}
\usepackage{comment}
\usepackage{listings}
\usepackage{inconsolata}
\usepackage{makecell}
\usepackage{tabularx} 
\usepackage{adjustbox}
\usepackage{rotating}
\usepackage{wrapfig}
\usepackage{caption} 

\definecolor{commentcolour}{rgb}{0.3,0.7,0.2}
\definecolor{backcolour}{rgb}{0.98,0.98,0.98}

\lstdefinelanguage{markdown}{
    comment=[l]{\#},
    morestring=[s]{```}{```},
    commentstyle=\color{commentcolour}\bfseries,
    stringstyle=\color{blue},
    basicstyle=\scriptsize\ttfamily,
    showstringspaces=false,
    breaklines=true,
    breakautoindent=false,
    breakindent=0pt,
    backgroundcolor=\color{backcolour},
}
\lstdefinestyle{mystyle}{
    morekeywords={self},
    basicstyle=\scriptsize\ttfamily,
    keywordstyle=\color{blue},
    commentstyle=\color{commentcolour}\bfseries,
    breaklines=true,
    breakautoindent=false,
    showstringspaces=false,
    backgroundcolor=\color{backcolour},
    stringstyle=\color{red},
}
\lstdefinelanguage{PythonPlus}[]{Python}{
  alsoother={@},
  morekeywords=[1]{,as,assert,nonlocal,with,yield,self,True,False,None} 
  morekeywords=[2]{,__init__,__add__,__mul__,__div__,__sub__,__call__,__getitem__,__setitem__,__eq__,__ne__,__nonzero__,__rmul__,__radd__,__repr__,__str__,__get__,__truediv__,__pow__,__name__,__future__,__all__,}, 
  morekeywords=[3]{,object,type,isinstance,copy,deepcopy,zip,enumerate,reversed,list,set,len,dict,tuple,range,xrange,append,execfile,real,imag,reduce,str,repr,}, 
  morekeywords=[4]{,Exception,NameError,IndexError,SyntaxError,TypeError,ValueError,OverflowError,ZeroDivisionError,}, 
  morekeywords=[5]{,ode,fsolve,sqrt,exp,sin,cos,arctan,arctan2,arccos,pi, array,norm,solve,dot,arange,isscalar,max,sum,flatten,shape,reshape,find,any,all,abs,plot,linspace,legend,quad,polyval,polyfit,hstack,concatenate,vstack,column_stack,empty,zeros,ones,rand,vander,grid,pcolor,eig,eigs,eigvals,svd,qr,tan,det,logspace,roll,min,mean,cumsum,cumprod,diff,vectorize,lstsq,cla,eye,xlabel,ylabel,squeeze,}, 
}

\usepackage{tikz}

\usepackage{amsmath} 
\usepackage{array} 
\usetikzlibrary{matrix, arrows}
\usepackage{amsmath,amssymb}
\usepackage{amsthm}
\usepackage{mathtools}
\usepackage{xspace}
\usepackage[noend]{algorithmic}
\usepackage[ruled,vlined]{algorithm2e}
\usepackage{url}
\usepackage{makeidx}
\usepackage{enumerate}
\usepackage{epstopdf}
\usepackage{booktabs}
\usepackage[table]{xcolor}

\usepackage[utf8]{inputenc}
\usepackage{thm-restate}
\usepackage{scalerel,stackengine}
\usepackage[shortlabels]{enumitem}
\usepackage{xr}
\usepackage{wrapfig}

\usepackage{fancyvrb}
\usepackage{xcolor}
\usepackage{bold-extra}
\usepackage{arydshln}

\usepackage{subcaption}
\usepackage[most]{tcolorbox}
\usepackage{fvextra}
\usepackage{float}
\usepackage{alltt}
\usepackage{soul}
\usepackage{MnSymbol,wasysym}
\usepackage{fancyvrb}
\usepackage{multirow}
\usepackage[final]{hyperref}
\usepackage[bottom]{footmisc}
\usepackage{diagbox}
\usepackage{mdframed}
\usepackage{todonotes}
\setuptodonotes{inline}
\usepackage{tikz}
\usetikzlibrary{shapes,calc,positioning}

\usepackage[most]{tcolorbox}
\definecolor{light-gray}{gray}{0.97}
\definecolor{light-gray1}{gray}{0.85}
\tcbset{
  aibox/.style={
    enhanced,
    breakable,
    width=\linewidth,             
    grow to left by=0pt,          
    colback=light-gray,
    colframe=light-gray1,
    colbacktitle=black,
    coltitle=white,
    fonttitle=\bfseries,
    boxrule=0.5pt,
    top=6pt, bottom=6pt, left=5pt, right=5pt,
    before skip=12pt, after skip=12pt,
    attach boxed title to top left={yshift=-0.07in,xshift=0pt}, 
    boxed title style={boxrule=0pt,colframe=white,},
    fontupper=\ttfamily\small,
  }
}
\newtcolorbox{AIbox}[2][]{aibox,title=#2,#1}

\definecolor{BestGreen}{HTML}{2ECC71}   
\definecolor{SecondAmber}{HTML}{F1C40F} 

\definecolor{modernblue}{HTML}{E6F0FA} 

\definecolor{teal}{HTML}{1ABC9C} 
\definecolor{brightblue}{HTML}{3498DB} 

\definecolor{orange}{HTML}{E67E22} 
\definecolor{softyellow}{HTML}{F39C12} 

\definecolor{aigold}{RGB}{244,210, 1} 
\definecolor{aigreen}{RGB}{210,244,211} 

\sethlcolor{aigreen}

\definecolor{aired}{RGB}{255,180,181}

\newtcbox{\mybox}[1][green]{on line,
arc=0pt,outer arc=0pt,colback=#1!10!white,colframe=#1!50!black,
boxsep=0pt,left=0pt,right=0pt,top=0pt,bottom=0pt,
boxrule=0pt,bottomrule=0pt,toprule=0pt}

\usepackage{todonotes}

\usepackage{array}
\usepackage{multirow}
\usepackage{pifont}
\usepackage{adjustbox}

\newcommand{\cmark}{\textcolor{green!70!black}{\ding{51}}} 
\newcommand{\xmark}{\textcolor{red}{\ding{55}}}            

\usepackage{acronym}
\newacro{llm}[LLM]{Large Language Model}
\newacro{nlp}[NLP]{Natural Language Processing}
\newacro{mt}[MT]{Machine Translation}
\newacro{nmt}[NMT]{Neural Machine Translation}
\newacro{sft}[SFT]{Supervised Fine-Tuning}
\newacro{lora}[LoRA]{Low Rank Approximation}

\newcommand{\BYOL}{\texttt{BYOL}\xspace}
\newcommand{\BYOLC}{\texttt{BYOL-nya}\xspace}
\newcommand{\BYOLM}{\texttt{BYOL-mri}\xspace}
\newcommand{\GEMMA}{{Gemma-3}\xspace}
\newcommand{\QWEN}{{Qwen-3}\xspace}
\newcommand{\APERTUS}{{Apertus}\xspace}
\newcommand{\OSS}{{GPT-OSS}\xspace}
\newcommand{\SMOLTALK}{{SmolTalk2}\xspace}
\newcommand{\FINEWEB}{{FineWeb2}\xspace}
\newcommand{\FineWebEdu}{{FineWeb-Edu}\xspace}
\newcommand{\AYA}{{Aya}\xspace}
\newcommand{\SMOL}{{Smol}\xspace}
\newcommand{\GLOBALMMLU}{{Global MMLU-Lite}\xspace}

\begin{document}
\title{BYOL: Bring Your Own Language Into LLMs}
\author{
Syed Waqas Zamir\textsuperscript{1*}, 
Wassim Hamidouche\textsuperscript{1*},  
Boulbaba Ben Amor\textsuperscript{2}, \\
Luana Marotti\textsuperscript{1},
Inbal Becker-Reshef\textsuperscript{1},
Juan Lavista Ferres\textsuperscript{1} \\ \vspace{0.2em}
\textsuperscript{1}Microsoft AI for Good Research Lab \quad
\textsuperscript{2}Inception, G42\\
\textsuperscript{*}\footnotesize Equal contribution
}

\date{}

\maketitle

\vspace{-2em}
\begin{abstract}
Large Language Models (LLMs) exhibit strong multilingual capabilities, yet remain fundamentally constrained by the severe imbalance in global language resources.
While over 7{,}000 languages are spoken worldwide, only a small subset ($<$100) has sufficient digital presence to meaningfully influence modern LLM training.
This disparity leads to systematic underperformance, cultural misalignment, and diminished accessibility for speakers of low-resource and extreme-low-resource languages.
To address this gap, we introduce \textbf{B}ring \textbf{Y}our \textbf{O}wn \textbf{L}anguage (BYOL), a unified framework that enables scalable, language-aware LLM development tailored to each language's digital footprint. BYOL begins with a language resource classification—mapping languages into four tiers (Extreme-Low, Low, Mid, High) based on curated web-scale corpora, and uses this classification to determine the appropriate integration strategy. For low-resource languages, we propose a full-stack data refinement and expansion pipeline, combining corpus cleaning, synthetic text generation, continual pretraining, and supervised finetuning. Applied to \texttt{Chichewa} and \texttt{Māori}, this pipeline yields two language-specific LLMs that achieve $\sim$12\% average improvement over strong multilingual baselines across 12 benchmarks, while preserving English and multilingual capabilities via weight-space model merging.  For extreme-low-resource languages, we introduce a translation-mediated inclusion pathway, demonstrating with \texttt{Inuktitut} that a tailored MT system can deliver +4 BLEU improvement over a commercial baseline, enabling high-accuracy LLM access in settings where direct modeling is otherwise infeasible. 
Our results show that BYOL offers a practical, extensible, and data-efficient recipe for expanding LLM capabilities to the long tail of the world's languages.
Finally, we release human-translated versions of the \GLOBALMMLU benchmark in \texttt{Chichewa}, \texttt{Māori}, and \texttt{Inuktitut}, and make our codebase and models publicly available at \url{https://github.com/microsoft/byol}.
\end{abstract}

\vspace{-1em}

\section{Introduction}
LLMs have achieved remarkable gains across natural language processing tasks, driven by large-scale pretraining on multilingual web corpora \cite{gao2020pile,weber2024redpajama,abadji-etal-2022-towards,cerebras2023slimpajama,penedo2024refinedweb,soldaini-etal-2024-dolma,penedo2024fineweb}. However, they are strongly affected by the uneven distribution of digital text across languages~\cite{joshi-etal-2020-state,penedo2025fineweb2}. While over 7{,}000 languages are spoken worldwide\footnote{https://www.ethnologue.com/}, only a small fraction dominates the web (e.g., $\sim$90\% of Common Crawl text comes from just twenty languages\footnote{https://commoncrawl.github.io/cc-crawl-statistics/plots/languages.html}). As generative AI increasingly becomes a general-purpose technology, this imbalance makes access to its benefits language-contingent, thereby reinforcing a systematic divide between {high-resource} languages (with abundant digital text) and {low-resource} languages (with minimal digital presence)~\cite{cohere2024languagegap,peppin2025multilingual}. 

\begin{figure}[t]
  \begin{center}
    \begin{tabular}{cc}\hspace{-4mm}
    \scalebox{0.9}{
    \begin{picture}(250,180)
    \put(0,0){\includegraphics[width=0.49\linewidth]{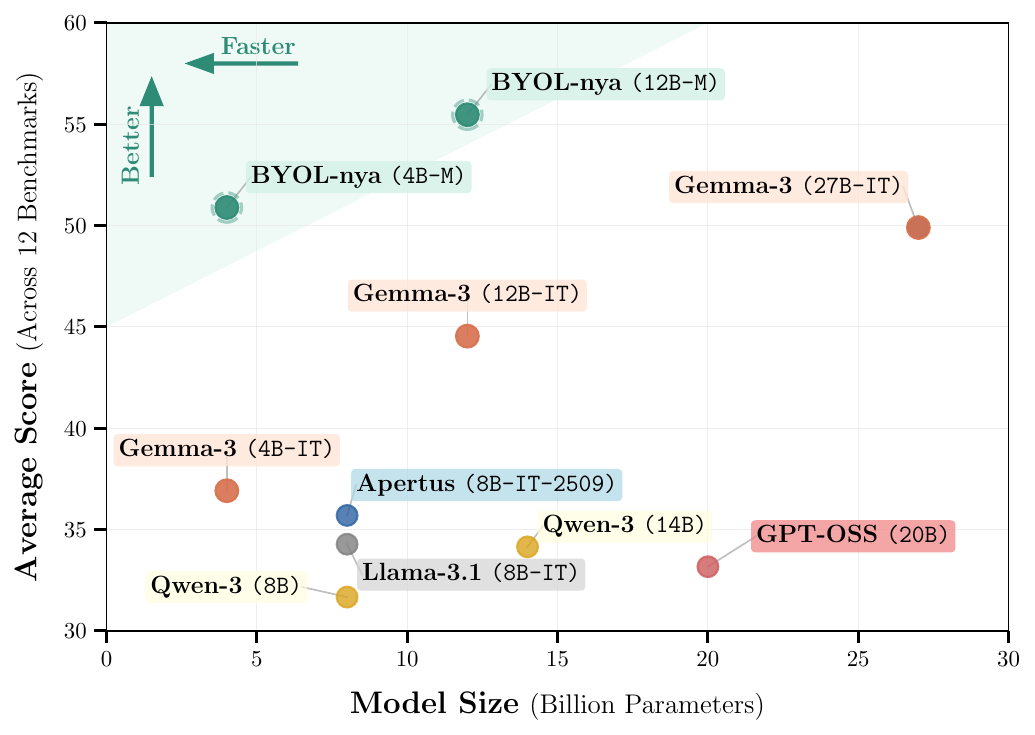}}
    \put(45,26){\tiny\cite{yang2025qwen3}}
    \put(35,57){\tiny\cite{team2025gemma}}
    \put(89,93){\tiny\cite{team2025gemma}}
    \put(170,119){\tiny\cite{team2025gemma}}
    \put(92,49){\tiny\cite{hernandez2025apertus}}
    \put(92,28){\tiny\cite{grattafiori2024llama}}
    \put(139,40){\tiny\cite{yang2025qwen3}}
    \put(186,37){\tiny\cite{openai2025gptoss120bgptoss20bmodel}}
    \end{picture}
    }

    & \hspace{-6mm}
    \scalebox{0.9}{
    \begin{picture}(250,180)
    \put(0,0){\includegraphics[width=0.49\linewidth]{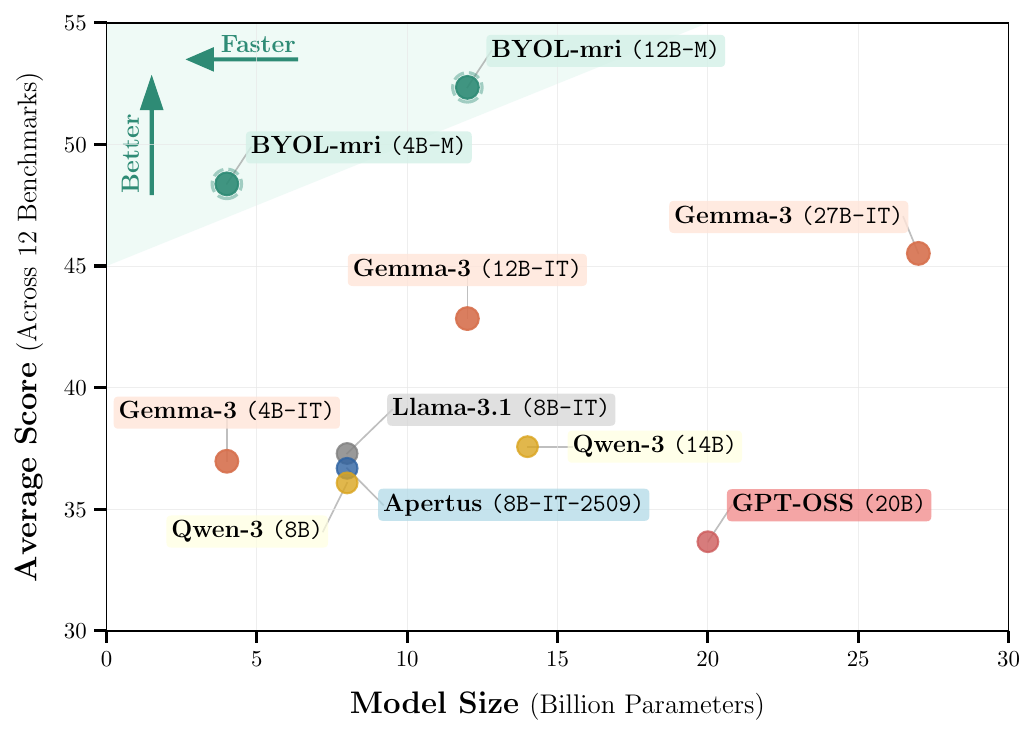}}
    \put(47,39){\tiny\cite{yang2025qwen3}}
    \put(35,66){\tiny\cite{team2025gemma}}
    \put(90,99){\tiny\cite{team2025gemma}}
    \put(170,112){\tiny\cite{team2025gemma}}
    \put(97,45){\tiny\cite{hernandez2025apertus}}
    \put(99,67){\tiny\cite{grattafiori2024llama}}
    \put(152,59){\tiny\cite{yang2025qwen3}}
    \put(186,45){\tiny\cite{openai2025gptoss120bgptoss20bmodel}}
    \end{picture}
    }

    \end{tabular}
  \end{center}\vspace{-2em}
\caption{\small \underline{\textbf{Performance comparison of LLMs}} on \texttt{Chichewa} (left) and \texttt{Māori} (right). On both languages, our \BYOL models deliver strong results; notably, the 4B variants outperform the $\sim\!7\times$ larger \GEMMA~(27B-IT).}
\vspace{-0.8em}
\label{fig:teaser-byol-chatmodels}
\end{figure}
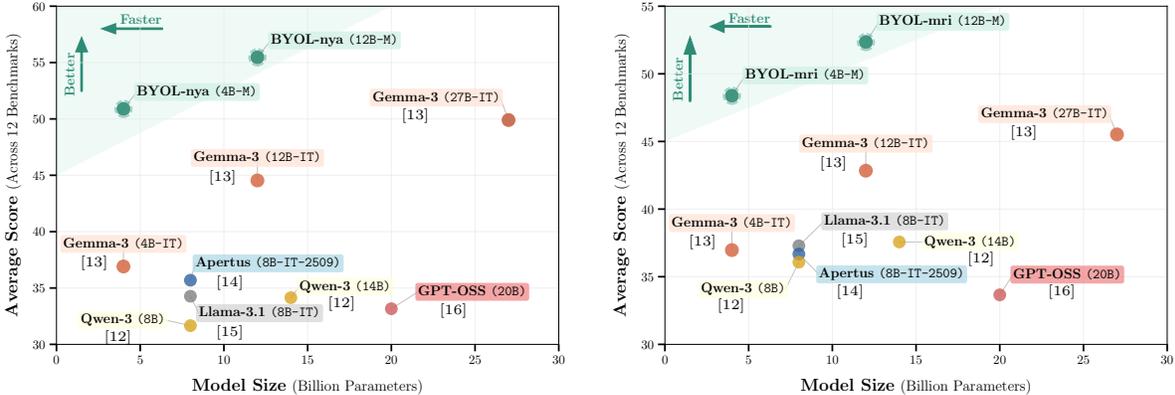

Crucially, this gap is not merely a quality issue. In countries where most people operate in local languages, a primarily English-centric LLM ecosystem effectively gates the benefits of AI behind a linguistic barrier. Malawi is a concrete illustration: although English is an official language, everyday communication is dominated by local languages such as Chichewa; fewer than 4\% of the population speaks English, and current LLM support for Chichewa remains limited. Without practical, repeatable ways to bring such languages into LLMs, the resulting imbalance becomes a widening AI diffusion divide~\cite{misra2025measuringaidiffusionpopulationnormalized,misra2025aidiffusionlowresource}, with downstream implications for access to AI-enabled services in education, health, legal systems, and economic productivity~\cite{appel2025anthropic_index_report,openai2025gdpval}.

The language divide also has broader technical consequences. LLMs trained primarily on English and a small set of high-resource languages show degraded performance on underrepresented languages (Fig.~\ref{fig:teaser-byol-chatmodels}), limited grounding in local context, and amplified cultural and epistemic biases~\cite{cohere2024languagegap,singh2024global}.
The lack of clean, sufficiently sized corpora constrains multilingual generalization~\cite{holtermann-etal-2024-evaluating,maini2025beyondweb,xuan-etal-2025-mmlu,ahuja2024megaverse}. Even when some data exists, expanding models to include more languages introduces the {curse of multilinguality}~\cite{chang-etal-2024-multilinguality,arivazhagan2019massivelymultilingual,conneau-etal-2020-unsupervised,pfeiffer2022liftingcurse}, where as language coverage increases, overall performance declines. Despite claims of broad multilingual support, most LLMs are evaluated primarily on high-resource languages, leaving the long tail of Low-Resource Languages (LRLs) untested~\cite{ustun-etal-2024-aya,adelani-etal-2025-irokobench,singh2024global,ahuja2024megaverse,hernandez2025apertus}. 

Beyond accuracy, speakers of underrepresented languages also face practical disadvantages, including higher inference costs and latency due to inefficient tokenization~\cite{ahia2023tokenizationcost,arnett2025language,arnett2025tokenizerfree,abagyan2025one,hernandez2025apertus}, and exclusion from safety-critical applications~\cite{peppin2025multilingual} as models insufficiently tuned for a language may distort meaning and produce harmful content~\cite{deng2024multilingualjailbreakchallenges,yong2023low,peppin2025multilingual}. Communities in LRL speaking countries~\cite{misra2025aidiffusionlowresource} also face structural barriers, i.e., limited access to compute, data, and research ecosystems, that further widen the technological gap~\cite{oecd2023ailanguagemodels,nekoto-etal-2020-participatory}. 
Addressing this challenge requires far more than scaling existing multilingual models. We argue that the solution must be a language-centric, efficient model development approach that adapts to each language’s digital footprint, resource quality, and practical constraints.

In this work, we introduce a unified framework, Bring Your Own Language (BYOL), designed to systematically enable LLM capabilities for low-resource and extreme-low-resource languages. 
\textbf{First}, we propose a \textbf{language resource classification} framework that maps each language to one of four tiers (Extreme-Low, Low, Mid, High) based on its effective digital footprint in curated web-scale corpora. This classification guides integration strategies: direct \emph{finetuning} for mid/high-resource languages, additional \emph{continual pretraining} for low-resource languages, and \emph{translation-based inclusion} for extreme-low-resource cases.  \textbf{Second}, for languages with limited noisy but usable corpora (low-resource tier), we develop a \textbf{data refinement and expansion pipeline} that cleans, augments, and enhances native-language text to support continual pretraining and downstream finetuning. We demonstrate this pipeline through two full-stack case studies\footnote{Chichewa (ISO 639-3: \texttt{nya}) is a low-resource Bantu language of Malawi, and Māori (ISO 639-3: \texttt{mri}) represents a revitalized Indigenous language of New Zealand. Chichewa and Māori were selected as representative low-resource languages from distinct linguistic families, allowing evaluation across typologically diverse, underserved languages.}: a \texttt{Chichewa LLM} (named \BYOLC) and a \texttt{Māori LLM} (\BYOLM),  each achieving roughly 12\% average improvement over strong multilingual baselines across 12 benchmarks (Fig.~\ref{fig:teaser-byol-chatmodels}).
\textbf{Third}, for languages with negligible digital presence (\textit{extreme-low-resource} tier), we introduce a \textbf{translation-mediated inclusion pathway} that enables access to LLM capabilities via high-quality forward- and back-translation. Using \texttt{Inuktitut}\footnote{Inuktitut (ISO 639-3 code: \texttt{iku}) is an Indigenous Inuit language spoken in Inuit Nunangat, Canada.} as a case study, we train a translation system achieving a $\sim$4 BLEU improvement over a commercial baseline and show that translation-mediated LLM use yields a $\sim$14\% accuracy gain over direct inference.
\textbf{Finally}, to support open, comparable evaluation for future research, we release \textit{human-translated versions} of \GLOBALMMLU~\cite{singh2024global} in \texttt{Chichewa}, \texttt{Māori}, and \texttt{Inuktitut} languages.
Our overarching goal is to demonstrate a scalable and extensible recipe for supporting the world's LRLs that, in contrast to generic multilingual scaling, shows the promise of language-aware, resource-adaptive LLM development for \textit{all} languages.

\begin{figure*}[t]
    \centering
    \includegraphics[width=0.95\textwidth]{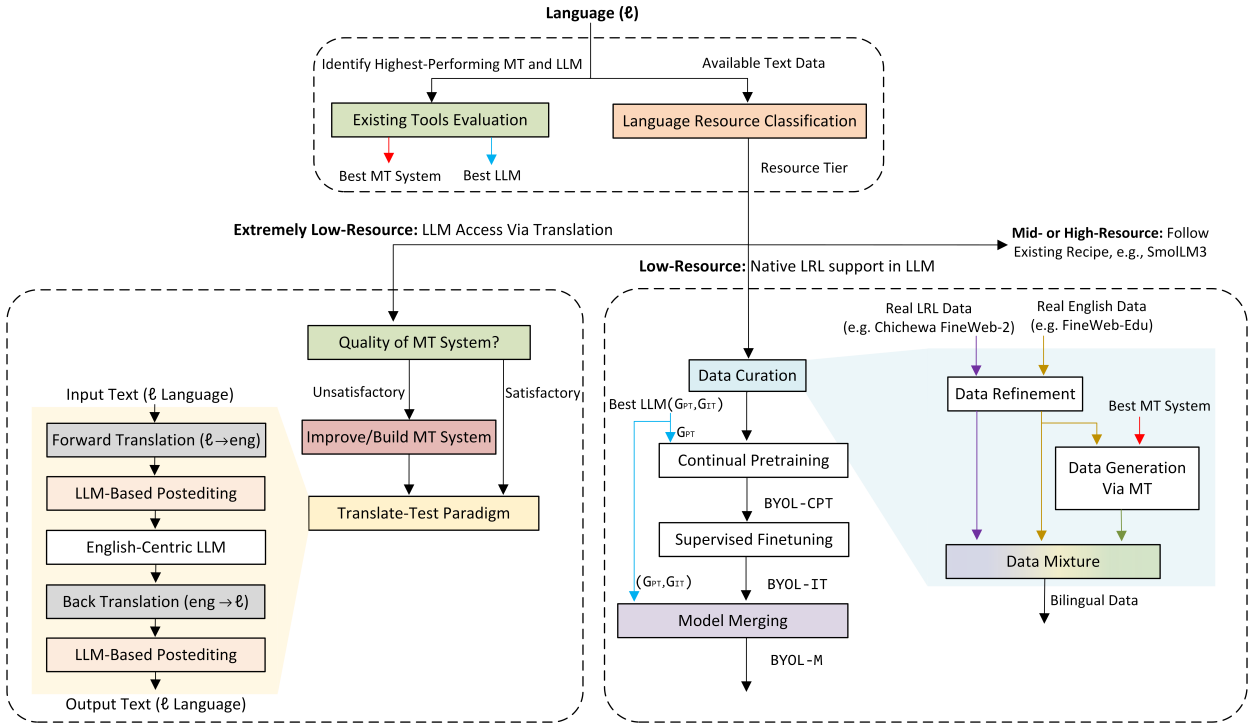}
    \caption{
        {\small \bf \underline{Overview of the BYOL pipeline.}} The system classifies a target language~$\ell$ by resource tier and selects the appropriate adaptation pathway. $G_{\mathrm{PT}}$ and $G_{\mathrm{IT}}$ denote the base and instruction-tuned variants of the generalist LLM selected through the initial tool evaluation.
    }
    \label{fig:pipeline}
\end{figure*}

\section{Bring Your Own Language (\BYOL) Framework}

\noindent\textbf{Overall pipeline.} Given a target language $\ell$, our pipeline (Fig.~\ref{fig:pipeline}) begins by evaluating existing tools to identify the best-performing LLM and machine translation (MT) systems. Concurrently, a language resource classification module analyzes the amount of text data available for language $\ell$ and assigns it to one of four tiers: {extreme-low-resource}, {low-resource}, {mid-resource}, or {high-resource}. This classification determines the route for language adaptation.
For \textbf{extreme-low-resource languages}, where textual data is insufficient for direct adaptation, access to LLM capabilities is enabled through a translation interface that follows the Translate-Test paradigm~\cite{artetxe2023revisitingmachinetranslationcrosslingual}. 
For \textbf{low-resource languages}, which have limited but usable data, the framework employs a data-centric strategy to enable native LLM support.
Languages classified as \textbf{mid- or high-resource}, which are typically well represented in multilingual models~\cite{abdin2024phi,openai2025gptoss120bgptoss20bmodel,team2025gemma,yang2025qwen3,liu2024deepseek,grattafiori2024llama}, fall outside the scope of this work. 

\subsection{Initial Assessment}
The assessment stage has two key modules: (1) {Language Resource Classification}, which assigns~$\ell$ to a resource tier, and (2) {Existing Tools Evaluation}, which benchmarks the performance of existing MT and LLM systems for $\ell$.

\begin{figure}[t]
  \centering
  \includegraphics[width=0.92\linewidth]{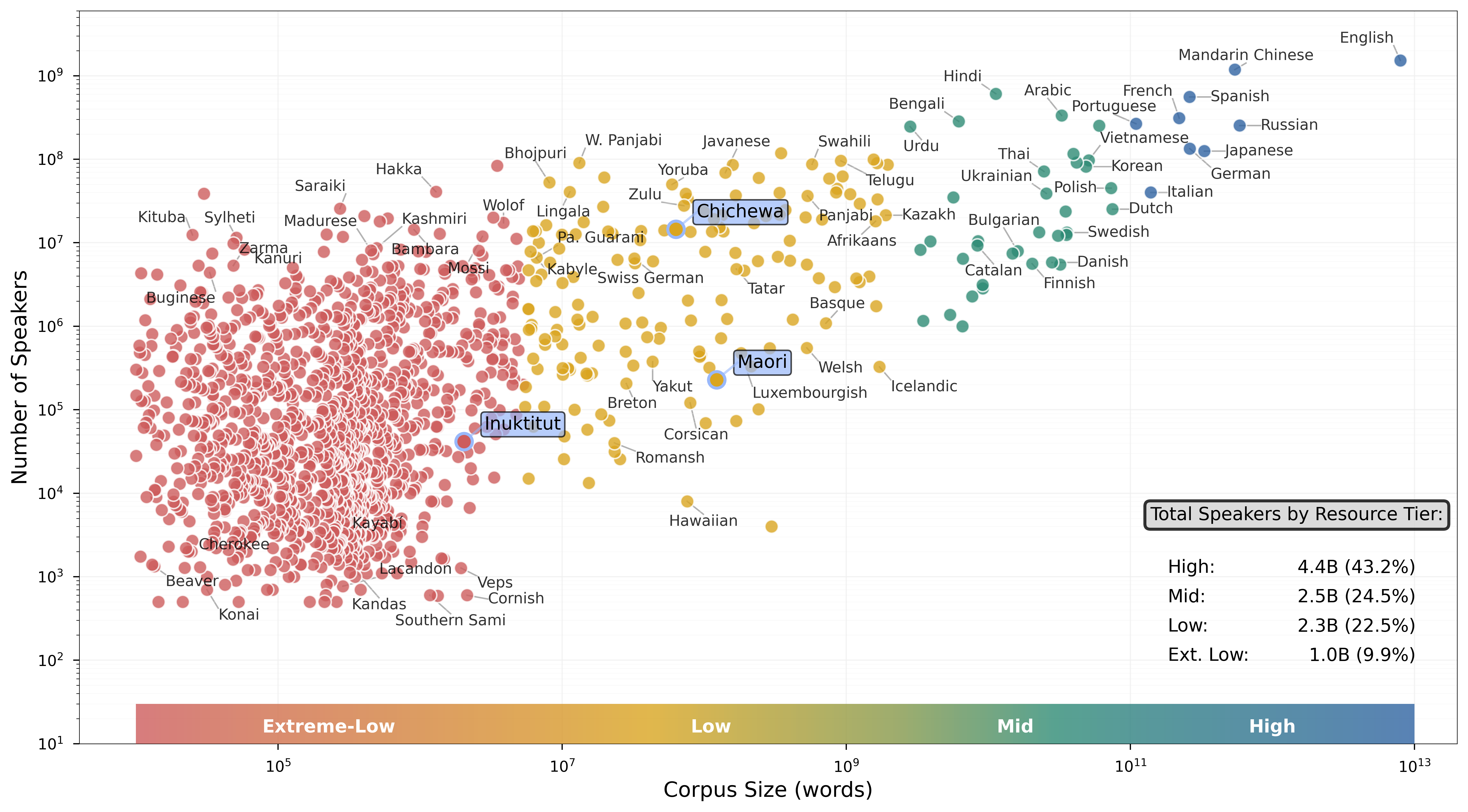} \vspace{-1em}
\caption{\small \underline{\textbf{Language resource classification}} derived from 
\FINEWEB~\cite{penedo2025fineweb2}. The continuous color bar shows the full spectrum of 
digital text availability, while the four discrete resource tiers ({Extreme-Low}, 
{Low}, {Mid}, {High}) use interpretable corpus-size boundaries to route 
languages through the appropriate pathway in the \BYOL framework.}
  \label{fig:language-classification}
\end{figure}

\subsubsection{Language Resource Classification} 
The taxonomy of Joshi \textit{et al.}~\cite{joshi-etal-2020-state} categorizes languages into six resource levels and remains a foundational reference for language classification. However, it predates the modern LLM era, and the web-scale data available for many languages has changed substantially since its publication. We therefore revisit language resource classification using \FINEWEB~\cite{penedo2025fineweb2}, a curated multilingual corpus derived from and deduplicated across Common Crawl~\cite{commoncrawl}. \FINEWEB provides a consistent estimate of the effective digital footprint for more than 1{,}000 written languages. For each language, we compute the total word count as a proxy for corpus size and pair it with speaker population to provide a two-dimensional view of digital representation (Fig.~\ref{fig:language-classification}).

{We group languages into four corpus-size tiers using interpretable boundaries that reflect the distribution in Fig.~\ref{fig:language-classification} and align with practical distinctions observed in multilingual LLM pretraining~\cite{kudugunta2023madlad,penedo2025fineweb2}.
}
\begin{itemize}
    \item \textbf{Extreme-Low-Resource} ($\leq 5\times10^{6}$ words):
    Languages with negligible digital presence and minimal-to-no LLM exposure.
    For these languages, native integration into LLMs is currently infeasible, and MT-based access is the most practical route.
    \item \textbf{Low-Resource} ($5\times10^{6}$ – $2\times10^{9}$ words): Languages with limited but usable textual data, making them candidates for native inclusion in LLMs through targeted continual pretraining. 
    \item \textbf{Mid-Resource} ($2\times10^{9}$ – $10^{11}$ words): Languages with substantial textual resources and moderate-to-strong LLM coverage, for which light adaptation (e.g., domain-specific finetuning) can typically close most of the remaining task-specific performance gap.
    \item \textbf{High-Resource} ($> 10^{11}$ words): Languages with abundant, high-quality web-scale corpora that enjoy comprehensive LLM support across diverse tasks.
\end{itemize}

These tier boundaries are indicative rather than absolute, since language resources evolve continuously as new text becomes available. In practice, the effort required for a language to approach reference English performance depends not only on corpus size, but also on linguistic characteristics, writing system, and the extent to which it can benefit from cross-lingual transfer. The continuous color bar in Fig.~\ref{fig:language-classification} reflects this spectrum of digital scarcity, while the four tiers introduced in this paper provide a simple, actionable routing scheme that guides the integration choices of the \BYOL framework. Figure~\ref{fig:language-classification} also shows that speaker population and corpus size are not strictly correlated: some languages with relatively small populations (e.g., Icelandic) have strong web presence, whereas others with millions of speakers (e.g., Saraiki, Kituba) remain digitally underrepresented.

\subsubsection{Existing Tools Evaluation: LLMs and MT Systems}
\label{sec:tool-assessment}
The second module identifies which LLMs and MT systems currently support language~$\ell$ and measures their performance.
This evaluation serves three purposes.
(1) For low-resource languages, MT systems can be used to generate synthetic training data by translating high-quality corpora from pivot languages (e.g., English or Spanish) into~$\ell$, supplementing limited real text.
(2) It identifies the highest-performing LLM with \emph{any} measurable representation of language~$\ell$, providing a baseline model for subsequent adaptation.
(3) For extreme-low-resource languages lacking direct LLM support, MT systems serve as a fallback interface, enabling indirect access to LLM capabilities via translation.

A major challenge is the lack of benchmarking datasets, which affects thousands of languages, as existing resources (e.g., FLORES-200~\cite{nllb-24}) cover only a small subset.
Creating new parallel benchmarks for each language is prohibitively costly and time-intensive.
We therefore adopt a scalable, reference-free evaluation framework.

\vspace{0.4em}\textbf{Round-trip translation as proxy evaluation.}
We employ the round-trip translation (RTT) strategy~\cite{zhou2023rethinking}. A sentence from a pivot language (e.g., English\footnote{While English is used as an example pivot, any high-resource language that is typologically and culturally closer to the target language can be substituted.}) is translated into~$\ell$ and then back into the pivot language. The reconstructed sentence is compared with the original source to measure how well meaning is preserved through the round-trip translation cycle.
RTT is domain-agnostic in its original form. However, translation quality varies significantly across domains~\cite{koretaka-etal-2023-mitigating,saunders-deneefe-2024-domain,shen2020sourcetargetdomainmismatchproblem,saunders2022domain}.
To address this, we extend RTT with a domain-conditioned evaluation that enables fine-grained analysis of how translation models generalize across different domains.
Overall, the process is defined as:

\begin{equation}
\text{RTTScore} = 
\frac{1}{\left| D \right|} 
\sum_{d \in D} 
\frac{1}{N_d} 
\sum_{i=1}^{N_d} 
\mathcal{M}\!\left(
s_i^{(d)},\, 
\mathcal{T}_{\ell\rightarrow\text{eng}}\!\left(
\mathcal{T}_{\text{eng}\rightarrow\ell}\!\left(
s_i^{(d)}
\right)\right)
\right),
\label{eq:rttscore}
\end{equation}
where $D$ is the set of domains, 
$s_i^{(d)}$ is the $i$-th sentence sampled from domain $d$, 
$N_d$ is the number of sentences in that domain, 
$\mathcal{T}_{(\cdot)}$ denotes an MT engine or an LLM, and $\mathcal{M}$ represents a fidelity metric such as SacreBLEU~\cite{post-2018-call}, chrF++~\cite{popovic-2017-chrf}, or embedding-based cosine similarity.
To operationalize this framework, we introduce a dedicated benchmark for cross-domain RTT evaluation.

\vspace{0.4em}\noindent\textbf{RTTBench-Mono: domain-balanced monolingual dataset.}
We develop RTTBench-Mono, a 1,250-sentence English dataset across 25 domains from NVIDIA's taxonomy\footnote{https://huggingface.co/nvidia/domain-classifier} after excluding the adult category. For each domain, 50 sentences are generated using \texttt{Azure OpenAI GPT-4.1} with varying lengths and syntactic complexity.
The dataset provides balanced coverage across diverse topics and serves as a standardized source for domain-conditioned RTT evaluation. 
We validate dataset quality using classifier-based and embedding-based checks; details on validation methodology, prompt design, and the procedure to prevent domain drift are provided in Appendix~\ref{app:rttbench}.

\subsection{Low-Resource Pathway: Native Language Support in LLMs}
\label{sec:lrl}
Training an LLM from scratch for a low-resource language is infeasible due to limited text. We therefore begin with a multilingual \emph{generalist} LLM that shows preliminary knowledge of~$\ell$, identified through the tool-assessment procedure in Sec.~\ref{sec:tool-assessment}.
We transform this model into a \emph{language-specific expert} through three stages: continual pretraining, supervised finetuning, and model merging.

\vspace{0.4em}\noindent\textbf{Continual pretraining (CPT).}
\FINEWEB~\cite{penedo2025fineweb2} is the primary source of multilingual non-English text for pretraining, but its coverage varies substantially across LRLs (Fig.~\ref{fig:language-classification}).
To expand the corpus, we translate the English \FineWebEdu dataset~\cite{penedo2024fineweb} into the target~$\ell$ using the best-performing MT system identified in Sec.~\ref{sec:tool-assessment}.
This resulting synthetic text is mixed with the real LRL data. 
We also include a random subset of English \FineWebEdu to preserve the baseline LLM's English competence and cross-lingual capabilities in the final model.
Although \FINEWEB applies rule-based filtering, its text still contains formatting errors, noise and limited coherence. 
We therefore refine all the pretraining data through guided rephrasing using a large multilingual LLM. Each text sample is treated as an initial draft and rewritten to improve clarity and structure, while expanding on-topic content. Additionally, toxic and harmful material is removed.
The final CPT corpus consists of:
(1) refined real LRL text from \FINEWEB,
(2) synthetic LRL data translated from the refined \FineWebEdu corpus, and
(3) refined real English data from \FineWebEdu~\cite{penedo2024fineweb}.
Further details on CPT dataset construction are provided in~\ref{annex:cpt datasets}, and the data-refinement prompt is given in~\ref{app:data-refinement-prompt}.
We perform CPT separately for
each target language using its corresponding bilingual data mixture. This stage strengthens the model’s internal representation of $\ell$ and mitigates language drift, preventing the model from unintentionally switching to other languages mid-sentence during text generation.

\vspace{0.4em}\noindent\textbf{Supervised finetuning (SFT).}
We assemble a bilingual instruction dataset for SFT. When available for the target language, we source native-language~$\ell$ samples from the \AYA dataset~\cite{singh-etal-2024-aya} and \SMOL~\cite{caswell2025smol}. Because most LRLs lack sufficient instruction data, we translate instruction-response pairs from five high-resource \AYA languages, as well as a subset of \SMOLTALK~\cite{bakouch2025smollm3}, into~$\ell$. We also include a  portion of English instruction data from \SMOLTALK to maintain cross-lingual alignment. Dataset composition is summarized in Appendix~\ref{annex:SFT_Data}.
Each CPT model is finetuned on this data mixture to enhance its instruction-following and response-generation capabilities in the target language~$\ell$.

\vspace{0.4em}\noindent\textbf{Model merging.}
Model merging combines a multilingual generalist model with a language-specific expert directly in weight space. This yields a unified network that gains native proficiency in~$\ell$ while retaining the multilingual behavior of the generalist model.
Let $G_{\mathrm{PT}}$ and $G_{\mathrm{IT}}$ denote the pretrained and instruction-tuned variants of the multilingual generalist model $G$.  
Let $E_{\ell}$ represent the language expert for ~$\ell$, obtained by applying CPT followed by SFT to $G_{\mathrm{PT}}$.  
Because all models originate from the same initialization, their parameter spaces remain aligned, allowing linear combination in weight space.
We define the merged model as:
\begin{equation}
M(\alpha,\beta)
=
G_{\mathrm{PT}}
+ \alpha \,(G_{\mathrm{IT}} - G_{\mathrm{PT}})
+ \beta \,(E_{\ell} - G_{\mathrm{PT}}),
\label{eq:model_merging}
\end{equation}

where $\alpha$ and $\beta$ are positive scaling coefficients. The first term $(G_{\mathrm{IT}} - G_{\mathrm{PT}})$ transfers the instruction-following behavior of the generalist model $G$, while the second term $(E_{\ell} - G_{\mathrm{PT}})$ injects the language-specific knowledge. This approach brings low-resource language expertise into the baseline LLM while preserving its multilingual and safety behaviors, without requiring additional training or alignment steps~\cite{huang-etal-2024-chat}.

\subsection{Extreme-Low-Resource Pathway: LLM Access via Machine Translation}
\label{sec:elrl}
For extreme-low-resource languages, available text is insufficient for direct model adaptation (Fig.~\ref{fig:language-classification}).
LLMs trained on such limited data produce unintelligible outputs if used directly.
Therefore, we adopt the \emph{Translate-Test} paradigm~\cite{artetxe2023revisitingmachinetranslationcrosslingual}, where input text is translated into English, processed by an English-centric LLM, and translated back into the source language.
This enables access to advanced LLM capabilities even for languages with minimal digital presence.
The effectiveness of the Translate-Test approach depends on MT quality: a usable MT system must exist for the target language, and its translations should be sufficiently accurate, since any errors or information loss propagate to the English LLM and degrade final responses. As a case study, we develop an MT system for \textit{Inuktitut}, and integrate it with an LLM.  

\vspace{0.4em}\noindent\textbf{Sentence alignment in extreme-low-resource languages.}
Sentence alignment identifies matching sentence pairs across bilingual documents and is critical for MT, as translation models are highly sensitive to noise and misaligned training examples.
Existing approaches rely on length-based heuristics~\cite{gale-church-1993}, lexical overlap~\cite{moore-2002-fast}, or pretrained embedding models~\cite{thompson-koehn-2019-vecalign,feng-etal-2022-language,heffernan-etal-2022-bitext} that are typically unavailable or unreliable for extreme-low-resource languages. Therefore, we employ a multilingual LLM to perform sentence alignment directly. Even with weak understanding of a given language, an LLM can still leverage structural and contextual cues to detect cross-lingual correspondences. In our pipeline, we use \texttt{Azure OpenAI GPT-5-chat} to extract aligned pairs from bilingual sources such as news articles and children’s books. The full alignment prompt is provided in Appendix~\ref{app:text-alignment-prompt}.

\vspace{0.4em}\noindent\textbf{Synthetic data generation via back-translation.}
High-quality parallel data is essential for MT training~\cite{costa2022no,kudugunta2023madlad}. However, extreme-low-resource languages typically have limited and domain-constrained bitext. We therefore use back-translation~\cite{sennrich-etal-2016-improving,edunov-etal-2018-understanding} to expand the available training data. We first train an initial \emph{$\ell$-to-English} model on the existing bitext and use it to translate monolingual $\ell$ sentences into English, producing synthetic English paired with human-written $\ell$ text. The synthetic and real pairs are mixed at a 1:1 ratio~\cite{haque-etal-2020-terminology} to balance quality and diversity.

Since the Translate-Test paradigm requires MT systems in both directions ({English-to-$\ell$} and {$\ell$-to-English}), we construct distinct monolingual corpora for each direction. For the {English-to-$\ell$} model, we rely on monolingual text in $\ell$, whose availability varies across languages~\cite{penedo2025fineweb2}. For the {$\ell$-to-English} model, we assemble an English corpus by sampling from diverse datasets (e.g., WikiMatrix~\cite{schwenk-etal-2021-wikimatrix}, News Commentary~\cite{tiedemann-2012-parallel}, Global Voices~\cite{nguyen-daume-iii-2019-global}); the complete list of datasets is in \ref{ref:mt-datasets} (Table~\ref{tab:bt-datasets}). The English samples undergo a two-stage cleanup: (1) rule-based filtering to remove duplicates, malformed text, and out-of-range lengths, and (2) LLM-based refinement to eliminate non-English or low-quality sentences and improve clarity while preserving meaning (see prompt for LLM-based refinement in Appendix~\ref{app:BT-english-text-filtering}). The resulting clean English corpus is then back-translated using the {English-to-$\ell$} model.

\vspace{0.4em}\noindent\textbf{LLM-based post-editing of MT outputs.}
MT systems trained on limited data often produce outputs with lexical inaccuracies or subtle word-choice errors, especially across semantically similar domains~\cite{bapna-etal-2022-next-thousand,nielsen-etal-2025-alligators}.
In contrast, multilingual LLMs, though performing poorly at \emph{direct} translation in low-resource languages, possess strong general-purpose reasoning capabilities, making them well suited as post-editors for refining MT outputs~\cite{nielsen-etal-2025-alligators}.
We therefore employ \texttt{Azure OpenAI GPT-5-chat}  to perform light post-editing, where the model applies minimal corrections to improve grammatical accuracy, lexical precision, and overall fluency while preserving the meaning and structure of the original NMT translation.
The LLM is explicitly prompted to avoid unnecessary rephrasing and instead focus on correcting mistranslated words and minor inconsistencies to enhance translation quality.
The post-editing prompt template is provided in Appendix~\ref{app:llm-postedit-prompt}.

\section{Experiments and Analysis}
To demonstrate the effectiveness of our pipeline, we evaluate two pathways: (1) adapting the LLM directly to target low-resource languages, and (2) enabling translation-mediated LLM access for extreme-low-resource languages.

\subsection{Direct LLM Adaptation for Chichewa and Māori}

\subsubsection{Experimental Details}
\vspace{0.4em}\noindent\textbf{Baseline model selection.}
We choose \GEMMA~\cite{team2025gemma} as our starting point due to its stronger performance on Chichewa and Māori, as we shall see in ablation experiments. We compare our adapted models against several LLMs, including Llama-3.1~\cite{grattafiori2024llama}, \QWEN~\cite{yang2025qwen3}, \OSS (medium reasoning mode)~\cite{openai2025gptoss120bgptoss20bmodel}, \APERTUS~\cite{hernandez2025apertus}, and Azure OpenAI GPT-4o.

\vspace{0.4em}\noindent\textbf{Training datasets.}
Our experiments use bilingual data mixtures (English and the target language) in both continual pretraining and supervised finetuning stages. For CPT, the corpus combines real LRL text from \FINEWEB~\cite{penedo2025fineweb2}, synthetic LRL text obtained by translating the English \FineWebEdu corpus~\cite{penedo2024fineweb}, and English text from \FineWebEdu~\cite{penedo2024fineweb}. These components are mixed at a 1:1:1 ratio, and all text is refined using the data curation strategy described in Sec.~\ref{sec:lrl}. The final CPT mixtures contain approximately 433M tokens for Chichewa and 745M tokens for Māori.

For SFT, we assemble a bilingual instruction dataset from several sources. We use native QA pairs from the \AYA dataset~\cite{singh-etal-2024-aya} when available for the target LRL. To increase coverage, we translate \AYA QA pairs from five high-resource languages (English, French, Dutch, Spanish, and Italian) into the target language. We also translate \SMOLTALK~\cite{bakouch2025smollm3} into the target LRL and include its English samples to maintain cross-lingual alignment. The detailed SFT dataset composition is shown in \ref{annex:SFT_Data} (Table~\ref{tab:chat-data-merged}).

\vspace{0.4em}\noindent\textbf{Benchmarking datasets and evaluation metrics.}
Existing multilingual benchmarks provide little-to-no coverage for most low-resource languages, and Chichewa and Māori are present only in FLORES-200~\cite{costa2022no} (translation) and Belebele~\cite{bandarkar-etal-2024-belebele} (reading comprehension). To perform comprehensive evaluation, we introduce professionally translated versions of \GLOBALMMLU~\cite{singh2024global} for both languages. We further generate machine-translated variants of ten benchmarks: ARC-Easy/Hard~\cite{clark2018think}, MGSM~\cite{shi2022language}, XCOPA~\cite{ponti2020xcopa}, StoryCloze~\cite{lin2021few}, PIQA~\cite{bisk2020piqa}, HellaSwag~\cite{zellers2019hellaswag}, XNLI-2.0~\cite{upadhyay2023xnli}, XWinograd~\cite{tikhonov2021s}, and TruthfulQA-Multi~\cite{lin2021truthfulqa}. All translated benchmarks are integrated into the \texttt{lm-evaluation-harness} framework~\cite{eval-harness}.
We also report English performance of competing models on HumanEval~\cite{chen2021evaluating}, BBH~\cite{suzgun2022challenging}, GPQA-Diamond~\cite{rein2024gpqa}, and IFEval~\cite{zhou2023instruction}. In addition, we run pairwise comparisons using an LLM-as-a-judge setup on MultiWikiQA~\cite{MultiWikiQA2025}.
Throughout the paper, we report scores for each benchmark/task using its standard evaluation metric (accuracy, BLEU, chrF++, etc.). The average score (reported as a percentage) is computed by normalizing each metric to the $[0, 1]$ range and using chrF++ for the translation task. Details on evaluation benchmarks and metrics are provided in Appendix~\ref{annex:base-eval-datasets} (Table~\ref{tab:pre-train-eval}) for base models, and in Appendix~\ref{annex:it-eval-datasets} (Table~\ref{tab:instruct-eval}) for instruction-tuned models.

\vspace{0.4em}\noindent\textbf{Hyperparameters.}
We train separate models, \BYOLC and \BYOLM, at three different sizes: 1B, 4B, and 12B parameters. All models are optimized with AdamW ($\beta_1=\text{0.9}$, $\beta_2=\text{0.999}$) for $4$ epochs. The learning rate is set to $2 \times 10^{-5}$ and gradually reduced to $2 \times 10^{-6}$ using cosine annealing, with a linear warm-up over the first 3\% of training iterations. During training, we set the maximum sequence length to 4096 tokens. We use the same hyperparameters for SFT, except that we use a lower learning rate of $1 \times 10^{-5}$ and train for 2 epochs.

\subsubsection{Performance Evaluation}
\vspace{0.4em}\noindent\textbf{Base model results.}
Tables \ref{tab:pre-train-resl-nya} and \ref{tab:pre-train-resl-mri} present base model comparisons on several Chichewa (\texttt{nya}) and Māori (\texttt{mri}) benchmarks. Across both languages, our \BYOL models provide consistent gains at similar parameter scales and often surpass significantly larger models. For example, in Table \ref{tab:pre-train-resl-nya}, the 4B \BYOLC model yields a 13.52 point average improvement over the 2$\times$ larger \APERTUS~(8B)~\cite{hernandez2025apertus} baseline and a 1.24 gain over the 3$\times$ larger \GEMMA~(12B-PT)~\cite{team2025gemma} model.
Similarly, Table \ref{tab:pre-train-resl-mri} shows that our \BYOLM model obtains an average score of 47.72, compared to 44.88 for the \GEMMA~(12B-PT) model, while having 3$\times$ fewer parameters.  Notably, our continual-pretrained \BYOL models achieve these gains while preserving the English performance of the \GEMMA~(PT) baselines; see \ref{annex:eng-perf-cpt} (Table~\ref{tab:pre-train-eng}).


\begin{table}[!t]
\centering
\caption{\small \underline{\textbf{Base model results on Chichewa}} (\texttt{nya}) language benchmarks. Our \BYOLC (CPT) models yield significant gains, and notably the 4B variant surpasses Gemma-3 (12B-PT), despite being 3$\times$ smaller.}
\label{tab:pre-train-resl-nya}
\adjustbox{width=\textwidth,center}{
\begin{tabular}{ll|cccc|ccccc|ccc}
\toprule[1.2pt]
\multirow{2}{*}{\textbf{Benchmarks}} & & 
\multicolumn{4}{c|}{\textbf{1B -- 2B Models}} & 
\multicolumn{5}{c|}{\textbf{4B -- 8B Models}} & 
\multicolumn{3}{c}{\textbf{12B+ Models}} \\ 
\cmidrule(lr){3-6} \cmidrule(lr){7-11} \cmidrule(lr){12-14}
& & 
\rotatebox[origin=c]{80}{\makecell{\textbf{Llama-3.2}~\cite{grattafiori2024llama} \\ \texttt{(1B)}}} &
\rotatebox[origin=c]{80}{\makecell{\textbf{Qwen-3}~\cite{yang2025qwen3} \\ \texttt{(1.7B-Base)}}} &
\rotatebox[origin=c]{80}{\makecell{\textbf{Gemma-3}~\cite{team2025gemma} \\ \texttt{(1B-PT)}}} &
\rotatebox[origin=c]{80}{\makecell{\textbf{\BYOLC} \\ \texttt{(1B-CPT)}}} &
\rotatebox[origin=c]{80}{\makecell{\textbf{Llama-3.1}~\cite{grattafiori2024llama} \\ \texttt{(8B)}}} &
\rotatebox[origin=c]{80}{\makecell{\textbf{Apertus}~\cite{hernandez2025apertus} \\ \texttt{(8B-2509)}}} &
\rotatebox[origin=c]{80}{\makecell{\textbf{Qwen-3}~\cite{yang2025qwen3} \\ \texttt{(8B-Base)}}} &
\rotatebox[origin=c]{80}{\makecell{\textbf{Gemma-3}~\cite{team2025gemma} \\ \texttt{(4B-PT)}}} &
\rotatebox[origin=c]{80}{\makecell{\textbf{\BYOLC} \\ \texttt{(4B-CPT)}}} &
\rotatebox[origin=c]{80}{\makecell{\textbf{Qwen-3}~\cite{yang2025qwen3} \\ \texttt{(14B-Base)}}} &
\rotatebox[origin=c]{80}{\makecell{\textbf{Gemma-3}~\cite{team2025gemma} \\ \texttt{(12B-PT)}}} &
\rotatebox[origin=c]{80}{\makecell{\textbf{\BYOLC} \\ \texttt{(12B-CPT)}}} \\
\midrule[1.2pt]
Global MMLU-Lite~\cite{singh2024global} & & 28.25 &  \bf 32.50 & 26.75 & 23.00 & 43.25 & 41.75 & 37.50 & 50.75 &   \bf 55.25 & 45.75 & 60.75 & \bf 64.50 \\
ARC-Easy~\cite{clark2018think}  & & 28.66 & 28.83 & 29.04 &  \bf 36.66 & 28.62 & 31.02 & 29.63 & 30.22 &   \bf 48.48 & 29.00 & 39.98 & \bf 51.14 \\
ARC-Hard~\cite{clark2018think} & & 22.44 & 21.93 & 23.72 &   \bf 27.56 & 24.06 & 27.73 & 23.98 & 27.13 & \bf 40.61 & 25.68 & 35.41 & \bf 42.41 \\
MGSM~\cite{shi2022language} &  &  \phantom{0}1.6 & \phantom{0}7.60 & \phantom{0}2.00 & \bf \phantom{0}3.20 & \phantom{0}8.40 & 10.40 & 10.00 & 17.20 &  \bf 31.60 & 14.40 & 44.00 & \bf 53.20 \\
 \midrule 
XCOPA~\cite{ponti2020xcopa}  & & 48.60 & 52.00 & 51.80 & \bf 61.20 & 51.20 & 53.20 & 52.20 & 57.20 &  \bf 70.00 & 51.00 & 60.80 & \bf 71.20 \\
XStoryCloze~\cite{lin2021few} & & 48.38 & 47.45 & 50.96 & \bf 55.79 & 50.43 & 53.74 & 49.11 & 54.40 & \bf  65.98 & 50.83 & 63.53 & \bf 67.90 \\ 
PIQA~\cite{bisk2020piqa}  & & 51.74 & 51.25 & 51.69 &  \bf 58.54 & 51.74 & 53.86 & 51.09 & 54.57 &  \bf 63.71 & 52.39 & 58.27 & \bf 64.96 \\
HellaSwag~\cite{zellers2019hellaswag} & & 29.19 & 29.93 & 29.05 &  \bf 37.13 & 29.56 & 31.98 & 27.81 & 33.45 & \bf 47.31 & 30.69 & 44.09 & \bf 51.89 \\
\midrule 
XNLI 2.0~\cite{upadhyay2023xnli} & & 33.79 & 33.07& 34.19 & \bf 37.92 & 33.81 & 35.01 & 34.73 & 37.82 &  \bf 40.32 & 33.95 & 40.98 & \bf 45.21 \\
XWinograd~\cite{tikhonov2021s}  & & 50.59 & 49.20 & 51.34 &  \bf 63.32 & 50.59 & 56.15 & 52.09 & 54.76 &   \bf 68.34 & 51.12 & 61.39 & \bf 70.37 \\
Belebele~\cite{bandarkar-etal-2024-belebele} & & 27.56 & \bf  29.22 & 28.11 &  26.00 & 29.22 & 38.78 & 32.33 & 38.22 &  \bf 45.44 & 36.56 & 59.56 & \bf 61.00 \\
\midrule
\multirow{2}{*}{ FLORES-200~\cite{costa2022no} (\texttt{nya}$\rightarrow$\texttt{eng})}  & BLEU &  \phantom{0}2.81 & \phantom{0}1.03 & \phantom{0}5.02 &  \bf 14.77 & \phantom{0}9.40 & 18.92 & \phantom{0}2.95 & 17.28 &  \bf 23.87 & \phantom{0}6.19 & 25.59 & \bf 27.84 \\
& chrF++ & 19.75 & 15.26 & 24.17 & \bf 38.18 & 31.38 & 42.25 & 21.97 & 40.37 &  \bf 47.95 & 27.85 & 48.91 & \bf 51.12 \\
\multirow{2}{*}{  FLORES-200~\cite{costa2022no}  (\texttt{eng}$\rightarrow$\texttt{nya}) } & BLEU &  \phantom{0}0.43 & \phantom{0}0.04 & \phantom{0}0.40 &  \bf \phantom{0}9.53 & \phantom{0}0.71 & \phantom{0}2.16 & \phantom{0}0.04 & \phantom{0}2.24 &   \bf 12.79 & \phantom{0}0.07 & \phantom{0}9.66 & \bf 13.82 \\
& chrF++  & 11.17 & \phantom{0}2.24 & 12.90 & \bf 40.52 & 13.69 & 21.98 & 2.18 & 23.22 &  \bf 48.66 & \phantom{0}2.77 & 41.13 & \bf 49.47 \\
\midrule[1.2pt]
\bf  Average Score  & &  30.90 & 30.81  & 31.98	&	\bf 39.16 & 34.30  & 38.30 & 32.66 &  39.95 & \bf 51.82		&34.77  & 50.68	& \bf 57.26 \\
\bottomrule[1.2pt]
\end{tabular}
}
\end{table}

\begin{table}[!t]
\centering
\caption{\small\underline{\textbf{Base model results on Māori}} (\texttt{mri}) benchmarks. Our \BYOLM (4B -CPT) model obtains an average score of 47.72  outperforming the Gemma-3 (12B-PT) model, which achieves 44.88.}
\label{tab:pre-train-resl-mri}
\adjustbox{width=\textwidth,center}{
\begin{tabular}{ll|cccc|ccccc|ccc}
\toprule[1.2pt]
\multirow{2}{*}{\textbf{Benchmarks}} & & 
\multicolumn{4}{c|}{\textbf{1B -- 2B Models}} & 
\multicolumn{5}{c|}{\textbf{4B -- 8B Models}} & 
\multicolumn{3}{c}{\textbf{12B+ Models}} \\ 
\cmidrule(lr){3-6} \cmidrule(lr){7-11} \cmidrule(lr){12-14}
& & 
\rotatebox[origin=c]{80}{\makecell{\textbf{Llama-3.2}~\cite{grattafiori2024llama} \\ \texttt{(1B)}}} &
\rotatebox[origin=c]{80}{\makecell{\textbf{Qwen-3}~\cite{yang2025qwen3} \\ \texttt{(1.7B-Base)}}} &
\rotatebox[origin=c]{80}{\makecell{\textbf{Gemma-3}~\cite{team2025gemma} \\ \texttt{(1B-PT)}}} &
\rotatebox[origin=c]{80}{\makecell{\textbf{\BYOLM} \\ \texttt{(1B-CPT)}}} &
\rotatebox[origin=c]{80}{\makecell{\textbf{Llama-3.1}~\cite{grattafiori2024llama} \\ \texttt{(8B)}}} &
\rotatebox[origin=c]{80}{\makecell{\textbf{Apertus}~\cite{hernandez2025apertus} \\ \texttt{(8B-2509)}}} &
\rotatebox[origin=c]{80}{\makecell{\textbf{Qwen-3}~\cite{yang2025qwen3} \\ \texttt{(8B-Base)}}} &
\rotatebox[origin=c]{80}{\makecell{\textbf{Gemma-3}~\cite{team2025gemma} \\ \texttt{(4B-PT)}}} &
\rotatebox[origin=c]{80}{\makecell{\textbf{\BYOLM} \\ \texttt{(4B-CPT)}}} &
\rotatebox[origin=c]{80}{\makecell{\textbf{Qwen-3}~\cite{yang2025qwen3} \\ \texttt{(14B-Base)}}} &
\rotatebox[origin=c]{80}{\makecell{\textbf{Gemma-3}~\cite{team2025gemma} \\ \texttt{(12B-PT)}}} &
\rotatebox[origin=c]{80}{\makecell{\textbf{\BYOLM} \\ \texttt{(12B-CPT)}}} \\
\midrule[1.2pt]
Global MMLU-Lite~\cite{singh2024global} & & 26.75& \bf  34.25 & 22.00 & 24.25 & 	38.00 & 	35.25 & 	42.00 & 	39.00 & 	\bf 45.50 & 	47.00 & 45.00 & 	\bf 49.00\\
ARC-Easy~\cite{clark2018think}  & & 25.63 & 26.05 & 26.35 & \bf 30.81 & 	26.60 & 	25.72 & 	26.81 & 	26.52 & 	\bf 43.73 & 	26.14 & 29.50 & 	\bf 41.16 \\
ARC-Hard~\cite{clark2018think} & & 18.86 & 19.62 &  18.52 & \bf 22.87 & 	20.05 & 	20.90 & 	21.42 & 	21.08 & 	\bf 32.17 & 	22.35 &  23.46 & 	\bf 32.59 \\
MGSM~\cite{shi2022language}  & & \phantom{0}2.00 & \bf  \phantom{0}6.80 &  \phantom{0}0.40 &  \phantom{0}2.80 & 	13.20 & 	15.60 & 	33.20 & 	14.00 & 	\bf 24.00 & 	37.60 & 58.44 & 	\bf 52.00 \\
 \midrule 
XCOPA~\cite{ponti2020xcopa}  & & 52.60 & 52.80 &  52.20 & \bf 57.00 & 	55.80 & 	54.80 & 	51.80 & 	52.80 & 	\bf 60.60 & 	53.60 & 55.60  & 	\bf 61.20 \\
XStoryClozee~\cite{lin2021few}  & & 47.85& 47.78 & 49.04 & \bf 56.59 & 	51.36 & 	54.40 & 	51.03 & 	51.75 & \bf	63.34 & 	52.42 & 42.80 & 	\bf 64.00 \\ 
PIQA~\cite{bisk2020piqa}  & & 53.59 & 53.81 &  52.45 & \bf 57.29 & 	54.57 & 	55.44 & 	54.62 & 	54.30 & 	\bf 61.86 & 	55.55 & 56.64 & 	\bf 61.43 \\
HellaSwag~\cite{zellers2019hellaswag}  & & 26.98 & 27.16&  26.98 & \bf 30.83 & 	28.23 & 	29.22 & 	27.84 & 	28.68 & \bf	37.80 & 	28.51 & 31.78 & \bf	38.11 \\
\midrule 
XNLI 2.0~\cite{upadhyay2023xnli} & & 33.57 & 32.38 & 32.61 & \bf 39.36 & 	35.21 & 	36.63 & 	35.85 & 	34.97 &  \bf	44.57 & 	40.98 & 41.66 & \bf	44.51 \\
XWinograd~\cite{tikhonov2021s} & & 49.41 & 48.66 &  49.95 & \bf 57.33 & 	52.83 & 	53.05 & 	52.30 & 	52.83 & 	\bf 59.68 & 	52.19 & 56.47 & \bf	62.67 \\
Belebele~\cite{bandarkar-etal-2024-belebele} & & 26.22 & \bf  30.44 & 27.44 &  27.67 & 	34.78 & 	36.67 & 	43.33 & 	34.44 & \bf	47.78 & 	48.33  & 59.11 & 	\bf 63.56 \\
\midrule
\multirow{2}{*}{ FLORES-200~\cite{costa2022no} (\texttt{mri}$\rightarrow$\texttt{eng})}  & BLEU  & \phantom{0}2.94 & \phantom{0}0.39&  \phantom{0}3.37 & \bf 17.45 & 	16.36 & 	18.93 & 	8.96 & 	14.35 & \bf	25.93 & 	17.31 & 23.26 & 	\bf 30.15 \\
& chrF++ & 20.09 & 10.31 & 21.11 & \bf 40.74 & 	40.23 & 	43.21 & 	32.89 & 	37.34 & \bf	49.78 & 	40.42 & 46.91 & \bf	53.04 \\
\multirow{2}{*}{  FLORES-200~\cite{costa2022no}  (\texttt{eng}$\rightarrow$\texttt{mri}) } & BLEU & \phantom{0}0.51 & \phantom{0}0.08 &  \phantom{0}0.44 & \bf 18.16 & 	\phantom{0}4.05 & 	\phantom{0}5.50 & 	\phantom{0}0.21 & 	\phantom{0}2.59 & 	\bf 24.41 & 	\phantom{0}0.55 & 10.55 & 	\bf 25.05 \\
& chrF++ & 13.24 & \phantom{0}3.42 &  14.68 & \bf 42.69 & 	25.63 & 	27.48 & 	\phantom{0}5.05 & 	21.14 & 	\bf 49.55 & 	\phantom{0}9.03 & 36.08 & \bf	49.68\\
\midrule[1.2pt]
\bf  Average Score  & &   24.98 &  25.55 & 30.29 & \bf 37.71 & 36.65 & 37.57 & 36.78 & 36.07 & \bf 47.72 & 39.55 & 44.88 & \bf 51.77 \\
\bottomrule[1.2pt]
\end{tabular}
}
\end{table}

\vspace{0.4em}\noindent\textbf{Instruction-tuned model results.}
Tables~\ref{tab:chat-resl-nya} and~\ref{tab:chat-resl-mri} report performance comparisons of our \BYOL models against several competing LLMs of varying parameter capacities on Chichewa and Māori benchmarks, respectively. These \BYOL models are obtained via supervised finetuning followed by model merging (see Sec.~\ref{sec:lrl}).
For Chichewa, Table \ref{tab:chat-resl-nya} shows that our \BYOLC (4B-M) model obtains an average performance boost of 15.20\% over \APERTUS~(8B-Instruct) model, and a 1.00\% gain over a $\sim$7$\times$ larger model \GEMMA~(27B-IT). Similar trends can be observed for \BYOLM in Table~\ref{tab:chat-resl-mri}. Despite these strong language-specific improvements, our models also preserve English performance, as shown in Table~\ref{tab:chat-eng}.

\begin{figure}[!t]
    \centering

    \subfloat[\scriptsize \BYOLC (4B-M) model performance.\label{fig:llm-judge-win-loss-4b-nya}]{
        \includegraphics[width=0.47\linewidth]{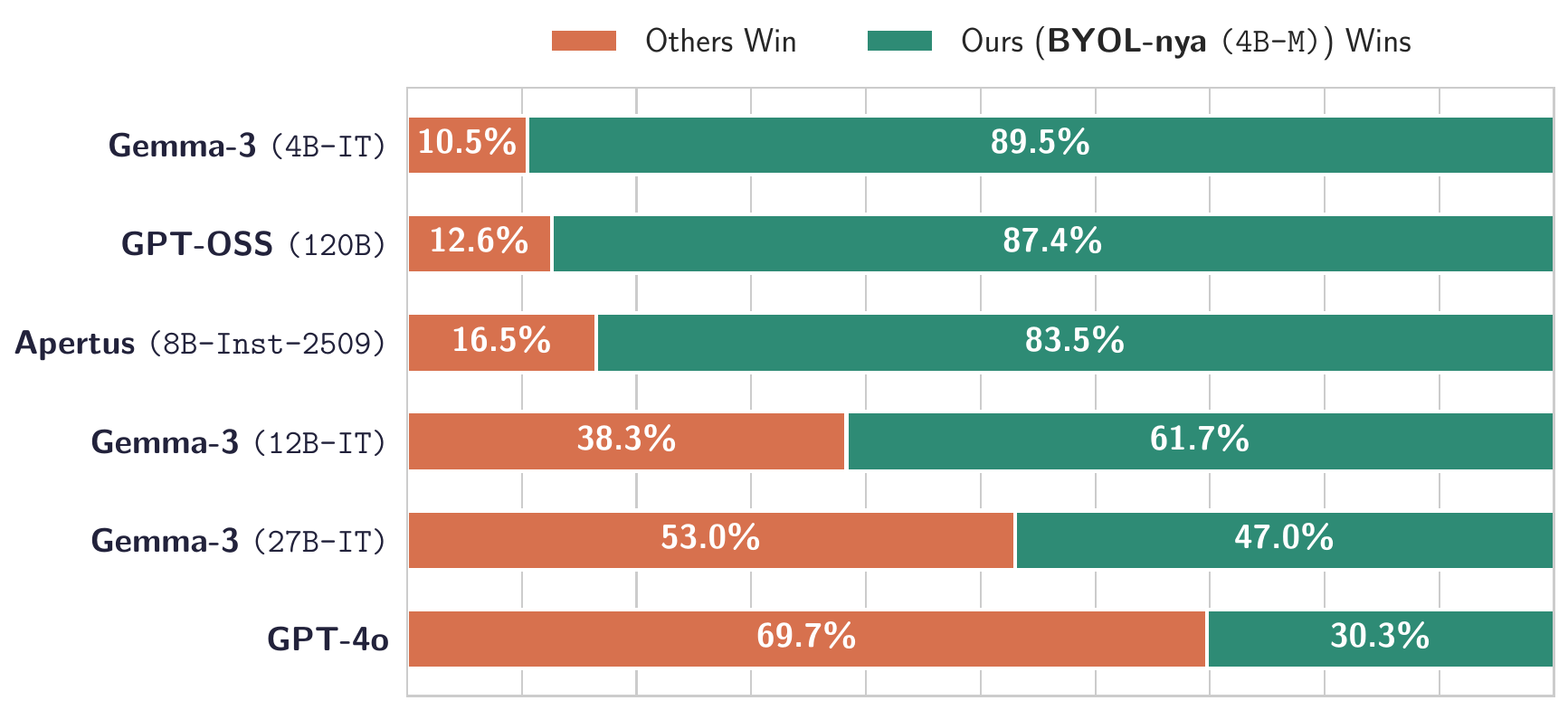}
    }
    \hfill
    \subfloat[\scriptsize \BYOLC (12B-M) model performance.\label{fig:llm-judge-win-loss-12b-nya}]{
        \includegraphics[width=0.47\linewidth]{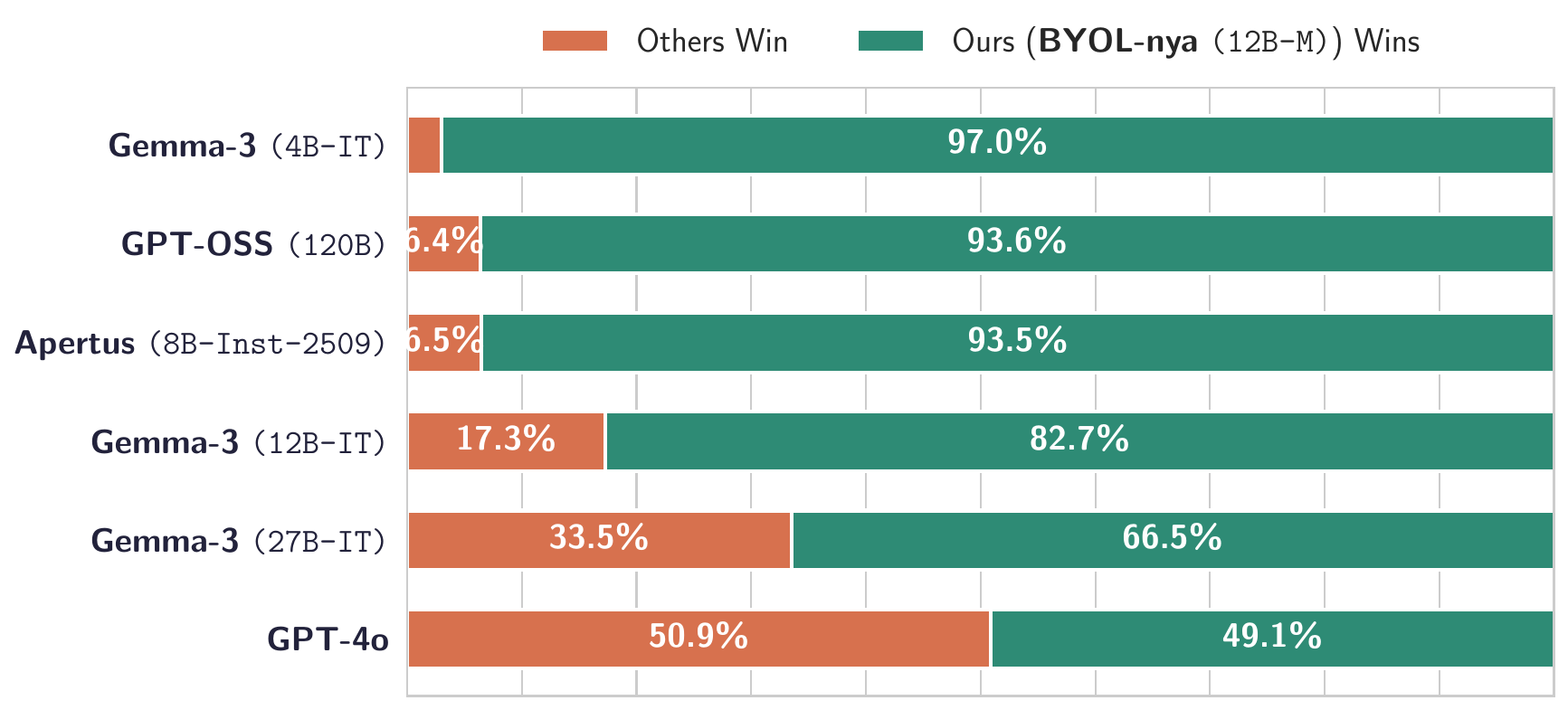}
    }

    \vspace{0.8em}

    \subfloat[\scriptsize \BYOLM (4B-M) model performance.\label{fig:llm-judge-win-loss-4b-mri}]{
        \includegraphics[width=0.47\linewidth]{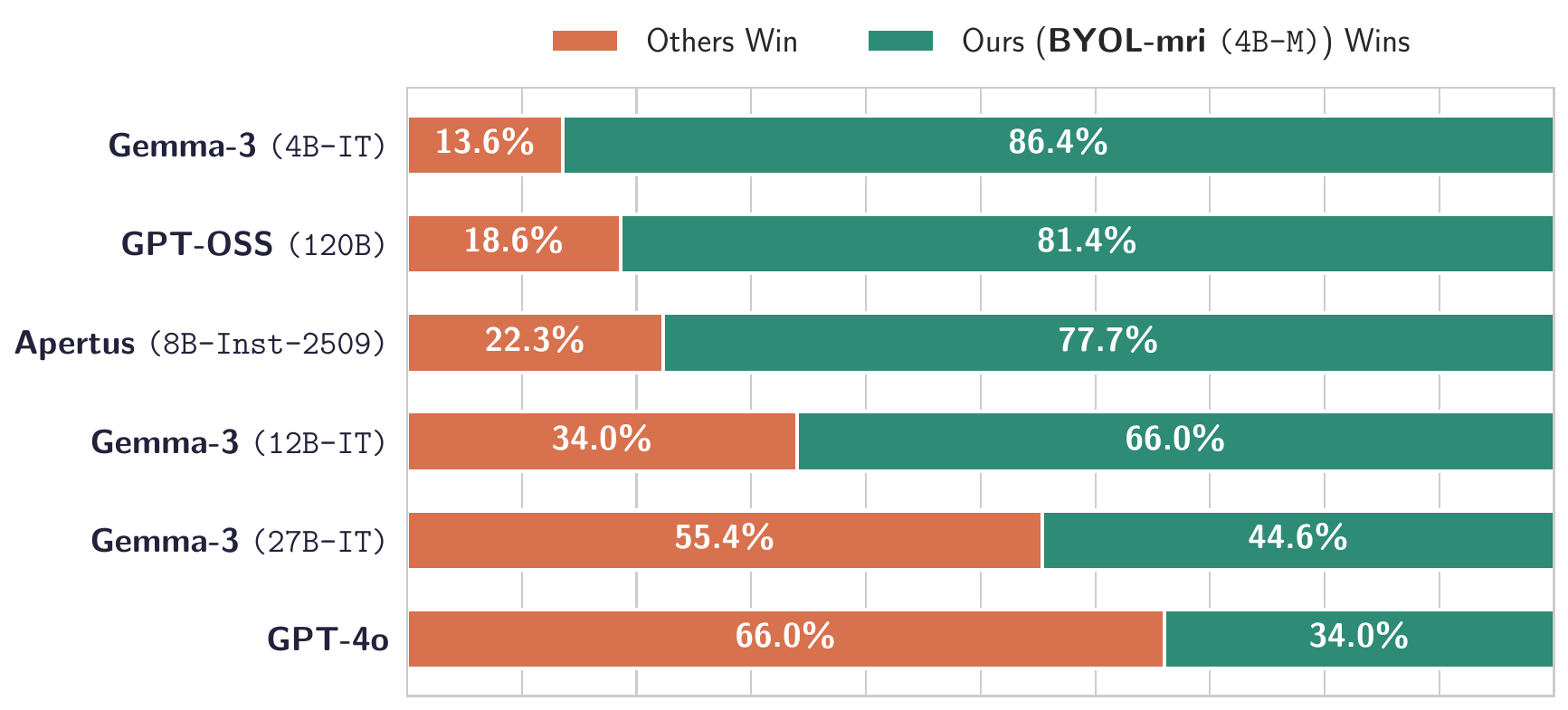}
    }
    \hfill
    \subfloat[\scriptsize  \BYOLM (12B-M) model performance.\label{fig:llm-judge-win-loss-12b-mri}]{
        \includegraphics[width=0.47\linewidth]{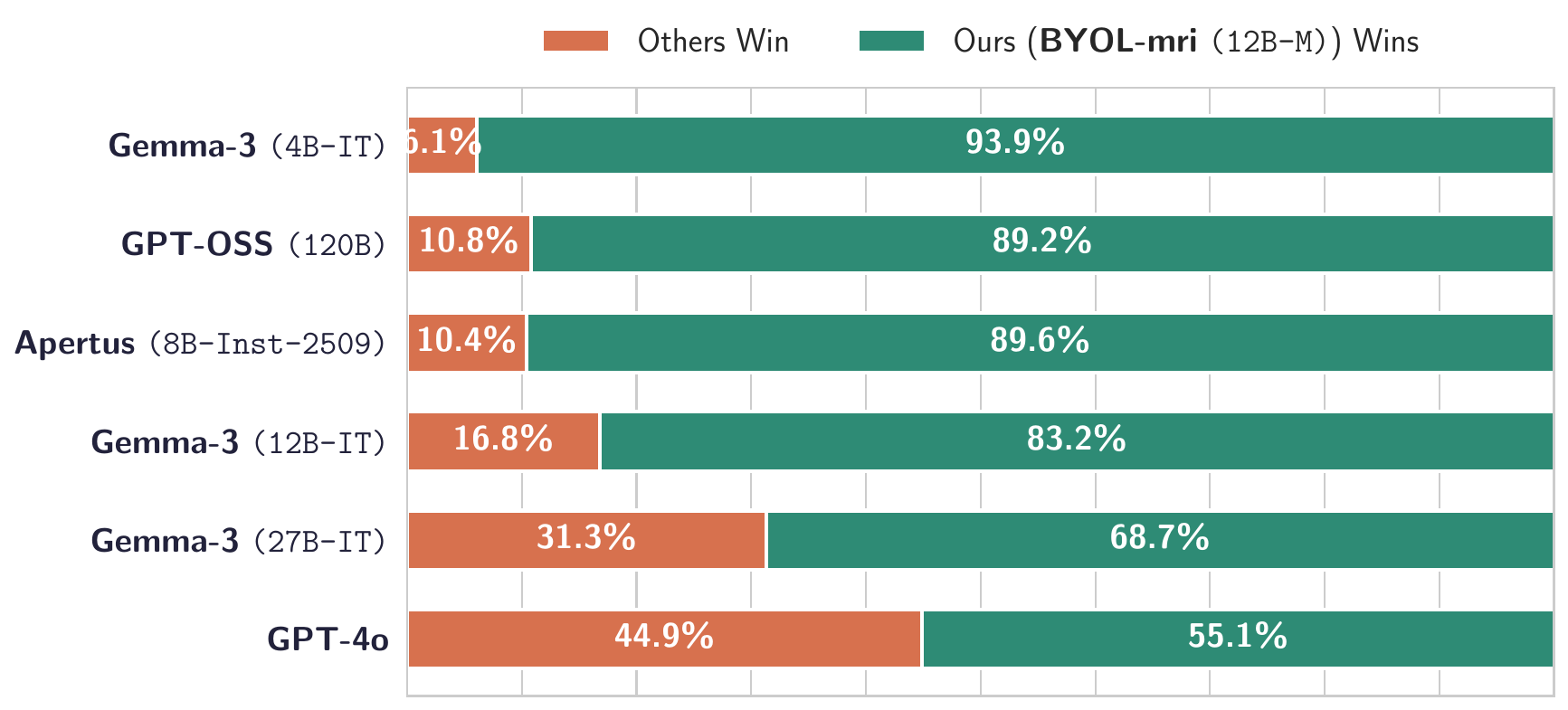}
    }

    \vspace{-0.5em}
    \caption{\small \underline{\textbf{LLM-as-a-judge win–loss comparisons}} on Multi-Wiki-QA~\cite{MultiWikiQA2025}. 
    Pairwise evaluations against competing LLMs using GPT-5-chat as the judge. Our \BYOL models achieve strong win rates on Chichewa and Māori; in particular, our 12B models, in (b) and (d), surpass \GEMMA~(27B-IT) and perform on par with GPT-4o.}
    \label{fig:llm-judge-win-loss}
\end{figure}

We further evaluate the generative capability of our chat models under an LLM-as-a-judge setting. For 1,000 questions from the MultiWikiQA~\cite{MultiWikiQA2025} reading comprehension dataset, we generate responses from all competing LLMs and compare them in a pairwise manner using GPT-5-chat as the judge. The evaluator selects the answer that is closer to the reference under a forced-choice protocol, i.e., no ties allowed. See Annex~\ref{subsec:llm-as-judge-template} for the evaluation prompt template.
The win-rate results in Fig.~\ref{fig:llm-judge-win-loss} show that our 4B models exceed the performance of substantially larger baselines. The \BYOL (12B-M) outputs are consistently preferred by the GPT-5-chat judge across both Chichewa and Māori, surpassing \GEMMA~(27B-IT) and achieving performance on par with GPT-4o.


\begin{table}[!t]
\centering
\caption{\small\underline{\textbf{Instruction-tuned model results on Chichewa}} (\texttt{nya}) language benchmarks. Zero-shot evaluation on 12 datasets shows our \BYOLC (M) models consistently achieve state-of-the-art performance.}
\label{tab:chat-resl-nya}
\adjustbox{width=\textwidth,center}{
\begin{tabular}{ll|ccccc|ccccc}
\toprule[1.2pt]
\multirow{2}{*}{\textbf{Benchmarks}} & & 
\multicolumn{5}{c|}{\textbf{4B -- 8B Models}} & 
\multicolumn{5}{c}{\textbf{12B+ Models}} \\ 
 \cmidrule(lr){3-7} \cmidrule(lr){8-12}
& & 
\rotatebox[origin=c]{65}{\makecell{\textbf{Llama-3.1}~\cite{grattafiori2024llama}\\\texttt{(Instruct-8B)}}} &
\rotatebox[origin=c]{65}{\makecell{\textbf{Apertus}~\cite{hernandez2025apertus}\\\texttt{(8B-Inst-2509)}}} &
\rotatebox[origin=c]{65}{\makecell{\textbf{Qwen-3}~\cite{yang2025qwen3}\\\texttt{(8B)}}} &
\rotatebox[origin=c]{65}{\makecell{\textbf{Gemma-3}~\cite{team2025gemma}\\\texttt{(4B-IT)}}} &
\rotatebox[origin=c]{65}{\makecell{\textbf{\BYOLC}\\\texttt{(4B-M)}}} &
\rotatebox[origin=c]{65}{\makecell{\textbf{Qwen-3}~\cite{yang2025qwen3}\\\texttt{(14B)}}} &
\rotatebox[origin=c]{65}{\makecell{\textbf{GPT-OSS}~\cite{openai2025gptoss120bgptoss20bmodel}\\\texttt{(20B)}}} &
\rotatebox[origin=c]{65}{\makecell{\textbf{Gemma-3}~\cite{team2025gemma}\\\texttt{(12B-IT)}}} &
\rotatebox[origin=c]{65}{\makecell{\textbf{Gemma-3}~\cite{team2025gemma}\\\texttt{(27B-IT)}}} &
\rotatebox[origin=c]{65}{\makecell{\textbf{\BYOLC}\\\texttt{(12B-M)}}} \\
\midrule[1.2pt]
Global MMLU-Lite~\cite{singh2024global} &  &33.30 & 34.83 & 36.45 & 45.36 & \bf  53.62 & 38.38 & 41.47 & 54.97 & 62.64 & \bf  66.15  \\
 ARC-Hard chat~\cite{clark2018think}  &  &  26.54 & 32.51 & 17.66 & 33.28 & \bf  50.43 & 27.82 & 31.57 & 55.38 & \bf 66.13 &  64.76  \\
 MGSM~\cite{shi2022language} &  &  	\phantom{0}8.80 & 	\phantom{0}2.40 & 	\phantom{0}2.40 & 11.20 &  \bf  30.00 & 	\phantom{0}3.60 & \phantom{0}3.50 & 37.60 & 38.80 & \bf  40.80 \\
 \midrule 
XCOPA~\cite{ponti2020xcopa} &  &  50.40 & 51.20 & 51.40 & 52.20 & \bf 66.40 & 54.00 & 52.20 & 54.00 & 54.80 & \bf   65.60 \\
XStoryCloze~\cite{lin2021few}  &  &  48.97 & 50.76 & 48.78 & 49.31 &  \bf 59.23  & 51.36 & 44.87& 55.53 & 58.31 & \bf  62.61 \\
PIQA~\cite{bisk2020piqa}  &  &  51.41 & 52.39 & 51.25 & 52.77 &\bf  61.75  & 52.01 & 50.44 & 55.28 & 58.38 & \bf 63.76 \\
HellaSwag~\cite{zellers2019hellaswag}  &  &  29.88 & 30.58 & 29.00 & 29.10 & \bf 45.32 & 29.54 & 25.72 & 35.08 & 40.99 & \bf  49.16 \\
\midrule 
XNLI 2.0~\cite{upadhyay2023xnli} &  & 33.05 & 35.81 & 32.63 & 35.75 & \bf 38.18 &  33.41 & 32.10 & 38.18 & 35.67	 & \bf 42.51 \\  
XWinograd~\cite{tikhonov2021s} &  &  51.98 & 50.80 & 51.98 & 52.41 & \bf 66.42  & 52.30 & 50.16 & 50.37 & 58.93 & \bf  66.20  \\
Belebele~\cite{bandarkar-etal-2024-belebele} &  &    27.11 & 34.56 & 22.11 & 29.00 & \bf   55.00 & 22.89 & 21.44& 43.22 & 52.67 & \bf 62.44  \\ \midrule
\multirow{2}{*}{ FLORES~\cite{costa2022no} (\texttt{nya}$\rightarrow$\texttt{eng})}  & BLEU   &  	\phantom{0}8.68 & 11.53 & 	\phantom{0}6.11 & 	11.97 &  \bf 24.96  & 	\phantom{0}8.38 & \phantom{0}1.70 & 	20.08 & 	22.84 & \bf 27.13  \\
& chrF++ &  30.84 & 37.71 & 27.08 & 35.26 & \bf 49.21  & 30.57 & 22.41 & 44.18 & 47.51 & \bf  50.77 \\
\multirow{2}{*}{  FLORES~\cite{costa2022no}  (\texttt{eng}$\rightarrow$\texttt{nya}) }& BLEU & 	\phantom{0}1.92 & 	\phantom{0}1.48 & 	\phantom{0}0.46 & 	\phantom{0}2.80 & \bf 13.31 & \phantom{0}0.94 & \phantom{0}1.00 & 	\phantom{0}0.98 & 	\phantom{0}8.17 & \bf 13.65  \\
& chrF++ & 22.23 & 18.77 & \phantom{0}9.95 & 25.38 & \bf  48.91 & 14.94 & 18.70 & 37.86 & 44.86 & \bf  49.43 \\
\midrule
TruthfulQA~\cite{lin2021truthfulqa} &  & 30.97 & 31.70 & 30.84 & 28.76 & \bf 36.11  & 33.05 & 36.47 & 17.38 & 29.01 & \bf  36.96  \\
\midrule[1.2pt]
\bf Average Score & & 34.27 & 35.69 & 31.66 & 36.91 & \bf 50.89 & 34.14 & 33.16  & 44.54 & 49.90 & \bf 55.47 \\
\bottomrule[1.2pt]
\end{tabular}
}
\end{table}

\begin{table}[!t]
\centering
\caption{\small\underline{\textbf{Instruction-tuned model results on Māori}} (\texttt{mri}) benchmarks. Zero-shot evaluation is performed. }
\label{tab:chat-resl-mri}
\adjustbox{width=\textwidth,center}{
\begin{tabular}{ll|ccccc|ccccc}
\toprule[1.2pt]
\multirow{2}{*}{\textbf{Benchmarks}} & & 
\multicolumn{5}{c|}{\textbf{4B--8B Models}} & 
\multicolumn{5}{c}{\textbf{12B+ Models}} \\ 
 \cmidrule(lr){3-7} \cmidrule(lr){8-12}
& & 
\rotatebox[origin=c]{65}{\makecell{\textbf{Llama-3.1}~\cite{grattafiori2024llama}\\\texttt{(Instruct-8B)}}} &
\rotatebox[origin=c]{65}{\makecell{\textbf{Apertus}~\cite{hernandez2025apertus}\\\texttt{(8B-Inst-2509)}}} &
\rotatebox[origin=c]{65}{\makecell{\textbf{Qwen-3}~\cite{yang2025qwen3}\\\texttt{(8B)}}} &
\rotatebox[origin=c]{65}{\makecell{\textbf{Gemma-3}~\cite{team2025gemma}\\\texttt{(4B-IT)}}} &
\rotatebox[origin=c]{65}{\makecell{\textbf{\BYOLM}\\\texttt{(4B-M)}}} &
\rotatebox[origin=c]{65}{\makecell{\textbf{Qwen-3}~\cite{yang2025qwen3}\\\texttt{(14B)}}} &
\rotatebox[origin=c]{65}{\makecell{\textbf{GPT-OSS}~\cite{openai2025gptoss120bgptoss20bmodel}\\\texttt{(20B)}}} &
\rotatebox[origin=c]{65}{\makecell{\textbf{Gemma-3}~\cite{team2025gemma}\\\texttt{(12B-IT)}}} &
\rotatebox[origin=c]{65}{\makecell{\textbf{Gemma-3}~\cite{team2025gemma}\\\texttt{(27B-IT)}}} &
\rotatebox[origin=c]{65}{\makecell{\textbf{\BYOLM}\\\texttt{(12B-M)}}} \\
\midrule[1.2pt]
Global MMLU-Lite~\cite{singh2024global}  &  &  30.44 & 31.05 & 33.27 & 35.10 & \bf 47.64 & 44.68 & 34.61 & 43.52 & \bf 54.64 &  52.48 \\
 ARC-Hard chat~\cite{clark2018think} &  &  31.91 & 27.30 & 34.04 & 37.50 &  \bf 51.11 & 35.75 & 32.85 & 43.09 & 49.66 & \bf  59.22 \\
 MGSM~\cite{shi2022language} &  &  15.20 & 4.80 & 7.20 & 10.80 & \bf  27.60 & \phantom{0}1.60 & \phantom{0}1.60 & 27.20 & \bf 41.60 &  38.80 \\
 \midrule 
XCOPA~\cite{ponti2020xcopa} &  &   54.60 & 52.00 & 52.60 & 53.60 & \bf 57.60  & 53.60 & 53.80 & 54.40 & 52.80 & \bf 59.80 \\
XStoryCloze~\cite{lin2021few} &  &  50.89 & 50.83 & 50.03 & 51.29 & \bf 57.78 & 51.03 & 48.58 & 54.00 & 54.53 & \bf 58.37  \\
PIQA~\cite{bisk2020piqa}   &  &  51.52 & 56.69 & 54.13 & 54.41 & \bf 60.17 & 53.70 & 53.54 & 53.10 & 56.37 & \bf 60.28 \\
HellaSwag~\cite{zellers2019hellaswag} &  &  31.20 & 29.12 & 27.57 & 28.21 & \bf 35.73 & 28.24 & 24.80 & 32.13 & 31.23 & \bf 37.46 \\
\midrule 
XNLI 2.0~\cite{upadhyay2023xnli} &  &  35.31 & 34.13 & 34.85 & 33.81 & \bf 42.55 & 34.89 & 34.31 & 39.50 & 37.78	 & \bf 42.51	\\
XWinograd~\cite{tikhonov2021s} &  &  49.52 & 51.12 & 50.80 & 51.55 & \bf 56.26 & 49.95 & 50.37 & 50.16 & 51.87 & \bf  57.33  \\
Belebele~\cite{bandarkar-etal-2024-belebele} &  &  28.00 & 33.22 & 22.89 & 25.56 & \bf 50.67  & 22.89 & 22.00 & 47.89 & 50.33 & \bf 62.78 \\ 
\midrule
\multirow{2}{*}{ FLORES~\cite{costa2022no} (\texttt{mri}$\rightarrow$\texttt{eng})}  & BLEU   &  12.78 & 14.22 & 11.83 & 11.64 & \bf 26.02  & 15.95 & \phantom{0}1.55 & 19.82 & 21.70 & \bf  28.14 \\
& chrF++  &  37.28 & 40.45 & 35.83 & 34.60 & \bf 50.75 & 40.48 & 21.31 & 43.90 & 46.79 & \bf  52.27 \\
\multirow{2}{*}{  FLORES~\cite{costa2022no}  (\texttt{eng}$\rightarrow$\texttt{mri}) }& BLEU & \phantom{0}6.69 & \phantom{0}4.42 & \phantom{0}2.71 & \phantom{0}5.53 & \bf 22.97 & \phantom{0}6.00 & \phantom{0}1.47 & 11.79 & 15.40 & \bf 24.28    \\
& chrF++  & 30.75 & 26.50 & 23.38 & 28.00 & \bf 48.48 & 29.34 & 18.87 & 36.94 & 41.51 & \bf 49.60 \\
\midrule
TruthfulQA~\cite{lin2021truthfulqa} &    &  38.19 & 39.66 & 42.47  & 36.23 & \bf 42.59 & 42.23 & 40.88 & 31.09 & 22.64 & \bf 49.69 \\
\midrule[1.2pt]
\bf Average Score & & 37.29 & 36.68 & 36.08 &  36.97	& \bf 48.38 & 37.57 & 33.66 & 42.84	&	45.52 &		\bf	52.35 \\
\bottomrule[1.2pt]
\end{tabular}
}
\end{table}

\subsubsection{Ablation Studies}
\label{sec:ablation}
We perform ablation experiments on Chichewa using the 4B \BYOL model unless mentioned otherwise.

\vspace{0.4em}\noindent\textbf{Identifying the baseline LLM and MT system.} 
To determine the base model for adaptation, we evaluate Llama-3.1~\cite{grattafiori2024llama}, \QWEN~\cite{yang2025qwen3}, and \GEMMA~\cite{team2025gemma} on RTTBench-Mono using the round-trip translation approach (Sec.~\ref{sec:tool-assessment}), where English sentences from 25 domains are translated into Chichewa and then back into English. We then compare the reconstructed English sentences with the original sentences.
Table~\ref{tab:init-assess-llm-ablation} shows that \GEMMA achieves the highest BLEU, chrF++, and embedding-similarity scores, and we therefore adopt it as our baseline model.

Using the same RTT setup, we evaluate MT systems and find Azure Translator to be the best performing: Table~\ref{tab:init-assess-mt-ablation} reports the fidelity scores, and Fig.~\ref{fig:init-assess-mt-per-domain-maps} (in \ref{annex:per-domain-mt-results}) shows that it leads in 18 of the 25 domains. We use it to generate synthetic Chichewa data by translating English text into Chichewa.

\begin{table*}[!t]
\centering

\begin{minipage}{0.4\linewidth}
\centering
\caption{\small \textbf{\underline{Baseline LLM selection ablation}} on RTTBench-Mono. 
Gemma-3 performs best and is therefore used for adaptation.}
\label{tab:init-assess-llm-ablation}

\adjustbox{width=\linewidth}{
\begin{tabular}{lccc}
\toprule[1.2pt]
& \multirow{2}{*}{\small \textbf{Llama-3.1}~\cite{grattafiori2024llama}}
& \multirow{2}{*}{\small \textbf{Qwen-3}~\cite{yang2025qwen3}}
& \multirow{2}{*}{\small \textbf{Gemma-3}~\cite{team2025gemma}} \vspace{0.3em} \\ 
& \small\texttt{(8B-Instruct)} 
& \small\texttt{(14B)} 
& \small\texttt{(12B-IT)} \\
\midrule
\small BLEU      $\uparrow$ & 3.07  & 3.33  & 11.01 \\
\small chrF++    $\uparrow$ & 15.86 & 20.01 & 34.21 \\
\small Similarity$\uparrow$ & 28.66 & 20.29 & 54.34 \\
\bottomrule[1.2pt]
\end{tabular}
}
\end{minipage}
\hfill
\begin{minipage}{0.57\linewidth}
\centering
\caption{\small \textbf{\underline{MT system selection ablation}} on RTTBench-Mono. 
Azure Translator achieves the best overall performance and leads on 18 of 25 domains; 
see Fig.~\ref{fig:init-assess-mt-per-domain-maps} for per-domain scores.}
\label{tab:init-assess-mt-ablation}

\adjustbox{width=\linewidth}{
\begin{tabular}{lccccc}
\toprule[1.2pt]
& \multirow{2}{*}{\small \textbf{NLLB-200}~\cite{costa2022no}}
& \multirow{2}{*}{\small \textbf{MADLAD-400}~\cite{kudugunta2023madlad}}
& \multirow{2}{*}{\small \textbf{GPT-4o}}
& \multirow{2}{*}{\small \textbf{Google}}
& \multirow{2}{*}{\small \textbf{Azure}} \vspace{0.3em}\\
& \small\texttt{(3.3B)} 
& \small\texttt{(7B-MT)} 
& \small\texttt{(OpenAI)} 
& \small\texttt{(Translate)} 
& \small\texttt{(Translator)} \\
\midrule
\small BLEU      $\uparrow$ & 18.29 & 22.91 & 34.02 & 42.40 & 44.94 \\
\small chrF++    $\uparrow$ & 42.76 & 47.26 & 59.91 & 64.95 & 67.58 \\
\small Similarity$\uparrow$ & 69.54 & 73.04 & 84.84 & 87.21 & 87.98 \\
\bottomrule[1.2pt]
\end{tabular}
}
\end{minipage}

\end{table*}

\begin{figure*}[!t]
\centering
\begin{minipage}[!t]{0.55\textwidth}
\centering
\captionof{table}{\small \underline{\textbf{Data mixture ablation.}}
Continual pretraining of the  \BYOLC (4B-CPT) model under different data mixtures is performed. 
Average score across several datasets is reported; see Table~\ref{tab:pre-train-ablation-sets-full} for per-dataset results. \GEMMA~(4B-PT) baseline scores are 39.95 (\texttt{nya}) and 65.17 (\texttt{eng}).}
\label{tab:pretrain_datasets}
\scalebox{0.7}{
\begin{tabular}{lccccc}
\toprule[1.2pt]
\textbf{Datasets} & \textbf{Language} & \textbf{C1} & \textbf{C2} & \textbf{C3} & \textbf{C4} \\
\midrule[1.2pt]
FineWeb2~\cite{penedo2025fineweb2} & \texttt{nya} & \cmark & \xmark & \xmark & \xmark \\
FineWeb2~\cite{penedo2025fineweb2} $\rightarrow$ \text{Refine (Sec.~\ref{sec:lrl})} & \texttt{nya} & \xmark & \cmark & \cmark & \cmark \\
\midrule
FineWeb-Edu~\cite{penedo2024fineweb} & \texttt{eng} & \xmark & \xmark & \xmark & \xmark \\
FineWeb-Edu~\cite{penedo2024fineweb} $\rightarrow$ \text{Refine (Sec.~\ref{sec:lrl})} & \texttt{eng} & \xmark & \xmark & \cmark & \cmark \\
\midrule
FineWeb-Edu~\cite{penedo2024fineweb} $\rightarrow$ \text{Refine} $\rightarrow$ \text{Translate} & \texttt{eng}$\rightarrow$\texttt{nya} & \xmark & \xmark & \xmark & \cmark \\
\midrule[1.2pt]
\multirow{2}{*}{\textbf{Average Score}} & \texttt{nya} & 48.77 & 49.44 & 49.60 & \textbf{51.82} \\
& \texttt{eng} & 64.58 & 65.24 & \textbf{65.37} & 65.29 \\
\bottomrule[1.2pt]
\end{tabular}
}

\end{minipage}
\hfill
\begin{minipage}[!t]{0.35\textwidth}
\centering
\includegraphics[width=\linewidth]{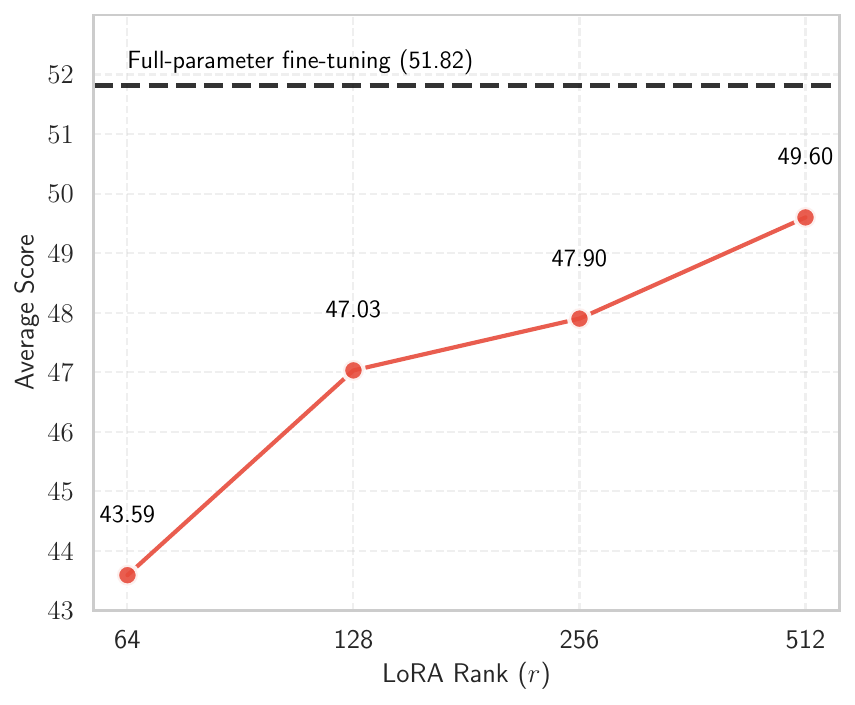}
\vspace{-2.2em}
\caption{\small \underline{\textbf{LoRA vs. full-parameter}} CPT of \BYOLC 4B. LoRA improves with rank but remains below full-param tuning (51.82).}
\label{fig:lora}
\end{minipage}
\end{figure*}

\vspace{0.4em}\noindent\textbf{Effect of data mixture.}
We evaluate four data mixtures for continual pretraining of the 4B \BYOLC model: C1 uses monolingual raw Chichewa data, C2 replaces it with our refined Chichewa corpus, C3 is a refined bilingual data mixture, and C4 further includes synthetic Chichewa obtained by translating refined English text.
Table~\ref{tab:pretrain_datasets} shows a steady improvement from C1 to C4. C4 provides the best results, increasing the Chichewa average score from 48.77 (C1) to 51.82, while preserving English performance. Per-dataset ablation results are provided in Table~\ref{tab:pre-train-ablation-sets-full}.

\vspace{0.4em}\noindent\textbf{LoRA vs. full-parameter finetuning.}
We assess the impact on the performance of the 4B \BYOLC model when updating either all parameters or only low-rank adapters. Figure~\ref{fig:lora} shows that LoRA~\cite{hu2022lora} performance improves with increasing rank, from 43.59 (r=64) to 49.60 (r=512), but remains below full-parameter tuning, which reaches 51.82. Based on these results, we adopt full-parameter training for all our models.

\begin{figure}[!t]
\centering
\includegraphics[width=\linewidth]{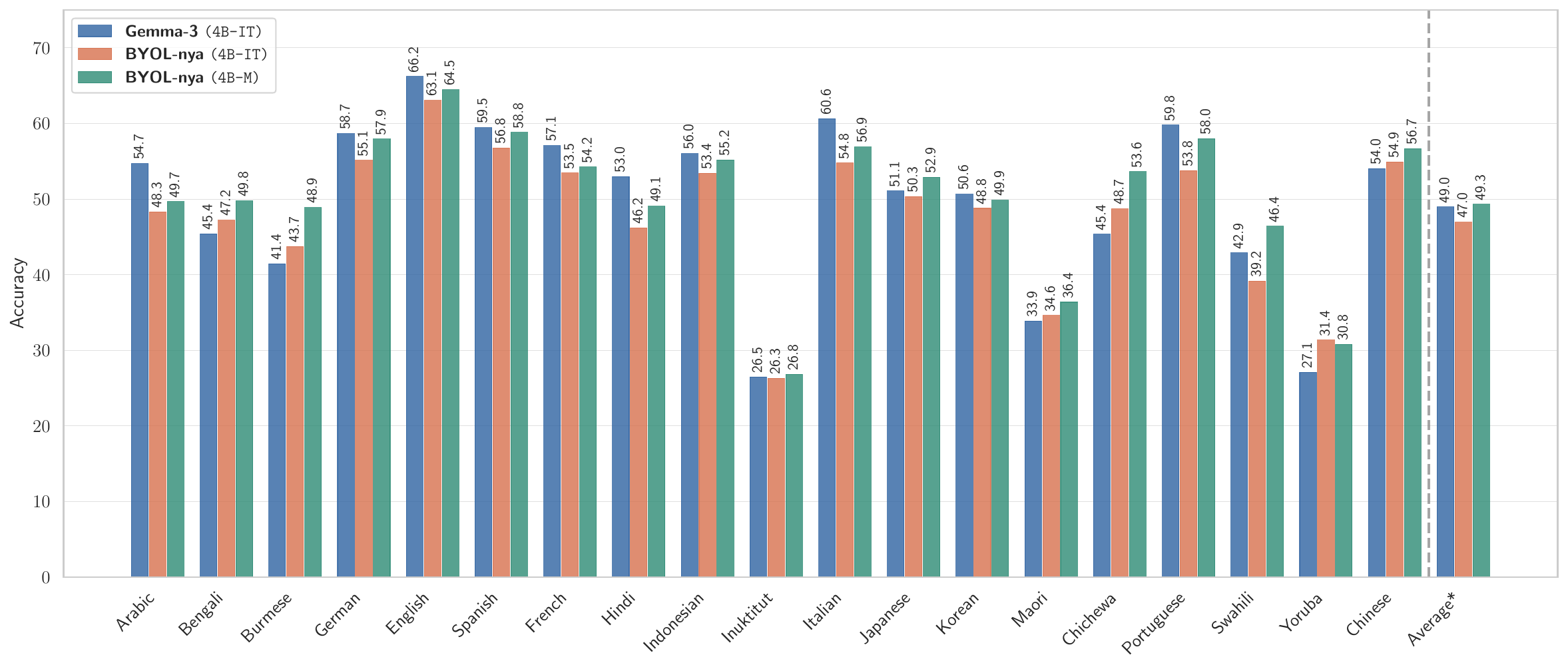}
    \vspace{-2em}
\caption{\small\underline{\textbf{Impact of model merging on multilingual performance}}. Evaluation on \GLOBALMMLU dataset~\cite{singh2024global}. Average$^\star$ accuracy excludes English and Chichewa to measure multilingual retention.}
\label{fig:model-merge}
\end{figure}

\vspace{0.4em}\noindent\textbf{Impact of model merging.}
We examine the impact of model merging on multilingual performance by comparing the 
(1) generalist multilingual model \GEMMA~(IT),
(2) language-specialist model \BYOLC (IT), and
(3) merged model \BYOLC (M) obtained using Eq.~\eqref{eq:model_merging}.
Figure~\ref{fig:model-merge} shows that the unmerged  \BYOLC (4B-IT) model yields strong gains on Chichewa but degrades performance on many other languages. In contrast, the merged model \BYOLC (4B-M) restores the multilingual capability of the \GEMMA~(IT) model across nearly all languages while retaining the improvements on Chichewa. Overall, model merging enables language-specific specialization without sacrificing multilinguality, showing that the procedure in Eq.~\eqref{eq:model_merging} effectively balances expert and generalist representations.

In addition to preserving multilingual accuracy, model merging also retains the safety characteristics of the generalist baseline. Table~\ref{tab:safety} reports bias and toxicity scores for the merged models, \BYOLC~(M) and \BYOLM~(M), compared with their unmerged IT variants, and  the baseline \GEMMA~(IT). We evaluate performance on three English benchmarks: BBQ~\cite{parrish-etal-2022-bbq} for social bias, and ToxiGen~\cite{hartvigsen2022toxigen} and RealToxicityPrompts~\cite{gehman-etal-2020-realtoxicityprompts} for toxicity. The merged models closely match baseline \GEMMA~(IT), whereas the unmerged IT models show the weakest safety performance. This indicates that merging maintains alignment and safety while adding low-resource language expertise, without any additional training or alignment steps.

\begin{table}[t]
\centering
\caption{\small \underline{\textbf{Effect of model merging on bias and toxicity.}} Merged models \BYOL (M) demonstrate safety characteristics much closer to the baseline than the unmerged models \BYOL(IT). }
\label{tab:safety}
\adjustbox{width=\textwidth,center}{
\begin{tabular}{l|c|cc|cc|c|cc|cc}
\toprule
& \multicolumn{5}{c|}{\bf 4B} & \multicolumn{5}{c}{\bf 12B} \\  \cmidrule(lr){2-6} 
        \cmidrule(lr){7-11} 
\bf Benchmarks 
& 
\rotatebox[origin=c]{70}{\small \makecell{\textbf{Gemma-3}~\cite{team2025gemma} \\ (IT) }} &
\rotatebox[origin=c]{70}{\small \makecell{\textbf{BYOL-nya} \\ (IT)}} &
\rotatebox[origin=c]{70}{\small \makecell{\textbf{BYOL-nya} \\ (M)}} &
\rotatebox[origin=c]{70}{\small \makecell{\textbf{BYOL-mri} \\ (IT)}} &
\rotatebox[origin=c]{70}{\small \makecell{\textbf{BYOL-mri} \\ (M)}} &
\rotatebox[origin=c]{70}{\small \makecell{\textbf{Gemma-3}~\cite{team2025gemma} \\ (IT) }} &
\rotatebox[origin=c]{70}{\small \makecell{\textbf{BYOL-nya} \\ (IT)}} &
\rotatebox[origin=c]{70}{\small \makecell{\textbf{BYOL-nya} \\ (M)}} &
\rotatebox[origin=c]{70}{\small \makecell{\textbf{BYOL-mri} \\ (IT)}} &
\rotatebox[origin=c]{70}{\small \makecell{\textbf{BYOL-mri} \\ (M)}} 
\\
\midrule[1.2pt]
BBQ~\cite{parrish-etal-2022-bbq} $\uparrow$ 
& 59.34 &  48.66 & 54.25 & 45.80 & 55.39 
&  67.43 &54.37 & 70.62 & 58.21 & 69.33 \\
ToxiGen~\cite{hartvigsen2022toxigen} $\uparrow$ 
& 81.49 &42.45 & 77.55 &43.30 & 79.47 
& 86.17 & 60.64 & 86.49 & 60.11 & 86.49 \\
RealToxicity Prompts~\cite{gehman-etal-2020-realtoxicityprompts} $\downarrow$ 
& \phantom{0}0.35 & \phantom{0}4.79
 & \phantom{0}1.44 & \phantom{0}5.87
 & \phantom{0}1.53 
&  \phantom{0}0.21 &\phantom{0}4.26 & \phantom{0}0.85 & \phantom{0}4.36
 & \phantom{0}0.85 \\
\bottomrule
\end{tabular}
}
\end{table}


\begin{figure}[!t]
    \centering

    \begin{subfigure}[b]{0.45\linewidth}
        \includegraphics[width=\linewidth]{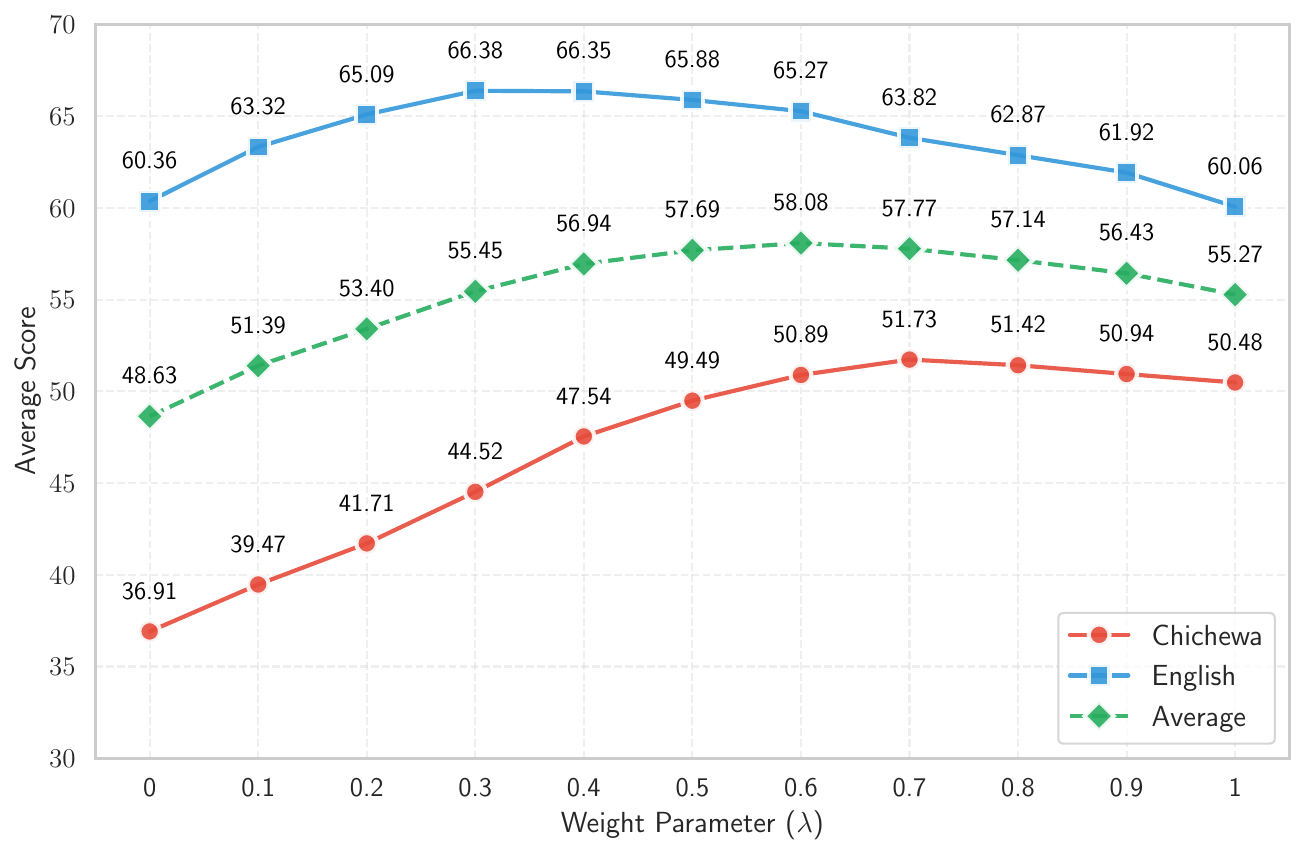}
    \end{subfigure}
    \hspace{0.02\linewidth}
    \begin{subfigure}[b]{0.45\linewidth}
        \includegraphics[width=\linewidth]{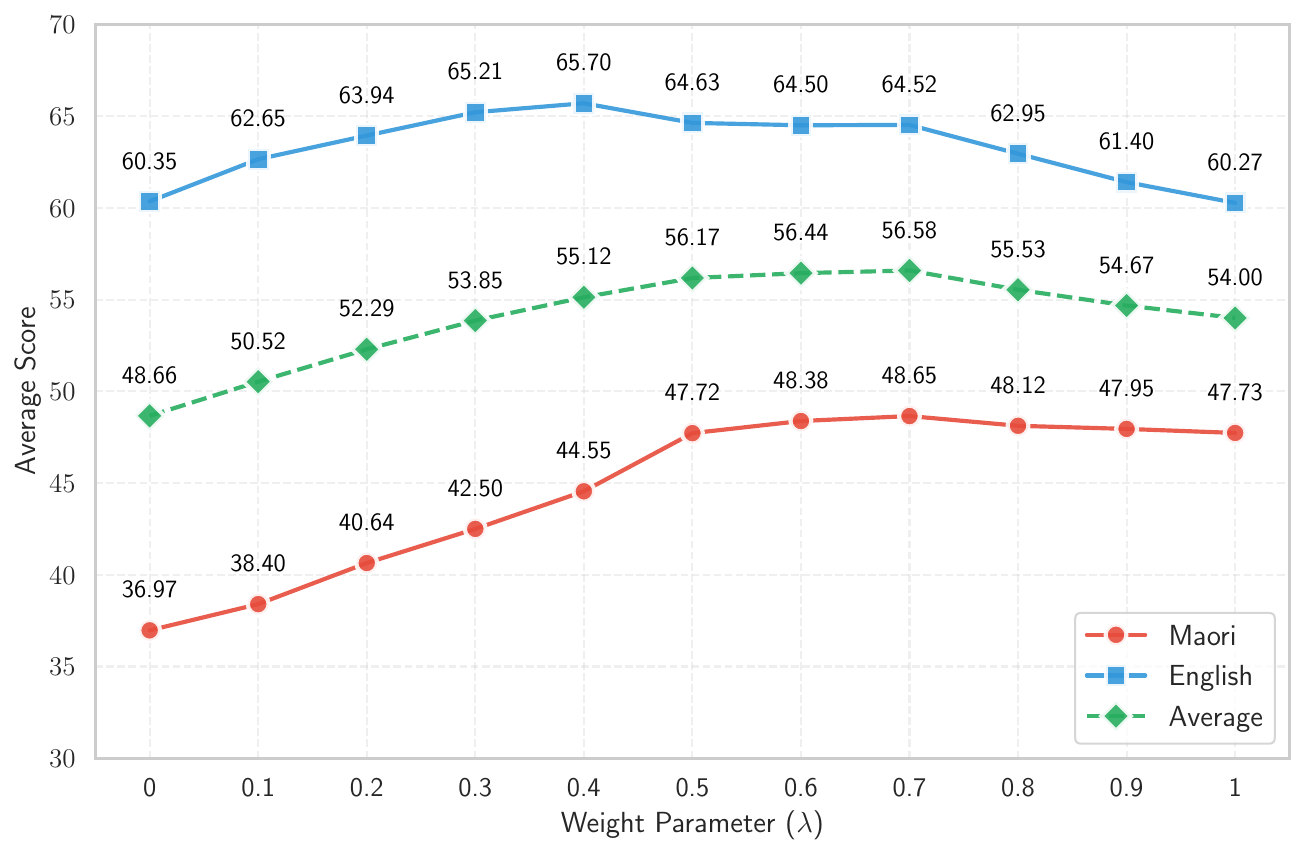}
    \end{subfigure}
    \vspace{-0.8em}
\caption{\small\underline{\textbf{Effect of model merging weight.}}  
\textbf{Left:} \BYOLC performance on Chichewa. \textbf{Right:} \BYOLM performance on Māori.  
Target-language accuracy increases with larger $\lambda$, while English performance declines; the best overall trade-off occurs near $\lambda=0.6$.}
    \label{fig:lambda-sweep}
\end{figure}

\vspace{0.4em}\noindent\textbf{Choice of model merging weights.}
Although Eq.~\eqref{eq:model_merging} uses two coefficients, in the ablation we examine a one-dimensional slice by enforcing $\alpha + \beta = 1$ and
reparameterizing $ \alpha = 1 - \lambda$ and $\beta = \lambda$, with $  \lambda \in [0,1].$

Figure~\ref{fig:lambda-sweep}(left) shows that increasing $\lambda$ steadily
improves Chichewa performance, rising from 36.91 at $\lambda=0$ (the pure
generalist baseline) to a peak of 51.73 at $\lambda=0.7$.  
English performance follows the opposite trend: it peaks at small $\lambda$
values (66.38 at $\lambda=0.3$) and gradually declines as the expert’s influence
grows.  
The bilingual average reaches its best overall score (58.08) at
$\lambda=0.6$, which we use for all merged Chichewa models.  
A similar pattern appears for Māori, as shown in
Fig.~\ref{fig:lambda-sweep}(right).

\subsection{Translation-Mediated LLM Access for Inuktitut}
\subsubsection{Experimental Details}
\noindent\textbf{Implementation details.} 
We train two Transformer models~\cite{transformers2017_vaswani}: one for Inuktitut to English translation and one for the reverse direction. Each model uses 9 encoder and 9 decoder layers with an embedding size of 512, 8 attention heads, and feed-forward size of 2048 dimensions. We use the BPE~\cite{sennrich2016neural_bpe} tokenizer with a shared vocabulary of 4096 tokens for both source and target text. 
We train models using the Adam optimizer~\cite{kingma2015adam} for 200K iterations with a batch size of 16K tokens. The learning rate follows the Noam schedule~\cite{transformers2017_vaswani} with a 10K-step warmup. For regularization, we apply dropout~\cite{srivastava2014dropout} and label smoothing~\cite{Szegedy2016_label}, both set to 0.1. After training, we average the last five checkpoints saved at 5K-step intervals~\cite{edunov-etal-2018-understanding}.

\vspace{0.4em}\noindent\textbf{Datasets.} 
For training the translation models, we use one publicly available dataset, the Nunavut Hansard (NH 3.0) corpus~\cite{joanis-etal-2020-nunavut}, and two internal datasets originating from children's books and news articles.
In addition to these human-translated datasets, we also include synthetic back-translated data collected from various sources~\cite{tiedemann-2012-parallel,schwenk-etal-2021-wikimatrix,nguyen-daume-iii-2019-global}. 
Details on MT datasets are summarized in Appendix~\ref{ref:mt-datasets}.
We convert Inuktitut text from syllabics to the romanized version~\cite{joanis-etal-2020-nunavut}. We apply several filtering operations to clean the data, including removing duplicate sentence pairs, discarding sentences shorter than 3 tokens or longer than 256 tokens, and filtering out pairs with a source–target character length ratio greater than 1.3. 
We apply on-the-fly data augmentation~\cite{post2023sotastreamstreamingapproachmachine} to improve generalization, including random punctuation removal, diacritic stripping, casing variation, and a copy mechanism that replaces the source sequence with the target sequence to enable identity mapping.
\subsubsection{Performance Evaluation}
\vspace{0.4em}\noindent\textbf{Evaluation of translators.}
We evaluate the translation accuracy of both LLMs and NMT models in the Inuktitut~$\leftrightarrow$~English setting. Results in Table~\ref{tab:nmt_results_iku2eng} and Table~\ref{tab:nmt_results_eng2iku} show that our NMT models achieve state-of-the-art performance. Averaged across all datasets, our method provides a 3.64 BLEU gain over Azure Translator when translating into English, and a 4.31 BLEU gain when translating out of English. Among LLMs, the reasoning model (GPT-5) performs better than GPT-4o and GPT-4.1, and 5-shot prompting provides noticeable improvements over zero-shot. However, even with few-shot inference, LLM accuracy remains far below that of dedicated NMT systems.

\begin{table}[!t]
    \centering
    \caption{\small \textbf{Inuktitut $\rightarrow$ English} translation results. BLEU and chrF++: higher is better.}
    \label{tab:nmt_results_iku2eng}
    \scalebox{0.7}{
    \begin{tabular}{c c c | c c | c c | c c | c c | c c}
        \toprule
        & & &
        \multicolumn{2}{c|}{\begin{tabular}{c}
            \textbf{NH 3.0}~\cite{joanis-etal-2020-nunavut}\\ \small{(Dev-Test)}
        \end{tabular}} &
        \multicolumn{2}{c|}{\begin{tabular}{c}
            \textbf{NH 3.0}~\cite{joanis-etal-2020-nunavut} \\ \small{(Test)}
        \end{tabular}} &
        \multicolumn{2}{c|}{\begin{tabular}{c}
            \textbf{News Articles} \\ \small{(Internal)}
        \end{tabular}} &
        \multicolumn{2}{c|}{\begin{tabular}{c}
            \textbf{Children Books} \\ \small{(Internal)}
        \end{tabular}} &
        \multicolumn{2}{c}{\textbf{Average}} \\

        \cmidrule(lr){4-5} 
        \cmidrule(lr){6-7} 
        \cmidrule(lr){8-9} 
        \cmidrule(lr){10-11}
        \cmidrule(lr){12-13}

        & & &
        {BLEU} & {chrF++} &
        {BLEU} & {chrF++} &
        {BLEU} & {chrF++} &
        {BLEU} & {chrF++} &
        {BLEU} & {chrF++} \\

        \midrule

        \multirow{4}{*}{\rotatebox[origin=c]{90}{\footnotesize \textbf{LLM}}}
        & \multirow{2}{*}{GPT-4o}
            & 0-shot
            & 9.59 & 28.45
            & 11.29 & 29.90
            & 9.08 & 32.24
            & 11.02 & 30.04
            & 10.25 & 30.16 \\
        &   & 5-shot
            & 13.27 & 31.77
            & 15.35 & 33.56
            & 12.82 & 34.49
            & 12.84 & 31.50
            & 13.57 & 32.83 \\
        \cmidrule{3-13}

        & \multirow{2}{*}{GPT-4.1}
            & 0-shot
            & 8.44 & 27.68
            & 5.59 & 21.69 
            & 7.39 & 30.88
            & 10.11 & 29.50
            & 7.88 & 27.94 \\
        &   & 5-shot
            & 12.14 & 30.76
            & 14.16 & 32.48 
            & 10.39 & 32.31
            & 12.73 & 31.26
            & 12.36 & 31.70 \\
        \cmidrule{3-13}

        & \multirow{2}{*}{GPT-5-Reasoning}
            & 0-shot
            & 12.80 & 33.10
            & 14.44 & 34.99
            & 9.88 & 34.47
            & 10.79 & 31.47
            & 12.00 & 33.51 \\
        &   & 5-shot
            & 16.33 & 35.75
            & 18.02 & 38.17 
            & 12.35 & 35.94
            & 12.50 & 32.81
            & 14.80 & 35.67 \\
        \midrule

        \multirow{2}{*}{\smash{\rotatebox[origin=c]{90}{\footnotesize \textbf{NMT}}}}  
        & Azure Translator
            & --
            & 31.31 & 49.29
            & 34.76 & 52.11
            & 28.01 & 49.80
            & 22.56 & 42.97
            & 29.16 & 48.54 \\
        & Ours
            & --
            & \textbf{35.73} & \textbf{52.44}
            & \textbf{40.24} & \textbf{55.89}
            & \textbf{28.49} & \textbf{50.06}
            & \textbf{26.72} & \textbf{46.48}
            & \textbf{32.80} & \textbf{51.22} \\
        \bottomrule
    \end{tabular}
    }
\end{table}

\begin{table}[!t]
    \centering
    \caption{\small \textbf{English $\rightarrow$ Inuktitut} translation results. BLEU and chrF++: higher is better.}
    \label{tab:nmt_results_eng2iku}
    \scalebox{0.7}{
    \begin{tabular}{c c c | c c | c c | c c | c c | c c}
        \toprule
        & & &
        \multicolumn{2}{c|}{\begin{tabular}{c}
            \textbf{NH 3.0~\cite{joanis-etal-2020-nunavut}}\\ \small{(Dev-Test)}
        \end{tabular}} &
        \multicolumn{2}{c|}{\begin{tabular}{c}
            \textbf{NH 3.0~\cite{joanis-etal-2020-nunavut}} \\ \small{(Test)}
        \end{tabular}} &
        \multicolumn{2}{c|}{\begin{tabular}{c}
            \textbf{News Articles} \\ \small{(Internal)}
        \end{tabular}} &
        \multicolumn{2}{c|}{\begin{tabular}{c}
            \textbf{Children Books} \\ \small{(Internal)}
        \end{tabular}} &
        \multicolumn{2}{c}{\textbf{Average}} \\
        \cmidrule(lr){4-5} 
        \cmidrule(lr){6-7} 
        \cmidrule(lr){8-9} 
        \cmidrule(lr){10-11}
        \cmidrule(lr){12-13}

        & & &
        BLEU & chrF++ &
        BLEU & chrF++ &
        BLEU & chrF++ &
        BLEU & chrF++ &
        BLEU & chrF++ \\

        \midrule

        \multirow{4}{*}{\rotatebox[origin=c]{90}{\footnotesize \textbf{LLM}}}
        & \multirow{2}{*}{GPT-4o}
            & 0-shot
            & 1.13 & 12.61
            & 1.28 & 12.01
            & 0.61 & 13.22
            & 0.50 & 13.19
            & 0.88 & 12.76 \\
        &   & 5-shot
            & 1.96 & 16.90
            & 2.56 & 17.46
            & 1.11 & 19.93
            & 1.21 & 16.13
            & 1.71 & 17.61 \\
        \cmidrule{3-13}

        & \multirow{2}{*}{GPT-4.1}
            & 0-shot
            & 1.20 & 14.10
            & 0.97 & 11.78
            & 0.84 & 15.68
            & 0.65 & 15.66
            & 0.92 & 14.31 \\
        &   & 5-shot
            & 2.63 & 20.69
            & 3.34 & 21.23
            & 1.68 & 23.34
            & 2.00 & 21.50
            & 2.41 & 21.69 \\
        \cmidrule{3-13}

        & \multirow{2}{*}{GPT-5-Reasoning}
            & 0-shot
            & 4.43 & 23.95
            & 5.68 & 24.42
            & 2.45 & 24.07
            & 1.91 & 25.58
            & 3.62 & 24.51 \\
        &   & 5-shot
            & 6.92 & 29.62
            & 6.61 & 28.31
            & 3.75 & 29.20
            & 4.91 & 29.69
            & 5.55 & 29.21 \\
        \midrule

        \multirow{2}{*}{\rotatebox[origin=c]{90}{\footnotesize \textbf{NMT}}}
        & Azure Translator
            & --
            & 15.14 & 43.76
            & 17.17 & 44.69
            & 6.89 & 42.69
            & 8.57 & 42.65
            & 11.94 & 43.45 \\
        & Ours
            & --
            & \textbf{18.82} & \textbf{44.59}
            & \textbf{21.08} & \textbf{46.34}
            & \textbf{12.27} & \textbf{46.57}
            & \textbf{14.82} & \textbf{45.29}
            & \textbf{16.25} & \textbf{45.70} \\
        \bottomrule
    \end{tabular}
    }
\end{table}

\begin{figure*}[!t]
\centering
\begin{minipage}{0.37\textwidth}
    \parbox[t]{\textwidth}{
        \caption{
            \textbf{\small {LLM accuracy on \GLOBALMMLU}} ~\cite{singh2024global}
            under three evaluation settings: (1) English text input,
            (2) direct Inuktitut text input, and (3) a translation-mediated LLM access (Inuktitut→English→LLM).
            Results show a large degradation when evaluating directly in Inuktitut and
            a notable accuracy recovery when our machine translator is used as an intermediate step.
        }
        \label{fig:translate-test-mmlu}
    }
\end{minipage}
\hfill
\begin{minipage}{0.62\textwidth}
    \parbox[t]{\textwidth}{
        \includegraphics[width=\textwidth]{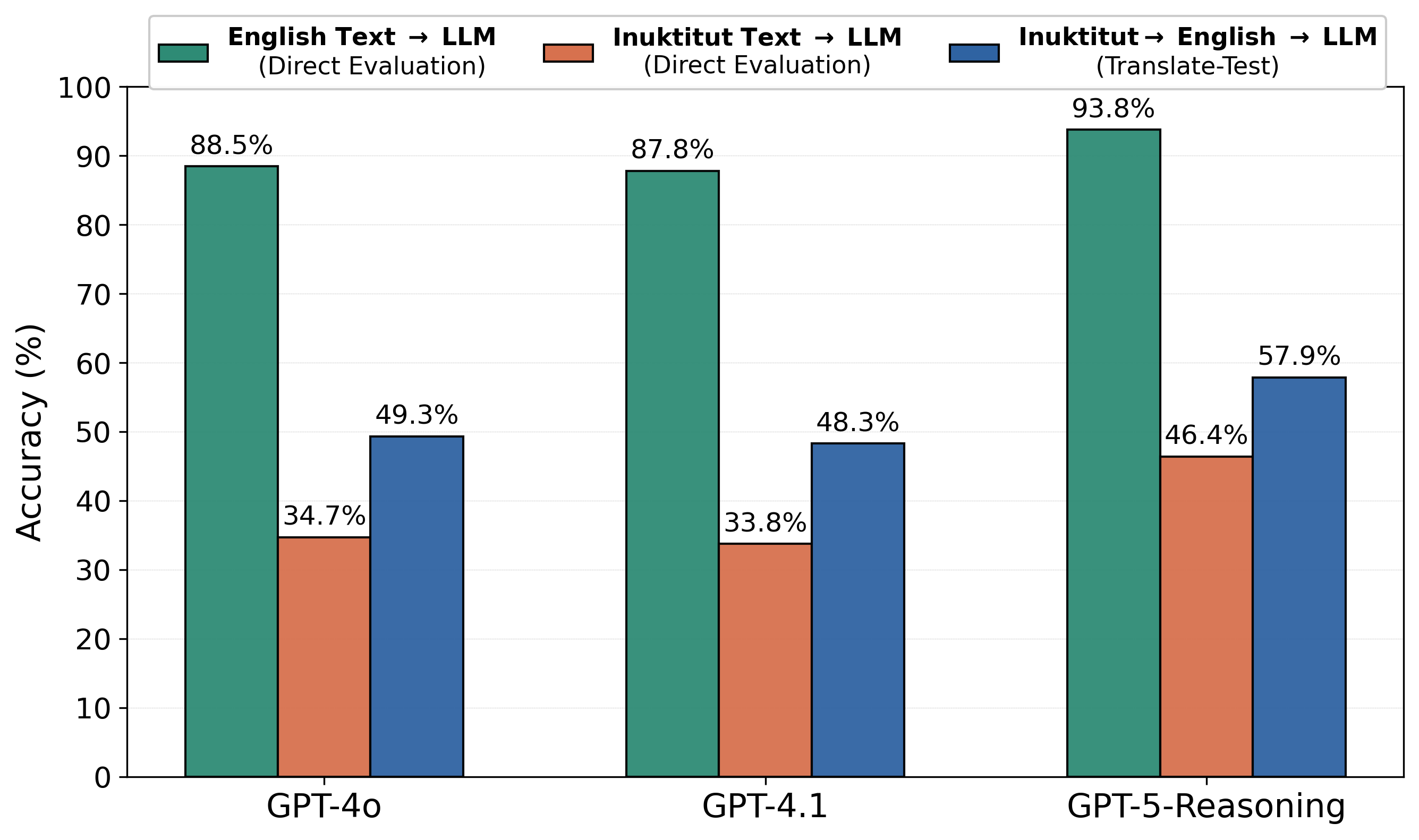}
    }
\end{minipage}
\end{figure*} 

\vspace{0.4em}\noindent\textbf{Evaluation of translation-mediated LLM access.}
We evaluate how LLMs perform on the \GLOBALMMLU benchmark~\cite{singh2024global} under three settings: (1) direct English text input, (2) direct Inuktitut text input, and (3) translation-mediated LLM access ((Inuktitut $\rightarrow$ English $\rightarrow$ LLM). Figure~\ref{fig:translate-test-mmlu} shows that while all LLMs are highly accurate in English, their performance drops drastically (over 50\%) when prompted directly in Inuktitut, indicating a weak understanding of the language. In contrast, inserting our MT model (Inuktitut→English) significantly recovers the lost accuracy in all LLMs. Specifically, compared to direct Inuktitut inference, translation-mediated access yields gains of 14.6\% for GPT-4o, 14.42\% for GPT-4.1, and 11.5\% for GPT-5 (reasoning).

\section{Challenges and Future Work}

\subsection{LLM Safety in Low-Resource Languages}
Model merging helps preserve English-aligned bias and toxicity behaviors in our adapted models, but it is unlikely to guarantee robust multilingual safety on its own. Recent open-weight work notes that safety alignment can be reversed by post-training, motivating stronger a priori data compliance during pretraining and broader, language-specific safety resources~\cite{hernandez2025apertus}. In addition, English-centric safety evaluation can lead to cross-lingual safety leaks, including translation-based jailbreaks~\cite{shen-etal-2024-language,ustun-etal-2024-aya}. Future work should expand low-resource safety datasets that separately cover harmful requests, bias, and toxicity, combining careful translation/adaptation of English resources with new human-curated, community-developed data that reflects local norms and realistic usage contexts. A complementary direction is to build specialized safety guard models for individual low-resource languages or small language families. Recent guardrails such as Qwen3Guard~\cite{zhao2025qwen3guard} suggest that language- or family-specific guards for \BYOL targets could provide more controllable safety than post-training alignment alone.

\subsection{Extension to Multilingual LLMs}
This work injected knowledge of a single low-resource language into multilingual 4B/12B models, showing strong target-language gains while largely preserving original capabilities. A natural next step is to move from single-language to multilingual specialization by integrating groups of related or typologically diverse languages~\cite{martins2025eurollm,dou2025sailor2}. This can be explored by training language-specific experts and merging them into a shared backbone, or by continual pretraining and instruction finetuning on a mixed multilingual dataset. Extending the same pipeline to larger backbones (e.g., \GEMMA 27B~\cite{team2025gemma}) may further improve performance and clarify how benefits scale with model size.

\subsection{Extension to Speech}
Many extreme-low-resource and LRLs are primarily spoken, making speech the most natural, and sometimes the only interface for accessing LLM-based tools. Future work should therefore extend \BYOL beyond text by integrating robust speech-to-text components (and eventually text-to-speech) as front-ends for language-specific adaptation. Recent progress such as Omnilingual Automatic Speech Recognition (ASR)~\cite{omnilingual2025omnilingual} suggests a promising direction with open, extensible multilingual ASR models that scale to 7B parameters and enable rapid extension to new languages. A key next step is to combine such ASR backbones with \BYOL-style text adaptation to build efficient end-to-end speech–LLM stacks for low-resource settings. Progress will likely depend on expanding high-quality speech--text data and improving robustness to real-world conditions such as noise, code-switching, and multi-speaker audio.

\subsection{Data Scarcity}
We extended \GLOBALMMLU~\cite{singh2024global} from 16 to 19 languages by adding high-quality, human-expert translations for Chichewa, Māori, and Inuktitut. In parallel, our partnership with the Government of Nunavut enabled the collection of high-quality, culturally grounded paired Inuktitut--English data. These contributions strengthen evaluation coverage and provide locally grounded bilingual resources, but translation alone is unlikely to reduce the broader bias of LLMs toward low-resource languages. More generally, data scarcity remains a central limitation: many LRLs still lack sufficient pretraining text, high-quality instruction and safety data, and robust language-specific evaluation sets~\cite{romanou2024include}. Future work should prioritize sustained, community-driven data creation across modalities, supported by strong data governance.

\section{Conclusion}
\label{sec:conclusion}
In this paper, we proposed \BYOL, an open framework for bringing LRLs into LLMs. \BYOL is guided by two key inputs: (i) an initial assessment of existing AI tools to identify the most suitable open-weight LLMs and MT systems for a target language, and (ii) a language digital-resource classification that assigns each language to one of four levels, {Extreme-Low}, {Low}, {Mid}, and {High}, based on the volume of available web-scale text. The classification determines the integration strategy: direct instruction finetuning for mid- and high-resource languages, additional continual pretraining for low-resource languages, and translation-based inclusion for extreme-low-resource settings. The best tools identified through the initial assessment were then used to support data curation and to serve as baseline models for training within the selected pathway. Using \BYOL, we instantiated the {Low}-resource path for Chichewa and Māori and trained two families of language-centric models, yielding four models—\BYOLC (M) and \BYOLM (M) at 4B and 12B parameters. Across 12 benchmarks, these models achieved an average improvement of around 12\% over a strong multilingual baseline. Under LLM-as-a-judge evaluation, \BYOLC (12B-M) and \BYOLM (12B-M) performed on par with GPT-4o on a question-answering benchmark, establishing a new state of the art among open models for Chichewa and Māori. For languages with extremely limited digital presence, we further explored translation-mediated inclusion on Inuktitut by training a neural machine translation system that yielded approximately +4 BLEU over a commercial baseline across three datasets, and showed that translation-mediated LLM use yields a $\sim$14\% accuracy gain over direct inference. Finally, we released human-translated versions of \GLOBALMMLU in Chichewa, Māori, and Inuktitut, improving the reliability of multiple-choice LLM evaluation for these languages. Looking ahead, we hope \BYOL will provide the NLP community with a practical, open recipe for extending LLM support to additional underrepresented languages and for releasing the data, models, and benchmarks needed to advance inclusive multilingual AI.

\section{Acknowledgments}
We would like to thank the Government of Nunavut, Canada, for their support and guidance on the Inuktitut language. Their collaboration enabled the human-expert translation of \GLOBALMMLU into Inuktitut and the collection of new paired English--Inuktitut data.

\newpage 
\bibliographystyle{ieeetr}
\bibliography{mainbib}

@article{gupta2020stochasticweightaveragingparallel,
  title   = {Stochastic Weight Averaging in Parallel: Large-Batch Training that Generalizes Well},
  author  = {Gupta, Vipul and Akle Serrano, Santiago and DeCoste, Dennis},
  journal = {arXiv:2001.02312},
  year    = {2020},
  url     = {https://arxiv.org/abs/2001.02312}
}

@inproceedings{Tao_2024,
   title={Unlocking the Potential of Model Merging for Low-Resource Languages},
   url={http://dx.doi.org/10.18653/v1/2024.findings-emnlp.508},
   DOI={10.18653/v1/2024.findings-emnlp.508},
   booktitle={Findings of the Association for Computational Linguistics: EMNLP 2024},
   author={Tao, Mingxu and Zhang, Chen and Huang, Quzhe and Ma, Tianyao and Huang, Songfang and Zhao, Dongyan and Feng, Yansong},
   year={2024}
   }

@article{hammoud2024modelmergingsafetyalignment,
  title   = {Model Merging and Safety Alignment: One Bad Model Spoils the Bunch},
  author  = {Hammoud, Hasan Abed Al Kader and Michieli, Umberto and Pizzati, Fabio and Torr, Philip and Bibi, Adel and Ghanem, Bernard and Ozay, Mete},
  journal = {arXiv:2406.14563},
  year    = {2024},
  url     = {https://arxiv.org/abs/2406.14563}
}

@inproceedings{sennrich2016neural_bpe,
  title={Neural machine translation of rare words with subword units},
  author={Sennrich, Rico and Haddow, Barry and Birch, Alexandra},
  booktitle={Proceedings of the 54th annual meeting of the association for computational linguistics},
  year={2016}
}

@article{misra2025measuringaidiffusionpopulationnormalized,
      title={Measuring {AI} Diffusion: A Population-Normalized Metric for Tracking Global AI Usage}, 
      author={Amit Misra and Jane Wang and Scott McCullers and Kevin White and Juan Lavista Ferres},
      year={2025},
      journal={arXiv:2511.02781},
      url={https://arxiv.org/abs/2511.02781}, 
}

@article{misra2025aidiffusionlowresource,
  title   = {{AI} Diffusion in Low Resource Language Countries},
  author  = {Misra, Amit and Zamir, Syed Waqas and Hamidouche, Wassim and Becker-Reshef, Inbal and Lavista Ferres, Juan},
  journal = {arXiv:2511.02752},
  year    = {2025},
  url     = {https://arxiv.org/abs/2511.02752}
}

@inproceedings{
hu2022lora,
title={Lo{RA}: Low-Rank Adaptation of Large Language Models},
author={Edward J Hu and Yelong Shen and Phillip Wallis and Zeyuan Allen-Zhu and Yuanzhi Li and Shean Wang and Lu Wang and Weizhu Chen},
booktitle={International Conference on Learning Representations},
year={2022},
url={https://openreview.net/forum?id=nZeVKeeFYf9}
}

@article{eval-harness,
  title    = {The Language Model Evaluation Harness},
  author   = {Gao, Leo and Tow, Jonathan and Abbasi, Baber and Biderman, Stella and Black, Sid and DiPofi, Anthony and Foster, Charles and Golding, Laurence and Hsu, Jeffrey and Le Noac'h, Alain and Li, Haonan and McDonell, Kyle and Muennighoff, Niklas and Ociepa, Chris and Phang, Jason and Reynolds, Laria and Schoelkopf, Hailey and Skowron, Aviya and Sutawika, Lintang and Tang, Eric and Thite, Anish and Wang, Ben and Wang, Kevin and Zou, Andy},
  journal  = {Zenodo},
  year     = {2024},
  doi      = {10.5281/zenodo.12608602},
  url      = {https://zenodo.org/records/12608602}
}

@misc{simplewiki,
  title        = {Simple English Wikipedia},
  author       = {Simple English Wikipedia Contributors},
  howpublished = {\url{https://simple.wikipedia.org/}}
}

@misc{tatoeba,
  title        = {The Tatoeba Project},
  author       = {Tatoeba Community},
  howpublished = {\url{https://tatoeba.org/}},
  year         = {2020}
}

@inproceedings{xue2021mt5,
  title     = {{mT5:} A Massively Multilingual Pre-trained Text-to-Text Transformer},
  author    = {Xue, Linting and Constant, Noah and Roberts, Adam and Kale, Mihir and Al-Rfou, Rami and Siddhant, Aditya and Barua, Aditya and Raffel, Colin},
  booktitle = {Proceedings of NAACL},
  year      = {2021}
}

@article{aryabumi2024aya,
  title={Aya 23: Open weight releases to further multilingual progress},
  author={Aryabumi, Viraat and Dang, John and Talupuru, Dwarak and Dash, Saurabh and Cairuz, David and Lin, Hangyu and Venkitesh, Bharat and Smith, Madeline and Campos, Jon Ander and Tan, Yi Chern and others},
  journal={arXiv:2405.15032},
  year={2024}
}

@article{martins2025eurollm,
  title={{EuroLLM}: Multilingual language models for europe},
  author={Martins, Pedro Henrique and Fernandes, Patrick and Alves, Jo{\~a}o and Guerreiro, Nuno M and Rei, Ricardo and Alves, Duarte M and Pombal, Jos{\'e} and Farajian, Amin and Faysse, Manuel and Klimaszewski, Mateusz and others},
  journal={Procedia Computer Science},
  year={2025},
}

@inproceedings{burchell2025expanded,
  title={An expanded massive multilingual dataset for high-performance language technologies ({HPLT})},
  author={Burchell, Laurie and Bonet, Ona De Gibert and Arefyev, Nikolay and Aulamo, Mikko and Ba{\~n}{\'o}n, Marta and Chen, Pinzhen and Fedorova, Mariia and Guillou, Liane and Haddow, Barry and Hajic, Jan and others},
  booktitle={Proceedings of the 63rd Annual Meeting of the Association for Computational Linguistics},
  year={2025}
}

@article{sengupta2023jais,
  title   = {Jais and Jais-chat: Arabic-Centric Foundation and Instruction-Tuned Open Generative Large Language Models},
  author  = {Sengupta, Neha and Sahu, Sunil Kumar and Jia, Bokang and Katipomu, Satheesh and Li, Haonan and Koto, Fajri and Marshall, William and Gosal, Gurpreet and Liu, Cynthia and Chen, Zhiming and others},
  journal = {arXiv:2308.16149},
  year    = {2023},
  url     = {https://arxiv.org/abs/2308.16149}
}

@article{nanda2025,
  title   = {{Llama-3-Nanda-10B-Chat}: An Open Generative Large Language Model for Hindi},
  author  = {Choudhury, Monojit and Chauhan, Shivam and Das, Rocktim Jyoti and Sahnan, Dhruv and Han, Xudong and others},
  journal = {arXiv:2504.06011},
  year    = {2025},
  url     = {https://arxiv.org/abs/2504.06011}
}

@article{gouvert2025lucie,
  title   = {The Lucie-7B LLM and the Lucie Training Dataset: Open resources for multilingual language generation},
  author  = {Gouvert, Olivier and Hunter, Julie and Louradour, Jérôme and Cerisara, Christophe and Dufraisse, Evan and Sy, Yaya and Rivière, Laura and Lorré, Jean-Pierre and OpenLLM-France community},
  journal = {arXiv:2503.12294},
  year    = {2025},
  url     = {https://arxiv.org/abs/2503.12294}
}

@article{croissantllm2024,
  title   = {{CroissantLLM:} A Truly Bilingual French-English Language Model},
  author  = {Faysse, Manuel and Fernandes, Patrick and Guerreiro, Nuno M. and Loison, António and Alves, Duarte M. and Corro, Caio and Boizard, Nicolas and Alves, João and Rei, Ricardo and Martins, Pedro H. and others},
  journal = {arXiv:2402.00786},
  year    = {2024},
  url     = {https://arxiv.org/abs/2402.00786}
}

@article{poro2024,
  title   = {Poro 34B and the Blessing of Multilinguality},
  author  = {Luukkonen, Risto and Burdge, Jonathan and Zosa, Elaine and Talman, Aarne and Komulainen, Ville and Hatanpää, Väinö and Sarlin, Peter and Pyysalo, Sampo},
  journal = {arXiv:2404.01856},
  year    = {2024},
  url     = {https://arxiv.org/abs/2404.01856}
}

@misc{aguila2023,
  title        = {{\`A}guila-7B: An Open-Source LLM for Catalan and Spanish},
  author       = {{Barcelona Supercomputing Center (Projecte AINA)}},
  year         = {2023},
  howpublished = {Project technical release / model documentation},
  url          = {https://projecteaina.cat/tech/en/aguila7b-the-new-open-source-llm-for-catalan-and-spanish-developed-by-the-bsc/}
}

@misc{ulizallama2023,
  title        = {{UlizaLlama:} An Open-Access Swahili Large Language Model},
  author       = {{Jacaranda Health}},
  year         = {2023},
  howpublished = {Model release / documentation},
  url          = {https://huggingface.co/Jacaranda/UlizaLlama}
}

@inproceedings{shen-etal-2024-language,
    title = "The Language Barrier: Dissecting Safety Challenges of {LLM}s in Multilingual Contexts",
    author = "Shen, Lingfeng  and
      Tan, Weiting  and
      Chen, Sihao  and
      Chen, Yunmo  and
      Zhang, Jingyu  and
      Xu, Haoran  and
      Zheng, Boyuan  and
      Koehn, Philipp  and
      Khashabi, Daniel",
    booktitle = "Findings of the Association for Computational Linguistics: ACL 2024",
    year = "2024",
    url = "https://aclanthology.org/2024.findings-acl.156/",
    doi = "10.18653/v1/2024.findings-acl.156"
}

@article{omnilingual2025omnilingual,
  title={{Omnilingual ASR:} Open-Source Multilingual Speech Recognition for 1600+ Languages},
  author={Omnilingual, ASR and Keren, Gil and Kozhevnikov, Artyom and Meng, Yen and Ropers, Christophe and Setzler, Matthew and Wang, Skyler and Adebara, Ife and Auli, Michael and Balioglu, Can and others},
  journal={arXiv:2511.09690},
  year={2025}
}

@article{romanou2024include,
  title={Include: Evaluating multilingual language understanding with regional knowledge},
  author={Romanou, Angelika and Foroutan, Negar and Sotnikova, Anna and Chen, Zeming and Nelaturu, Sree Harsha and Singh, Shivalika and Maheshwary, Rishabh and Altomare, Micol and Haggag, Mohamed A and Amayuelas, Alfonso and others},
  journal={arXiv:2411.19799},
  year={2024}
}

@article{ali2024teuken,
  title={Teuken-7b-base \& teuken-7b-instruct: Towards european llms},
  author={Ali, Mehdi and Fromm, Michael and Thellmann, Klaudia and Ebert, Jan and Weber, Alexander Arno and Rutmann, Richard and Jain, Charvi and L{\"u}bbering, Max and Steinigen, Daniel and Leveling, Johannes and others},
  journal={arXiv:2410.03730},
  year={2024}
}

@inproceedings{zhang2015character,
  title        = {Character-level Convolutional Networks for Text Classification},
  author       = {Zhang, Xiang and Zhao, Junbo and LeCun, Yann},
  booktitle    = {Advances in Neural Information Processing Systems},
  year         = {2015}
}

@inproceedings{narayan2018don,
  title        = {Don't Give Me the Details, Just the Summary! Topic-Aware Convolutional Neural Networks for Extreme Summarization},
  author       = {Narayan, Shashi and Cohen, Shay B. and Lapata, Mirella},
  booktitle    = {Proceedings of the 2018 Conference on Empirical Methods in Natural Language Processing},
  year         = {2018}
}

@inproceedings{koupaee2018wikihow,
  title        = {{WikiHow}: A Large Scale Text Summarization Dataset},
  author       = {Koupaee, Mahnaz and Wang, William Yang},
  booktitle    = {Proceedings of the 2018 EMNLP Workshop on Analysis of Abusive Language},
  year         = {2018}
}

@inproceedings{khashabi2021gooaq,
  title        = {{GooAQ:} Open Question Answering from Questions Asked to a Large Language Model},
  author       = {
    Khashabi, Daniel and Kordi, Yao and Khot, Tushar and Sabharwal, Ashutosh
    and Tafjord, Oyvind and Clark, Peter and Hajishirzi, Hannaneh
  },
  booktitle    = {Proceedings of the 2021 Conference on Empirical Methods in Natural Language Processing},
  year         = {2021}
}

@inproceedings{kwiatkowski2019natural,
  title        = {{Natural Questions:} A Benchmark for Question Answering Research},
  author       = {
    Kwiatkowski, Tom and Palomaki, Juliusz and Redfield, Olivia and Collins, Michael
    and Parikh, Ankur P. and Alberti, Chris and Epstein, Danielle and Polosukhin, Ilya
    and Kelcey, Jacob and Devlin, Jacob and Lee, Kenton and Toutanova, Kristina
    and Jones, Llion and Dai, Andrew M. and Uszkoreit, Jakob and Le, Quoc V.
    and Petrov, Slav
  },
  booktitle    = {Transactions of the Association for Computational Linguistics},
  year         = {2019}
}

@inproceedings{fan2019eli5,
  title        = {{ELI5}: Long Form Question Answering},
  author       = {Fan, Angela and Jernite, Yacine and Weston, Jason},
  booktitle    = {Proceedings of the 57th Annual Meeting of the Association for Computational Linguistics},
  year         = {2019}
}

@inproceedings{rajpurkar2016squad,
  title        = {{SQuAD:} 100,000+ Questions for Machine Comprehension of Text},
  author       = {Rajpurkar, Pranav and Zhang, Jian and Lopyrev, Konstantin and Liang, Percy},
  booktitle    = {Proceedings of the 2016 Conference on Empirical Methods in Natural Language Processing},
  year         = {2016}
}

@inproceedings{cettolo2012ted,
  title        = {{WIT3:} Web Inventory of Transcribed and Translated Talks},
  author       = {Cettolo, Mauro and Girardi, Christian and Federico, Marcello},
  booktitle    = {Proceedings of the 16th Annual Conference of the European Association for Machine Translation},
  year         = {2012}
}

@article{post2023sotastreamstreamingapproachmachine,
  title   = {{SOTASTREAM}: A Streaming Approach to Machine Translation Training},
  author  = {Post, Matt and Gowda, Thamme and Grundkiewicz, Roman and Khayrallah, Huda and Jain, Rohit and Junczys-Dowmunt, Marcin},
  journal = {arXiv:2308.07489},
  year    = {2023},
  url     = {https://arxiv.org/abs/2308.07489}
}

@article{srivastava2014dropout,
  title={Dropout: A simple way to prevent neural networks from overfitting},
  author={Srivastava, Nitish and Hinton, Geoffrey E. and Krizhevsky, Alex and Sutskever, Ilya and Salakhutdinov, Ruslan},
  journal={Journal of Machine Learning Research},
  year={2014}
}

@INPROCEEDINGS{Szegedy2016_label,
  author={Szegedy, Christian and Vanhoucke, Vincent and Ioffe, Sergey and Shlens, Jon and Wojna, Zbigniew},
  booktitle={IEEE Conference on Computer Vision and Pattern Recognition (CVPR)}, 
  title={Rethinking the Inception Architecture for Computer Vision}, 
  year={2016},
  }

@inproceedings{kingma2015adam,
  title={Adam: A method for stochastic optimization},
  author={Kingma, Diederik P. and Ba, Jimmy},
  booktitle={International Conference on Learning Representations},
  year={2015}
}

@techreport{bapna-etal-2022-next-thousand,
  author    = {Bapna, Ankur and Caswell, Isaac and Kreutzer, Julia and Firat, Orhan and van Esch, Daan and Siddhant, Aditya and Niu, Mengmeng and Baljekar, Pallavi Nikhil and Garcia, Xavier and Macherey, Wolfgang and Breiner, Theresa and Axelrod, Vera Saldinger and Riesa, Jason and Cao, Yuan and Chen, Mia and Macherey, Klaus and Krikun, Maxim and Wang, Pidong and Gutkin, Alexander and Shah, Apu and Huang, Yanping and Chen, Zhifeng and Wu, Yonghui and Hughes, Macduff Richard},
  title     = {Building Machine Translation Systems for the Next Thousand Languages},
  institution = {Google Research},
  year      = {2022},
  type      = {Technical Report},
  url       = {https://arxiv.org/abs/2212.10481},
}

@inproceedings{nielsen-etal-2025-alligators,
    title = "Alligators All Around: Mitigating Lexical Confusion in Low-resource Machine Translation",
    author = "Nielsen, Elizabeth  and
      Caswell, Isaac Rayburn  and
      Luo, Jiaming  and
      Cherry, Colin",
    booktitle = "Proceedings of the 2025 Conference of the Nations of the Americas Chapter of the Association for Computational Linguistics: Human Language Technologies (Volume 2: Short Papers)",
    year = "2025"
}

@inproceedings{tiedemann-2012-parallel,
  author    = {Tiedemann, J{\"o}rg},
  title     = {Parallel Data, Tools and Interfaces in OPUS},
  booktitle = {Proceedings of the 8th International Conference on Language Resources and Evaluation (LREC'12)},
  year      = {2012},
}

@inproceedings{nguyen-daume-iii-2019-global,
    title = "{G}lobal {V}oices: Crossing Borders in Automatic News Summarization",
    author = "Nguyen, Khanh  and
      Daum{\'e} III, Hal",
    booktitle = "Proceedings of the 2nd Workshop on New Frontiers in Summarization",
    year = "2019",
}

@inproceedings{schwenk-etal-2021-wikimatrix,
    title = "{W}iki{M}atrix: Mining 135{M} Parallel Sentences in 1620 Language Pairs from {W}ikipedia",
    author = "Schwenk, Holger  and
      Chaudhary, Vishrav  and
      Sun, Shuo  and
      Gong, Hongyu  and
      Guzm{\'a}n, Francisco",
    booktitle = "Proceedings of the 16th Conference of the European Chapter of the Association for Computational Linguistics: Main Volume",
    year = "2021",
}

@inproceedings{haque-etal-2020-terminology,
  author    = {Haque, Rejwanul and Moslem, Yasmin and Way, Andy},
  title     = {Terminology-Aware Sentence Mining for NMT Domain Adaptation: ADAPT’s Submission to the Adap-MT 2020 English-to-Hindi AI Translation Shared Task},
  booktitle = {Proceedings of the 17th International Conference on Natural Language Processing (ICON): Adap-MT 2020 Shared Task},
  year      = {2020}
}

@inproceedings{edunov-etal-2018-understanding,
  author    = {Edunov, Sergey and Ott, Myle and Auli, Michael and Grangier, David},
  title     = {Understanding Back-Translation at Scale},
  booktitle = {Proceedings of the 2018 Conference on Empirical Methods in Natural Language Processing},
  year      = {2018},
  url       = {https://aclanthology.org/D18-1045},
}

@inproceedings{sennrich-etal-2016-improving,
    title = "Improving Neural Machine Translation Models with Monolingual Data",
    author = "Sennrich, Rico  and
      Haddow, Barry  and
      Birch, Alexandra",
    booktitle = "Proceedings of the 54th Annual Meeting of the Association for Computational Linguistics)",
    year = "2016"
}

@inproceedings{moore-2002-fast,
  author    = {Moore, Robert C.},
  title     = {Fast and Accurate Sentence Alignment of Bilingual Corpora},
  booktitle = {Proceedings of the 5th Conference of the Association for Machine Translation in the Americas (AMTA 2002)},
  year      = {2002}
}

@article{gale-church-1993,
  author    = {William A. Gale and Kenneth W. Church},
  title     = {A Program for Aligning Sentences in Bilingual Corpora},
  journal   = {Computational Linguistics},
  year      = {1993},
}

@inproceedings{thompson-koehn-2019-vecalign,
    title = "{V}ecalign: Improved Sentence Alignment in Linear Time and Space",
    author = "Thompson, Brian  and
      Koehn, Philipp",
    booktitle = "Proceedings of the 2019 Conference on Empirical Methods in Natural Language Processing and the 9th International Joint Conference on Natural Language Processing (EMNLP-IJCNLP)",
    year = "2019"
}

@inproceedings{feng-etal-2022-language,
    title = "Language-agnostic {BERT} Sentence Embedding",
    author = "Feng, Fangxiaoyu  and
      Yang, Yinfei  and
      Cer, Daniel  and
      Arivazhagan, Naveen  and
      Wang, Wei",
    booktitle = "Proceedings of the 60th Annual Meeting of the Association for Computational Linguistics (Volume 1: Long Papers)",
    year = "2022"
}

@inproceedings{heffernan-etal-2022-bitext,
    title = "Bitext Mining Using Distilled Sentence Representations for Low-Resource Languages",
    author = "Heffernan, Kevin  and
      {\c{C}}elebi, Onur  and
      Schwenk, Holger",
    booktitle = "Findings of the Association for Computational Linguistics: EMNLP 2022",
    year = "2022",
}

@inproceedings{joanis-etal-2020-nunavut,
    title = "The {N}unavut {H}ansard {I}nuktitut{--}{E}nglish Parallel Corpus 3.0 with Preliminary Machine Translation Results",
    author = "Joanis, Eric  and
      Knowles, Rebecca  and
      Kuhn, Roland  and
      Larkin, Samuel  and
      Littell, Patrick  and
      Lo, Chi-kiu  and
      Stewart, Darlene  and
      Micher, Jeffrey",
    booktitle = "Proceedings of the Twelfth Language Resources and Evaluation Conference",
    year = "2020",
}

@inproceedings{transformers2017_vaswani,
 author = {Vaswani, Ashish and Shazeer, Noam and Parmar, Niki and Uszkoreit, Jakob and Jones, Llion and Gomez, Aidan N and Kaiser, \L ukasz and Polosukhin, Illia},
 booktitle = {Advances in Neural Information Processing Systems},
 title = {Attention is All you Need},
 year = {2017}
}

@InProceedings{pmlr-v162-wortsman22a,
  title = 	 {Model soups: averaging weights of multiple fine-tuned models improves accuracy without increasing inference time},
  author =       {Wortsman, Mitchell and Ilharco, Gabriel and Gadre, Samir Ya and Roelofs, Rebecca and Gontijo-Lopes, Raphael and Morcos, Ari S and Namkoong, Hongseok and Farhadi, Ali and Carmon, Yair and Kornblith, Simon and Schmidt, Ludwig},
  booktitle = 	 {Proceedings of the 39th International Conference on Machine Learning},
  year = 	 {2022},
}

@inproceedings{yu2024language,
  title={Language Models are Super Mario: Absorbing Abilities from Homologous Models as a Free Lunch},
  author={Yu, Le and Yu, Bowen and Yu, Haiyang and Huang, Fei and Li, Yongbin},
  booktitle={International Conference on Machine Learning},
  year={2024},
}

@misc{commoncrawl,
  author = {{Common Crawl}},
  howpublished = {\url{https://commoncrawl.org}},
  note = {Accessed: 2025-10-21}
}

@inproceedings{artetxe2023revisitingmachinetranslationcrosslingual,
    title = "Revisiting Machine Translation for Cross-lingual Classification",
    author = "Artetxe, Mikel  and
      Goswami, Vedanuj  and
      Bhosale, Shruti  and
      Fan, Angela  and
      Zettlemoyer, Luke",
    booktitle = "Proceedings of the 2023 Conference on Empirical Methods in Natural Language Processing",
    year = "2023",
    url = "https://aclanthology.org/2023.emnlp-main.399/",
    doi = "10.18653/v1/2023.emnlp-main.399"
}

@inproceedings{koretaka-etal-2023-mitigating,
    title = "Mitigating Domain Mismatch in Machine Translation via Paraphrasing",
    author = "Koretaka, Hyuga  and
      Kajiwara, Tomoyuki  and
      Fujita, Atsushi  and
      Ninomiya, Takashi",
    booktitle = "Proceedings of the 10th Workshop on Asian Translation",
    year = "2023",
    url = "https://aclanthology.org/2023.wat-1.2/"
}

@inproceedings{saunders-deneefe-2024-domain,
    title = "Domain adapted machine translation: What does catastrophic forgetting forget and why?",
    author = "Saunders, Danielle  and
      DeNeefe, Steve",
    booktitle = "Proceedings of the 2024 Conference on Empirical Methods in Natural Language Processing",
    year = "2024",
    url = "https://aclanthology.org/2024.emnlp-main.704/",
    doi = "10.18653/v1/2024.emnlp-main.704"
}

@inproceedings{shen2020sourcetargetdomainmismatchproblem,
    title = "The Source-Target Domain Mismatch Problem in Machine Translation",
    author = "Shen, Jiajun  and
      Chen, Peng-Jen  and
      Le, Matthew  and
      He, Junxian  and
      Gu, Jiatao  and
      Ott, Myle  and
      Auli, Michael  and
      Ranzato, Marc{'}Aurelio",
    booktitle = "Proceedings of the 16th Conference of the European Chapter of the Association for Computational Linguistics",
    year = "2021"
}

@article{saunders2022domain,
  title={Domain adaptation and multi-domain adaptation for neural machine translation: A survey},
  author={Saunders, Danielle},
  journal={Journal of Artificial Intelligence Research},
  year={2022}
}

@inproceedings{popovic-2017-chrf,
    title = "chr{F}++: words helping character n-grams",
    author = "Popovi{\'c}, Maja",
    booktitle = "Proceedings of the Second Conference on Machine Translation",
    year = "2017",
    url = "https://aclanthology.org/W17-4770",
    doi = "10.18653/v1/W17-4770"
}

@inproceedings{post-2018-call,
    title = "A Call for Clarity in Reporting {BLEU} Scores",
    author = "Post, Matt",
    booktitle = "Proceedings of the Third Conference on Machine Translation: Research Papers",
    year = "2018",
    url = "https://www.aclweb.org/anthology/W18-6319"
}

@article{nllb-24,
    author="{NLLB Team} and Costa-juss{\`a}, Marta R. and Cross, James and {\c{C}}elebi, Onur and others",
    title="Scaling neural machine translation to 200 languages",
    journal="Nature",
    year="2024",
    issn="1476-4687",
    doi="10.1038/s41586-024-07335-x",
    url="https://doi.org/10.1038/s41586-024-07335-x"
}

@inproceedings{zhou2023rethinking,
  title={Rethinking Round-Trip Translation for Machine Translation Evaluation},
  author={Zhou, Terry Yue and Xu, Qiongkai and He, Xuanli and Cohn, Trevor},
  booktitle={Findings of the Association for Computational Linguistics: EMNLP 2023},
  year={2023},
  url={https://aclanthology.org/2023.findings-emnlp.123}
}

@techreport{oecd2023ailanguagemodels,
  title        = {AI Language Models: Technological, Socio-Economic and Policy Considerations},
  author       = {{OECD}},
  year         = {2023},
  institution  = {Organisation for Economic Co-operation and Development},
  series       = {OECD Digital Economy Papers},
  number       = {352},
  url          = {https://www.oecd-ilibrary.org/science-and-technology/ai-language-models_13d38f92-en},
  urldate      = {2025-10-08}
}

@article{abdin2024phi,
  title   = {Phi-4 Technical Report},
  author  = {Abdin, Marah and Aneja, Jyoti and Behl, Harkirat and Bubeck, Sébastien and Others},
  journal = {arXiv:2412.08905},
  year    = {2024},
  url     = {https://arxiv.org/abs/2412.08905}
}

@article{grattafiori2024llama,
  title={The {Llama} 3 herd of models},
  author={Grattafiori, Aaron and Dubey, Abhimanyu and Jauhri, Abhinav and Pandey, Abhinav and Kadian, Abhishek and Al-Dahle, Ahmad and Letman, Aiesha and Mathur, Akhil and Schelten, Alan and Vaughan, Alex and others},
  journal={arXiv:2407.21783},
  year={2024}
}

@inproceedings{nekoto-etal-2020-participatory,
    title = "Participatory Research for Low-resourced Machine Translation: A Case Study in {A}frican Languages",
    author = {Nekoto, Wilhelmina  and
      Marivate, Vukosi  and
      Matsila, Tshinondiwa  and
      Fasubaa, Timi  and
      Fagbohungbe, Taiwo  and
      Akinola, Solomon Oluwole  and
      Muhammad, Shamsuddeen  and
      Kabongo Kabenamualu, Salomon  and
      Osei, Salomey  and
      Sackey, Freshia  and
      Niyongabo, Rubungo Andre  and
      Macharm, Ricky  and
      Ogayo, Perez  and
      Ahia, Orevaoghene  and
      Berhe, Musie Meressa  and
      Adeyemi, Mofetoluwa  and
      Mokgesi-Selinga, Masabata  and
      Okegbemi, Lawrence  and
      Martinus, Laura  and
      Tajudeen, Kolawole  and
      Degila, Kevin  and
      Ogueji, Kelechi  and
      Siminyu, Kathleen  and
      Kreutzer, Julia  and
      Webster, Jason  and
      Ali, Jamiil Toure  and
      Abbott, Jade  and
      Orife, Iroro  and
      Ezeani, Ignatius  and
      Dangana, Idris Abdulkadir  and
      Kamper, Herman  and
      Elsahar, Hady  and
      Duru, Goodness  and
      Kioko, Ghollah  and
      Espoir, Murhabazi  and
      van Biljon, Elan  and
      Whitenack, Daniel  and
      Onyefuluchi, Christopher  and
      Emezue, Chris Chinenye  and
      Dossou, Bonaventure F. P.  and
      Sibanda, Blessing  and
      Bassey, Blessing  and
      Olabiyi, Ayodele  and
      Ramkilowan, Arshath  and
      {\"O}ktem, Alp  and
      Akinfaderin, Adewale  and
      Bashir, Abdallah},
    booktitle = "Findings of the Association for Computational Linguistics: EMNLP 2020",
    year = "2020",
    url = "https://aclanthology.org/2020.findings-emnlp.195/",
    doi = "10.18653/v1/2020.findings-emnlp.195",
}

@inproceedings{pfeiffer2022liftingcurse,
  title        = {Lifting the Curse of Multilinguality by Pre-Training Modular Transformers},
  author       = {Jonas Pfeiffer and Naman Goyal and Xi Victoria Lin and Xian Li and James Cross and Sebastian Riedel and Mikel Artetxe},
  booktitle    = {Proceedings of the 2022 Conference of the North American Chapter of the Association for Computational Linguistics (NAACL)},
  year         = {2022},
  url          = {https://api.semanticscholar.org/CorpusID:248721770}
}

@inproceedings{chang-etal-2024-multilinguality,
    title = "When Is Multilinguality a Curse? Language Modeling for 250 High- and Low-Resource Languages",
    author = "Chang, Tyler A.  and
      Arnett, Catherine  and
      Tu, Zhuowen  and
      Bergen, Ben",
    booktitle = "Proceedings of the 2024 Conference on Empirical Methods in Natural Language Processing",
    year = "2024",
    url = "https://aclanthology.org/2024.emnlp-main.236/",
    doi = "10.18653/v1/2024.emnlp-main.236"
}

@article{arivazhagan2019massivelymultilingual,
  title        = {Massively Multilingual Neural Machine Translation in the Wild: Findings and Challenges},
  author       = {Arivazhagan, Naveen and Bapna, Ankur and Firat, Orhan and Lepikhin, Dmitry and Johnson, Melvin and Krikun, Maxim and Chen, Mia Xu and Cao, Yuan and Foster, George F. and Cherry, Colin and Macherey, Wolfgang and Chen, Zhifeng and Wu, Yonghui},
  journal      = {arXiv:1907.05019},
  year         = {2019},
  url          = {https://arxiv.org/abs/1907.05019}
}

@inproceedings{deng2024multilingualjailbreakchallenges,
  title        = {Multilingual Jailbreak Challenges in Large Language Models},
  author       = {Yue Deng and Wenxuan Zhang and Sinno Jialin Pan and Lidong Bing},
  booktitle    = {Proceedings of the International Conference on Learning Representations (ICLR)},
  year         = {2024},
  eprint       = {2310.06474},
  archivePrefix= {arXiv},
  primaryClass = {cs.CL},
  url          = {https://arxiv.org/abs/2310.06474}
}

@article{yong2023low,
  title={Low-resource languages jailbreak gpt-4},
  author={Yong, Zheng-Xin and Menghini, Cristina and Bach, Stephen H},
  journal={arXiv:2310.02446},
  year={2023}
}

@article{hernandez2025apertus,
  title={Apertus: Democratizing open and compliant llms for global language environments},
  author={Hern{\'a}ndez-Cano, Alejandro and H{\"a}gele, Alexander and Huang, Allen Hao and Romanou, Angelika and Solergibert, Antoni-Joan and Pasztor, Barna and Messmer, Bettina and Garbaya, Dhia and {\v{D}}urech, Eduard Frank and Hakimi, Ido and others},
  journal={arXiv:2509.14233},
  year={2025}
}

@article{abagyan2025one,
  title={One Tokenizer To Rule Them All: Emergent Language Plasticity via Multilingual Tokenizers},
  author={Abagyan, Diana and Salamanca, Alejandro R and Cruz-Salinas, Andres Felipe and Cao, Kris and Lin, Hangyu and Locatelli, Acyr and Fadaee, Marzieh and {\"U}st{\"u}n, Ahmet and Hooker, Sara},
  journal={arXiv:2506.10766},
  year={2025}
}

@inproceedings{ahia2023tokenizationcost,
  title        = {Do All Languages Cost the Same? Tokenization in the Era of Commercial Language Models},
  author       = {Ahia, Orevaoghene and Kumar, Sachin and Gonen, Hila and Kasai, Jungo and Mortensen, David and Smith, Noah and Tsvetkov, Yulia},
  booktitle    = {Proceedings of the 2023 Conference on Empirical Methods in Natural Language Processing (EMNLP)},
  year         = {2023},
  doi          = {10.18653/v1/2023.emnlp-main.614},
  url          = {https://aclanthology.org/2023.emnlp-main.614}
}

@misc{arnett2025tokenizerfree,
  title        = {There is no such thing as a tokenizer-free lunch},
  author       = {Catherine Arnett},
  howpublished = {\url{https://huggingface.co/blog/catherinearnett/in-defense-of-tokenizers}},
  year         = {2025},
  note         = {Hugging Face Community Article},
  urldate      = {2025-10-08}
}

@inproceedings{arnett2025language,
  title={Why do language models perform worse for morphologically complex languages?},
  author={Arnett, Catherine and Bergen, Benjamin},
  booktitle={Proceedings of the 31st International Conference on Computational Linguistics},
  year={2025}
}

@inproceedings{holtermann-etal-2024-evaluating,
    title = "Evaluating the Elementary Multilingual Capabilities of Large Language Models with {M}ulti{Q}",
    author = {Holtermann, Carolin  and
      R{\"o}ttger, Paul  and
      Dill, Timm  and
      Lauscher, Anne},
    booktitle = "Findings of the Association for Computational Linguistics: ACL 2024",
    year = "2024",
    url = "https://aclanthology.org/2024.findings-acl.265/",
    doi = "10.18653/v1/2024.findings-acl.265"
}

@article{bakouch2025smollm3,
  title   = {{SmolLM3}: smol, multilingual, long-context reasoner},
  author  = {Bakouch, Elie and Ben Allal, Loubna and Lozhkov, Anton and Tazi, Nouamane and Tunstall, Lewis and Patiño, Carlos Miguel and Beeching, Edward and Roucher, Aymeric and Reedi, Aksel Joonas and Gallouédec, Quentin and Rasul, Kashif and Habib, Nathan and Fourrier, Clémentine and Kydlicek, Hynek and Penedo, Guilherme and Larcher, Hugo and Morlon, Mathieu and Srivastav, Vaibhav and Lochner, Joshua and Nguyen, Xuan-Son and Raffel, Colin and von Werra, Leandro and Wolf, Thomas},
  journal = {Hugging Face Blog},
  year    = {2025},
  url     = {https://huggingface.co/blog/smollm3}
}

@article{openai2025gptoss120bgptoss20bmodel,
  title     =   {gpt-oss-120b \& gpt-oss-20b model card},
  author    =   {Agarwal, Sandhini and Ahmad, Lama and Ai, Jason and Altman, Sam and Applebaum, Andy and Arbus, Edwin and Arora, Rahul K and Bai, Yu and Baker, Bowen and Bao, Haiming and others},
  journal   =   {arXiv preprint arXiv:2508.10925},
  year      =   {2025}
}

@article{xuan-etal-2025-mmlu,
  title   = {{MMLU-ProX:} A Multilingual Benchmark for Advanced Large Language Model Evaluation},
  author  = {Xuan, Weihao and Yang, Rui and Qi, Heli and Zeng, Qingcheng and Xiao, Yunze and Others},
  journal = {arXiv:2503.10497},
  year    = {2025},
  url     = {https://arxiv.org/abs/2503.10497}
}

@inproceedings{ahuja2024megaverse,
  title={{MEGAVERSE}: Benchmarking large language models across languages, modalities, models and tasks},
  author={Ahuja, Sanchit and Aggarwal, Divyanshu and Gumma, Varun and Watts, Ishaan and Sathe, Ashutosh and Ochieng, Millicent and Hada, Rishav and Jain, Prachi and Ahmed, Mohamed and Bali, Kalika and others},
  booktitle={Proceedings of the 2024 Conference of the North American Chapter of the Association for Computational Linguistics: Human Language Technologies (Volume 1: Long Papers)},
  year={2024}
}

@article{gao2020pile,
  title={The pile: An 800gb dataset of diverse text for language modeling},
  author={Gao, Leo and Biderman, Stella and Black, Sid and Golding, Laurence and Hoppe, Travis and Foster, Charles and Phang, Jason and He, Horace and Thite, Anish and Nabeshima, Noa and others},
  journal={arXiv:2101.00027},
  year={2020}
}

@inproceedings{abadji-etal-2022-towards,
    title = "Towards a Cleaner Document-Oriented Multilingual Crawled Corpus",
    author = "Abadji, Julien  and
      Ortiz Suarez, Pedro  and
      Romary, Laurent  and
      Sagot, Beno{\^i}t",
    booktitle = "Proceedings of the Thirteenth Language Resources and Evaluation Conference",
    year = "2022",
    url = "https://aclanthology.org/2022.lrec-1.463/"
}

@article{cerebras2023slimpajama,
  title   = {{SlimPajama:} A 627B Token Cleaned and Deduplicated Version of RedPajama},
  author  = {Soboleva, Daria and Al-Khateeb, Faisal and Myers, Robert and Steeves, Jacob R. and Hestness, Joel and Dey, Nolan},
  journal = {Cerebras Blog},
  year    = {2023},
  url     = {https://www.cerebras.net/blog/slimpajama-a-627b-token-cleaned-and-deduplicated-version-of-redpajama}
}

@inproceedings{penedo2024refinedweb,
  title={The {RefinedWeb} dataset for {Falcon LLM}: Outperforming curated corpora with web data only},
  author={Penedo, Guilherme and Malartic, Quentin and Hesslow, Daniel and Cojocaru, Ruxandra and Alobeidli, Hamza and Cappelli, Alessandro and Pannier, Baptiste and Almazrouei, Ebtesam and Launay, Julien},
  booktitle={Advances in Neural Information Processing Systems},
  year={2024}
}

@inproceedings{soldaini-etal-2024-dolma,
    title = "Dolma: an Open Corpus of Three Trillion Tokens for Language Model Pretraining Research",
    author = "Soldaini, Luca  and
      Kinney, Rodney  and
      Bhagia, Akshita  and
      Schwenk, Dustin  and
      Others",
    booktitle = "Proceedings of the 62nd Annual Meeting of the Association for Computational Linguistics (Volume 1: Long Papers)",
    year = "2024",
    url = "https://aclanthology.org/2024.acl-long.840/",
    doi = "10.18653/v1/2024.acl-long.840"
}

@article{peppin2025multilingual,
  title={The Multilingual Divide and Its Impact on Global AI Safety},
  author={Peppin, Aidan and Kreutzer, Julia and Sebag, Alice Schoenauer and Marchisio, Kelly and Ermis, Beyza and Dang, John and Cahyawijaya, Samuel and Singh, Shivalika and Goldfarb-Tarrant, Seraphina and Aryabumi, Viraat and others},
  journal={arXiv:2505.21344},
  year={2025}
}

@techreport{cohere2024languagegap,
  title        = {The AI Language Gap: Considerations on the Multilingual Capabilities of AI Language Models},
  institution  = {Cohere Labs},
  year         = {2024},
  note         = {Policy Primer},
  url          = {https://cohere.com/research/papers/the-ai-language-gap.pdf},  
}

@techreport{openai2025gdpval,
  author      = {Tejal Patwardhan and Rachel Dias and Elizabeth Proehl and Grace Kim and Michele Wang and Olivia Watkins and Simón Posada Fishman and Marwan Aljubbeh and Phoebe Thacker and Laurance Fauconnet and Natalie S. Kim and Patrick Chao and Samuel Miserendino and Gildas Chabot and David Li and Michael Sharman and Alexandra Barr and Amelia Glaese and Jerry Tworek},
  title       = {{GDPVal}: Evaluating AI Model Performance on Real-World Economically Valuable Tasks},
  institution = {OpenAI},
  year        = {2025},
  url         = {https://cdn.openai.com/pdf/d5eb7428-c4e9-4a33-bd86-86dd4bcf12ce/GDPval.pdf},
  note        = {Accessed: 2025-10-06}
}

@techreport{appel2025anthropic_index_report,
  author       = {Ruth Appel and Peter McCrory and Alex Tamkin and Miles McCain and Tyler Neylon and Michael Stern},
  title        = {The Anthropic Economic Index Report: Uneven Geographic and Enterprise AI Adoption},
  institution  = {Anthropic},
  year         = {2025},
  url          = {https://assets.anthropic.com/m/218c82b858610fac/original/Economic-Index.pdf}, 
  note         = {Accessed: 2025-10-06}
}

@inproceedings{huang-etal-2024-chat,
  title     = "{{Chat Vector}: A Simple Approach to Equip {LLM}s with Instruction Following and Model Alignment in New Languages}",
  author    = "Shih‑Cheng Huang and Pin‑Zu Li and Yu‑chi Hsu and Kuang‑Ming Chen and Yu Tung Lin and Shih‑Kai Hsiao and Richard Tsai and Hung‑yi Lee",
  booktitle = "Proceedings of the 62nd Annual Meeting of the Association for Computational Linguistics",
  year      = "2024",

}

@article{wu2025shadow,
  title     = "{{Shadow-FT}: Tuning Instruct Model via Training on Paired Base Model}",
  author    = "Wu, Taiqiang and Yang, Runming and Li, Jiayi and Hu, Pengfei and Wong, Ngai and Yang, Yujiu",
  journal   = "arXiv:2505.12716",
  year      = "2025",
}

@article{tikhonov2021s,
  title={It's all in the heads: Using attention heads as a baseline for cross-lingual transfer in commonsense reasoning},
  author={Tikhonov, Alexey and Ryabinin, Max},
  journal={arXiv:2106.12066},
  year={2021}
}

@inproceedings{ponti2020xcopa,
  title={{XCOPA: A} Multilingual Dataset for Causal Commonsense Reasoning},
author    = {Edoardo M. Ponti and Goran Glava{\v{s}} and Olga Majewska and Qianchu Liu and Ivan Vuli{\'c} and Anna Korhonen},
  booktitle={Proceedings of the 2020 Conference on Empirical Methods in Natural Language Processing (EMNLP)},
  year={2020},
  url={https://ducdauge.github.io/files/xcopa.pdf}
}

@article{lin2021few,
  title={Few-shot learning with multilingual language models},
  author={Lin, Xi Victoria and Mihaylov, Todor and Artetxe, Mikel and Wang, Tianlu and Chen, Shuohui and Simig, Daniel and Ott, Myle and Goyal, Naman and Bhosale, Shruti and Du, Jingfei and others},
  journal={arXiv:2112.10668},
  year={2021}
}

@inproceedings{upadhyay2023xnli,
  title={{XNLI 2.0:} Improving XNLI dataset and performance on Cross Lingual Understanding ({XLU})},
  author={Upadhyay, Ankit Kumar and Upadhya, Harsit Kumar},
  booktitle={2023 IEEE 8th International Conference for Convergence in Technology (I2CT)},
  year={2023}
}

@inproceedings{bandarkar-etal-2024-belebele,
    title = "The Belebele Benchmark: a Parallel Reading Comprehension Dataset in 122 Language Variants",
    author = "Bandarkar, Lucas  and
      Liang, Davis  and
      Muller, Benjamin  and
      Artetxe, Mikel  and
      Shukla, Satya Narayan  and
      Husa, Donald  and
      Goyal, Naman  and
      Krishnan, Abhinandan  and
      Zettlemoyer, Luke  and
      Khabsa, Madian",
    booktitle = "Proceedings of the 62nd Annual Meeting of the Association for Computational Linguistics (Volume 1: Long Papers)",
    year = "2024",
    url = "https://aclanthology.org/2024.acl-long.44/",
    doi = "10.18653/v1/2024.acl-long.44"
}

@article{shi2022language,
  title={Language models are multilingual chain-of-thought reasoners},
  author={Shi, Freda and Suzgun, Mirac and Freitag, Markus and Wang, Xuezhi and Srivats, Suraj and Vosoughi, Soroush and Chung, Hyung Won and Tay, Yi and Ruder, Sebastian and Zhou, Denny and others},
  journal={arXiv:2210.03057},
  year={2022}
}

@article{caswell2025smol,
  title={{SMOL:} Professionally translated parallel data for 115 under-represented languages},
  author={Caswell, Isaac and Nielsen, Elizabeth and Luo, Jiaming and Cherry, Colin and Kovacs, Geza and Shemtov, Hadar and Talukdar, Partha and Tewari, Dinesh and Diane, Baba Mamadi and Doumbouya, Koulako Moussa and others},
  journal={arXiv:2502.12301},
  year={2025}
}

@inproceedings{upadhayay2024taco,
title={{TaCo:} Enhancing Cross-Lingual Transfer for Low-Resource Languages in {LLM}s through Translation-Assisted Chain-of-Thought Processes},
author={Bibek Upadhayay and Vahid Behzadan},
booktitle={5th Workshop on practical ML for limited/low resource settings, ICLR},
year={2024},
url={https://openreview.net/forum?id=02MLWBj8HP}
}

@inproceedings{singh-etal-2024-aya,
    title = "{Aya Dataset:} An Open-Access Collection for Multilingual Instruction Tuning",
    author = {Singh, Shivalika  and
      Vargus, Freddie  and
      D{'}souza, Daniel  and
      Karlsson, B{\"o}rje F.  and
      Mahendiran, Abinaya  and
      Others},
    booktitle = "Proceedings of the 62nd Annual Meeting of the Association for Computational Linguistics (Volume 1: Long Papers)",
    year = "2024",
    url = "https://aclanthology.org/2024.acl-long.620/"
}

@article{maini2025beyondweb,
  title={{BeyondWeb:} Lessons from Scaling Synthetic Data for Trillion-scale Pretraining},
  author={Maini, Pratyush and Dorna, Vineeth and Doshi, Parth and Carranza, Aldo and Pan, Fan and Urbanek, Jack and Burstein, Paul and Fang, Alex and Deng, Alvin and Abbas, Amro and others},
  journal={arXiv:2508.10975},
  year={2025}
}

@article{du2024chinese,
  title={Chinese tiny llm: Pretraining a chinese-centric large language model},
  author={Du, Xinrun and Yu, Zhouliang and Gao, Songyang and Pan, Ding and Cheng, Yuyang and Ma, Ziyang and Yuan, Ruibin and Qu, Xingwei and Liu, Jiaheng and Zheng, Tianyu and others},
  journal={arXiv:2404.04167},
  year={2024}
}

@article{shantipriya2024building,
  title={Building Pre-train LLM Dataset for the Indic Languages: A Case Study on Hindi},
  author={Shantipriya, Parida and Kusum, Lata and  Shakshi, Panwar and  Sanskruti, Mishra},
  journal={arXiv:2407.09855v1},
  year={2024}
}

@article{turker2024vbart,
  title={Vbart: The turkish llm},
  author={Turker, Meliksah and Ari, Mehmet Erdi and Han, Aydin},
  journal={arXiv:2403.01308},
  year={2024}
}

@article{dou2025sailor2,
  title={Sailor2: Sailing in South-East Asia with Inclusive Multilingual LLMs},
  author={Dou, Longxu and Liu, Qian and Zhou, Fan and Chen, Changyu and Wang, Zili and Jin, Ziqi and Liu, Zichen and Zhu, Tongyao and Du, Cunxiao and Yang, Penghui and others},
  journal={arXiv:2502.12982},
  year={2025}
}

@article{aloui2024101,
  title={101 billion arabic words dataset},
  author={Aloui, Manel and Chouikhi, Hasna and Chaabane, Ghaith and Kchaou, Haithem and Dhaouadi, Chehir},
  journal={arXiv:2405.01590},
  year={2024}
}

@article{khan2024indicllmsuite,
  title={{IndicLLMSuite:} a blueprint for creating pre-training and fine-tuning datasets for indian languages},
  author={Khan, Mohammed Safi Ur Rahman and Mehta, Priyam and Sankar, Ananth and Kumaravelan, Umashankar and Doddapaneni, Sumanth and Jain, Sparsh and Kunchukuttan, Anoop and Kumar, Pratyush and Dabre, Raj and Khapra, Mitesh M and others},
  journal={arXiv:2403.06350},
  year={2024}
}

@article{zhao2025qwen3guard,
  title={{Qwen3Guard} Technical Report},
  author={Zhao, Haiquan and Yuan, Chenhan and Huang, Fei and Hu, Xiaomeng and Zhang, Yichang and Yang, An and Yu, Bowen and Liu, Dayiheng and Zhou, Jingren and Lin, Junyang and others},
  journal={arXiv:2510.14276},
  year={2025}
}

@misc{issai2024kazllm10_8b,
  title        = {{LLama-3.1-KazLLM-1.0-8B}},
  author       = {{ISSAI, Nazarbayev University}},
  year         = {2024},
  howpublished = {Model release},
  url          = {https://huggingface.co/issai/LLama-3.1-KazLLM-1.0-8B},
  urldate      = {2025-12-05}
}

@article{koto2025sherkala_chat,
  title   = {{Sherkala-Chat}: Building a State-of-the-Art LLM for Kazakh in a Moderately Resourced Setting},
  author  = {Koto, Fajri and Joshi, Rituraj and Mukhituly, Nurdaulet and Others},
  journal = {arXiv:2503.01493},
  year    = {2025},
  url     = {https://arxiv.org/abs/2503.01493}
}

@article{zellers2019hellaswag,
  title={{HellaSwag:} Can a machine really finish your sentence?},
  author={Zellers, Rowan and Holtzman, Ari and Bisk, Yonatan and Farhadi, Ali and Choi, Yejin},
  journal={arXiv:1905.07830},
  year={2019}
}

@inproceedings{bisk2020piqa,
  title={{PIQA:} Reasoning about physical commonsense in natural language},
  author={Bisk, Yonatan and Zellers, Rowan and Gao, Jianfeng and Choi, Yejin and others},
  booktitle={Proceedings of the AAAI conference on artificial intelligence},
  year={2020}
}

@article{clark2018think,
  title={Think you have solved question answering? try arc, the ai2 reasoning challenge},
  author={Clark, Peter and Cowhey, Isaac and Etzioni, Oren and Khot, Tushar and Sabharwal, Ashish and Schoenick, Carissa and Tafjord, Oyvind},
  journal={arXiv:1803.05457},
  year={2018}
}

@article{costa2022no,
  title={No language left behind: Scaling human-centered machine translation},
  author={Costa-Juss{\`a}, Marta R and Cross, James and {\c{C}}elebi, Onur and Elbayad, Maha and Heafield, Kenneth and Heffernan, Kevin and Kalbassi, Elahe and Lam, Janice and Licht, Daniel and Maillard, Jean and others},
  journal={arXiv:2207.04672},
  year={2022}
}

@inproceedings{adelani-etal-2025-irokobench,
    title = "{I}roko{B}ench: A New Benchmark for {A}frican Languages in the Age of Large Language Models",
    author = "Adelani, David Ifeoluwa  and
      Ojo, Jessica  and
      Azime, Israel Abebe  and
      Zhuang, Jian Yun  and
      Alabi, Jesujoba Oluwadara  and
      He, Xuanli  and
      Ochieng, Millicent  and
      Hooker, Sara  and
      Bukula, Andiswa  and
      Lee, En-Shiun Annie  and
      Chukwuneke, Chiamaka Ijeoma  and
      Buzaaba, Happy  and
      Sibanda, Blessing Kudzaishe  and
      Kalipe, Godson Koffi  and
      Mukiibi, Jonathan  and
      Kabongo Kabenamualu, Salomon  and
      Yuehgoh, Foutse  and
      Setaka, Mmasibidi  and
      Ndolela, Lolwethu  and
      Odu, Nkiruka  and
      Mabuya, Rooweither  and
      Osei, Salomey  and
      Muhammad, Shamsuddeen Hassan  and
      Samb, Sokhar  and
      Guge, Tadesse Kebede  and
      Sherman, Tombekai Vangoni  and
      Stenetorp, Pontus",
    booktitle = "Proceedings of the 2025 Conference of the Nations of the Americas Chapter of the Association for Computational Linguistics: Human Language Technologies (Volume 1: Long Papers)",
    year = "2025",
    url = "https://aclanthology.org/2025.naacl-long.139/",
    doi = "10.18653/v1/2025.naacl-long.139",
    ISBN = "979-8-89176-189-6",
}

@article{lin2021truthfulqa,
  title={{TruthfulQA}: Measuring how models mimic human falsehoods},
  author={Lin, Stephanie and Hilton, Jacob and Evans, Owain},
  journal={arXiv:2109.07958},
  year={2021}
}

@article{ahmadian2024mix,
  title={Mix data or merge models? optimizing for diverse multi-task learning},
  author={Ahmadian, Arash and Goldfarb-Tarrant, Seraphina and Ermis, Beyza and Fadaee, Marzieh and Hooker, Sara and others},
  journal={arXiv:2410.10801},
  year={2024}
}

@article{liu2024deepseek,
  title={Deepseek-v3 technical report},
  author={Liu, Aixin and Feng, Bei and Xue, Bing and Wang, Bingxuan and Wu, Bochao and Lu, Chengda and Zhao, Chenggang and Deng, Chengqi and Zhang, Chenyu and Ruan, Chong and others},
  journal={arXiv:2412.19437},
  year={2024}
}

@article{team2025gemma,
  title={Gemma 3 technical report},
  author={Team, Gemma and Kamath, Aishwarya and Ferret, Johan and Pathak, Shreya and Vieillard, Nino and Merhej, Ramona and Perrin, Sarah and Matejovicova, Tatiana and Ram{\'e}, Alexandre and Rivi{\`e}re, Morgane and others},
  journal={arXiv:2503.19786},
  year={2025}
}

@inproceedings{parrish-etal-2022-bbq,
    title = "{BBQ}: A hand-built bias benchmark for question answering",
    author = "Parrish, Alicia  and
      Chen, Angelica  and
      Nangia, Nikita  and
      Padmakumar, Vishakh  and
      Phang, Jason  and
      Thompson, Jana  and
      Htut, Phu Mon  and
      Bowman, Samuel",
    booktitle = "Findings of the Association for Computational Linguistics: ACL 2022",
    year = "2022",
    url = "https://aclanthology.org/2022.findings-acl.165",
    doi = "10.18653/v1/2022.findings-acl.165"
}

@inproceedings{hartvigsen2022toxigen,
  title={{ToxiGen}: A Large-Scale Machine-Generated Dataset for Implicit and Adversarial Hate Speech Detection},
  author={Hartvigsen, Thomas and Gabriel, Saadia and Palangi, Hamid and Sap, Maarten and Ray, Dipankar and Kamar, Ece},
  booktitle={Proceedings of the 60th Annual Meeting of the Association for Computational Linguistics},
  year={2022}
}

@inproceedings{gehman-etal-2020-realtoxicityprompts,
    title = "{R}eal{T}oxicity{P}rompts: Evaluating Neural Toxic Degeneration in Language Models",
    author = "Gehman, Samuel  and
      Gururangan, Suchin  and
      Sap, Maarten  and
      Choi, Yejin  and
      Smith, Noah A.",
    booktitle = "Findings of the Association for Computational Linguistics: EMNLP 2020",
    year = "2020",
    url = "https://aclanthology.org/2020.findings-emnlp.301/",
    doi = "10.18653/v1/2020.findings-emnlp.301"
}

@article{yang2025qwen3,
  title={Qwen3 technical report},
  author={Yang, An and Li, Anfeng and Yang, Baosong and Zhang, Beichen and Hui, Binyuan and Zheng, Bo and Yu, Bowen and Gao, Chang and Huang, Chengen and Lv, Chenxu and others},
  journal={arXiv:2505.09388},
  year={2025}
}

@article{penedo2024fineweb,
  title={The fineweb datasets: Decanting the web for the finest text data at scale},
  author={Penedo, Guilherme and Kydl{\'\i}{\v{c}}ek, Hynek and Lozhkov, Anton and Mitchell, Margaret and Raffel, Colin A and Von Werra, Leandro and Wolf, Thomas and others},
  journal={Advances in Neural Information Processing Systems},
  year={2024}
}

@article{singh2024global,
  title={Global {MMLU}: Understanding and addressing cultural and linguistic biases in multilingual evaluation},
  author={Singh, Shivalika and Romanou, Angelika and Fourrier, Cl{\'e}mentine and Adelani, David I and Ngui, Jian Gang and Vila-Suero, Daniel and Limkonchotiwat, Peerat and Marchisio, Kelly and Leong, Wei Qi and Susanto, Yosephine and others},
  journal={arXiv:2412.03304},
  year={2024}
}

@inproceedings{joshi-etal-2020-state,
    title = "The State and Fate of Linguistic Diversity and Inclusion in the {NLP} World",
    author = "Joshi, Pratik  and
      Santy, Sebastin  and
      Budhiraja, Amar  and
      Bali, Kalika  and
      Choudhury, Monojit",
    booktitle = "Proceedings of the 58th Annual Meeting of the Association for Computational Linguistics",
    year = "2020",
    url = "https://aclanthology.org/2020.acl-main.560/",
    doi = "10.18653/v1/2020.acl-main.560"
}

@article{yadav2023ties,
  title={Ties-merging: Resolving interference when merging models},
  author={Yadav, Prateek and Tam, Derek and Choshen, Leshem and Raffel, Colin A and Bansal, Mohit},
  journal={Advances in Neural Information Processing Systems},
  year={2023}
}

@article{matena2022merging,
  title={Merging models with fisher-weighted averaging},
  author={Matena, Michael S and Raffel, Colin A},
  journal={Advances in Neural Information Processing Systems},
  year={2022}
}

@article{rafailov2023direct,
  title={Direct preference optimization: Your language model is secretly a reward model},
  author={Rafailov, Rafael and Sharma, Archit and Mitchell, Eric and Manning, Christopher D and Ermon, Stefano and Finn, Chelsea},
  journal={Advances in neural information processing systems},
  year={2023}
}

@article{MultiWikiQA2025,
  title={{MultiWikiQA:} A Reading Comprehension Benchmark in 300+ Languages},
  author={ Dan Saattrup Smart},
  journal={arXiv:2509.04111v1},
  year={2025}
}

@article{dubois2023alpacafarm,
  title={{AlpacaFarm:} A simulation framework for methods that learn from human feedback},
  author={Dubois, Yann and Li, Chen Xuechen and Taori, Rohan and Zhang, Tianyi and Gulrajani, Ishaan and Ba, Jimmy and Guestrin, Carlos and Liang, Percy S and Hashimoto, Tatsunori B},
  journal={Advances in Neural Information Processing Systems},
  year={2023}
}

@misc{kydlicek2025finepdfs,
  title        = {{FinePDFs}},
  author       = {Hynek Kydlíček and Guilherme Penedo and Leandro von Werra},
  year         = {2025},
  howpublished = {\url{https://huggingface.co/datasets/HuggingFaceFW/finepdfs}},
  note         = {Hugging Face dataset}
}

@inproceedings{ustun-etal-2024-aya,
    title = "{Aya Model:} An Instruction Finetuned Open-Access Multilingual Language Model",
    author = {{\"U}st{\"u}n, Ahmet  and
      Aryabumi, Viraat  and
      Yong, Zheng  and
      Ko, Wei-Yin  and
      D{'}souza, Daniel  and
      Onilude, Gbemileke  and
      Bhandari, Neel  and
      Singh, Shivalika  and
      Ooi, Hui-Lee  and
      Kayid, Amr  and
      Vargus, Freddie  and
      Blunsom, Phil  and
      Longpre, Shayne  and
      Muennighoff, Niklas  and
      Fadaee, Marzieh  and
      Kreutzer, Julia  and
      Hooker, Sara},
    booktitle = "Proceedings of the 62nd Annual Meeting of the Association for Computational Linguistics (Volume 1: Long Papers)",
    year = "2024",
    url = "https://aclanthology.org/2024.acl-long.845/",
    doi = "10.18653/v1/2024.acl-long.845"
}

@article{penedo2025fineweb2,
  title={{FineWeb2}: One Pipeline to Scale Them All--Adapting Pre-Training Data Processing to Every Language},
  author={Penedo, Guilherme and Kydl{\'\i}{\v{c}}ek, Hynek and Sabol{\v{c}}ec, Vinko and Messmer, Bettina and Foroutan, Negar and Kargaran, Amir Hossein and Raffel, Colin and Jaggi, Martin and Von Werra, Leandro and Wolf, Thomas},
  journal={arXiv:2506.20920},
  year={2025}
}

@inproceedings{wenzek-etal-2020-ccnet,
    title = "{CCN}et: Extracting High Quality Monolingual Datasets from Web Crawl Data",
    author = "Wenzek, Guillaume  and
      Lachaux, Marie-Anne  and
      Conneau, Alexis  and
      Chaudhary, Vishrav  and
      Guzm{\'a}n, Francisco  and
      Joulin, Armand  and
      Grave, Edouard",
    booktitle = "Proceedings of the Twelfth Language Resources and Evaluation Conference",
    year = "2020",
    url = "https://aclanthology.org/2020.lrec-1.494/",
}

@article{chung2023unimax,
  title={Unimax: Fairer and more effective language sampling for large-scale multilingual pretraining},
  author={Chung, Hyung Won and Constant, Noah and Garcia, Xavier and Roberts, Adam and Tay, Yi and Narang, Sharan and Firat, Orhan},
  journal={arXiv:2304.09151},
  year={2023}
}

@article{team2024gemma,
  title={Gemma 2: Improving open language models at a practical size},
  author={Team, Gemma and Riviere, Morgane and Pathak, Shreya and Sessa, Pier Giuseppe and Hardin, Cassidy and Bhupatiraju, Surya and Hussenot, L{\'e}onard and Mesnard, Thomas and Shahriari, Bobak and Ram{\'e}, Alexandre and others},
  journal={arXiv:2408.00118},
  year={2024}
}

@inproceedings{conneau-etal-2020-unsupervised,
    title = "Unsupervised Cross-lingual Representation Learning at Scale",
    author = "Conneau, Alexis  and
      Khandelwal, Kartikay  and
      Goyal, Naman  and
      Chaudhary, Vishrav  and
      Wenzek, Guillaume  and
      Guzm{\'a}n, Francisco  and
      Grave, Edouard  and
      Ott, Myle  and
      Zettlemoyer, Luke  and
      Stoyanov, Veselin",
    booktitle = "Proceedings of the 58th Annual Meeting of the Association for Computational Linguistics",
    year = "2020",
    url = "https://aclanthology.org/2020.acl-main.747/",
    doi = "10.18653/v1/2020.acl-main.747",
}

@article{de2024new,
  title={A new massive multilingual dataset for high-performance language technologies},
  author={De Gibert, Ona and Nail, Graeme and Arefyev, Nikolay and Ba{\~n}{\'o}n, Marta and Van Der Linde, Jelmer and Ji, Shaoxiong and Zaragoza-Bernabeu, Jaume and Aulamo, Mikko and Ram{\'\i}rez-S{\'a}nchez, Gema and Kutuzov, Andrey and others},
  journal={arXiv:2403.14009},
  year={2024}
}

@inproceedings{nguyen-etal-2024-culturax,
    title = "{C}ultura{X}: A Cleaned, Enormous, and Multilingual Dataset for Large Language Models in 167 Languages",
    author = "Nguyen, Thuat  and
      Nguyen, Chien Van  and
      Lai, Viet Dac  and
      Man, Hieu  and
      Ngo, Nghia Trung  and
      Dernoncourt, Franck  and
      Rossi, Ryan A.  and
      Nguyen, Thien Huu",
    booktitle = "Proceedings of the 2024 Joint International Conference on Computational Linguistics, Language Resources and Evaluation (LREC-COLING 2024)",
    year = "2024",
    url = "https://aclanthology.org/2024.lrec-main.377/"
}

@inproceedings{rein2024gpqa,
  title={{GPQA:} A graduate-level google-proof q\&a benchmark},
  author={Rein, David and Hou, Betty Li and Stickland, Asa Cooper and Petty, Jackson and Pang, Richard Yuanzhe and Dirani, Julien and Michael, Julian and Bowman, Samuel R},
  booktitle={First Conference on Language Modeling},
  year={2024}
}

@article{suzgun2022challenging,
  title={Challenging big-bench tasks and whether chain-of-thought can solve them},
  author={Suzgun, Mirac and Scales, Nathan and Sch{\"a}rli, Nathanael and Gehrmann, Sebastian and Tay, Yi and Chung, Hyung Won and Chowdhery, Aakanksha and Le, Quoc V and Chi, Ed H and Zhou, Denny and others},
  journal={arXiv:2210.09261},
  year={2022}
}

@article{chen2021evaluating,
  title={Evaluating Large Language Models Trained on Code},
  author={Chen, Mark and Tworek, Jerry and Jun, Heewoo and Yuan, Qiming and Pinto, Henrique Ponde de Oliveira and Kaplan, Jared and Edwards, Harri and Burda, Yuri and Joseph, Nicholas and Brockman, Greg and others},
  journal={arXiv:2107.03374},
  year={2021}
}

@article{zhou2023instruction,
  title={Instruction-Following Evaluation for Large Language Models},
  author={Zhou, Jeffrey and Lu, Tianjian and Mishra, Swaroop and Brahma, Siddhartha and Basu, Sujoy and Luan, Yi and Zhou, Denny and Hou, Le},
  journal={arXiv:2311.07911},
  year={2023}
}

@article{laurenccon2022bigscience,
  title={The bigscience roots corpus: A 1.6 tb composite multilingual dataset},
  author={Lauren{\c{c}}on, Hugo and Saulnier, Lucile and Wang, Thomas and Akiki, Christopher and Villanova del Moral, Albert and Le Scao, Teven and Von Werra, Leandro and Mou, Chenghao and Gonz{\'a}lez Ponferrada, Eduardo and Nguyen, Huu and others},
  journal={Advances in Neural Information Processing Systems},
  year={2022}
}

@article{kudugunta2023madlad,
  title={Madlad-400: A multilingual and document-level large audited dataset},
  author={Kudugunta, Sneha and Caswell, Isaac and Zhang, Biao and Garcia, Xavier and Xin, Derrick and Kusupati, Aditya and Stella, Romi and Bapna, Ankur and Firat, Orhan},
  journal={Advances in Neural Information Processing Systems},
  year={2023}
}

@article{weber2024redpajama,
  title={{RedPajama:} an open dataset for training large language models},
  author={Weber, Maurice and Fu, Dan and Anthony, Quentin and Oren, Yonatan and Adams, Shane and Alexandrov, Anton and Lyu, Xiaozhong and Nguyen, Huu and Yao, Xiaozhe and Adams, Virginia and others},
  journal={Advances in neural information processing systems},
  year={2024}
}

@article{roussis2025krikri,
  title={Krikri: Advancing Open Large Language Models for Greek},
  author={Roussis, Dimitris and Voukoutis, Leon and Paraskevopoulos, Georgios and Sofianopoulos, Sokratis and Prokopidis, Prokopis and Papavasileiou, Vassilis and Katsamanis, Athanasios and Piperidis, Stelios and Katsouros, Vassilis},
  journal={arXiv:2505.13772},
  year={2025}
}
\appendix

\newpage

\section{Related Work}
\textbf{Multilingual pretraining datasets.}
Large-scale multilingual datasets~\cite{penedo2025fineweb2,kydlicek2025finepdfs,de2024new,nguyen-etal-2024-culturax,weber2024redpajama,laurenccon2022bigscience,wenzek-etal-2020-ccnet,conneau-etal-2020-unsupervised,xue2021mt5} are the key component of building effective multilingual LLMs. Early corpora such as CC-100~\cite{wenzek-etal-2020-ccnet}, mC4~\cite{xue2021mt5}, and HPLT~\cite{de2024new,burchell2025expanded} rely on uniform, language-agnostic filtering pipelines, which often lead to uneven coverage and degraded quality for low- and mid-resource languages.
\FINEWEB~\cite{penedo2025fineweb2} introduces a language-adaptive filtering and deduplication pipeline covering 1,868 language–script pairs. It provides detailed per-language metadata and yields monolingual models that outperform counterparts trained on HPLT~\cite{de2024new}, HPLT-2~\cite{burchell2025expanded}, CulturaX~\cite{nguyen-etal-2024-culturax}, and CC-100~\cite{wenzek-etal-2020-ccnet,conneau-etal-2020-unsupervised}. FinePDFs~\cite{kydlicek2025finepdfs} complements this work with a 3T-token, 475M-document corpus spanning 1,733 language–script pairs, built from a dedicated PDF extraction and OCR pipeline.

In parallel, a growing number of high-quality \emph{language-specific} pretraining datasets has emerged for individual languages, including Arabic 
(ArabicWeb24\footnote{https://huggingface.co/blog/MayFarhat/arabicweb24}, Arabic-101B~\cite{aloui2024101}), Hindi and Telugu (Sangraha~\cite{khan2024indicllmsuite}), Hindi (Odaigen~\cite{shantipriya2024building}), French (Croissant~\cite{croissantllm2024}), Russian (OmniaRussica\footnote{https://omnia-russica.github.io/}), Thai (SeaCommonCrawl~\cite{dou2025sailor2}), Turkish (VNGRS-WebCorpus~\cite{turker2024vbart}), Chinese (Tiger-Bot\footnote{https://github.com/TigerResearch/TigerBot}, MAP-CC~\cite{du2024chinese}), Icelandic (The Icelandic Gigaword Corpus\footnote{https://clarin.is/en/}), and Catalan\footnote{{https://huggingface.co/datasets/projecte-aina/catalan\_general\_crawling}}. 
These datasets highlight the benefits of language-aware curation for improving downstream model quality.

\vspace{0.4em}\textbf{Multilingual and language-centric LLMs.}
Recent progress in multilingual LLMs has been driven by both closed-weight and open-weight frontier systems with broad language coverage, including mT5~\cite{xue2021mt5}, Aya-101~\cite{ustun-etal-2024-aya}, Aya-23~\cite{aryabumi2024aya}, \QWEN~\cite{yang2025qwen3}, and \GEMMA~\cite{team2025gemma}. \GEMMA notably revised its training mixture relative to Gemma-2~\cite{team2024gemma} by increasing multilingual data, adding monolingual and parallel corpora, and mitigating language imbalance using a UNIMAX-inspired sampling strategy~\cite{chung2023unimax}.
Open and region-focused initiatives have expanded accessible multilingual coverage: Aya-101 increases breadth and improves instruction mixture balance~\cite{ustun-etal-2024-aya}; Aya-23 adopts a depth-focused decoder-only design over 23 languages~\cite{aryabumi2024aya}; Sailor2 targets Southeast Asian languages through large-scale continual pretraining with open recipes~\cite{dou2025sailor2}; and \APERTUS emphasizes transparency and full reproducibility across a wide multilingual scope~\cite{hernandez2025apertus}. Europe-centric efforts such as EuroLLM and Teuken further underscore the importance of regional coverage and tokenizer design~\cite{martins2025eurollm,ali2024teuken}.

Complementing these multilingual generalist models, several language-centric open-weight models demonstrate the value of targeted specialization for mid- and high-resource settings, including Arabic (\textsc{Jais}~\cite{sengupta2023jais}), Hindi (\textsc{Nanda}~\cite{nanda2025}), French (Lucie-7B~\cite{gouvert2025lucie}, CroissantLLM~\cite{croissantllm2024}), Finnish (Poro-34B~\cite{poro2024}), Catalan (Àguila-7B~\cite{aguila2023}), Kazakh (Sherkala-Chat~\cite{koto2025sherkala_chat}, KazLLM-1.0-8B~\cite{issai2024kazllm10_8b}), Greek (Llama-Krikri-8B~\cite{roussis2025krikri}) and Swahili (UlizaLlama~\cite{ulizallama2023}).

Despite these advances, the long tail of low- and extreme-low-resource languages still lacks a unified, open, and tier-aware integration framework that links realistic data availability to concrete adaptation strategies. \BYOL addresses this gap by introducing a four-tier digital-resource taxonomy and mapping each tier to an appropriate integration path—direct finetuning, continual pretraining, or translation-mediated inclusion.

\vspace{0.4em}\textbf{Model Merging.}
Model merging has emerged as an effective strategy for consolidating the complementary strengths of multiple LLMs.
Early efforts demonstrated that finetuned models sharing the same pretrained backbone can be combined through simple parameter averaging, often yielding merged models that outperform the individual experts~\cite{pmlr-v162-wortsman22a,matena2022merging,gupta2020stochasticweightaveragingparallel}. Subsequent research explored more structured and non-linear merging techniques~\cite{yadav2023ties,yu2024language}, aiming to improve generalization and robustness across diverse downstream tasks. Alongside these technical advances, a growing body of work highlights the safety risks of merging: harmful or misaligned behaviors can transfer directly from one model to the merged result~\cite{hammoud2024modelmergingsafetyalignment}. This has motivated techniques that identify and manipulate safety subspaces, allowing aligned models to be fused with domain-specific experts while preserving desirable behaviors~\cite{yu2024language}.

Complementary lines of work apply merging to domain-specialized experts in safety, coding, and reasoning~\cite{wu2025shadow,ahmadian2024mix}. Shadow-FT~\cite{wu2025shadow} adopts a paired base–instruct setup, finetuning the base model and grafting its updates onto the instruct model, and reports consistent gains over alternative merging schemes. Ahmadian et al.~\cite{ahmadian2024mix} show that objective-driven merging of safety and general-purpose experts, including language-focused experts, outperforms simply mixing their training data, particularly in multilingual settings.
Merging has also been extended to multilingual settings, enabling the construction of task-capable LLMs for high-resource languages without requiring supervised finetuning data in the target language~\cite{Tao_2024}. Our work builds on these developments with a goal of bringing low-resource language expertise into the baseline LLM while preserving its multilingual and safety behaviors, without requiring additional alignment training.

\section{Training Datasets}
\subsection{Continual Pretraining Datasets}
\label{annex:cpt datasets}

Table~\ref{tab:pretrain_datasets} (Sec.~\ref{sec:ablation}) summarizes the four dataset configurations (C1–C4) used in our continual pretraining ablation study. The goal is to quantify, relative to a low-resource raw corpus (C1), the individual and combined contributions of refined low-resource corpora (C2), refined English corpora (C3), and their combination with the translation of refined English data (C4). The raw pretraining data for C1 are drawn from the multilingual \FINEWEB corpus~\cite{penedo2025fineweb2}, while the English corpus is derived from the high-quality \FineWebEdu English dataset~\cite{penedo2024fineweb}.

We use \texttt{Azure OpenAI GPT-5-mini} to refine the English corpus and \texttt{Azure OpenAI GPT-5} (reasoning) to refine the target-language corpora, following the prompt structure in Annex~\ref{app:data-refinement-prompt}. Although many large open- and closed-weight language models could be used to refine English text, refinement in low-resource languages such as Chichewa and Māori requires particular care, as it demands strong comprehension and generation capabilities in those languages. For this reason, we use the larger-capacity \texttt{Azure OpenAI GPT-5 (reasoning)} model for the low-resource corpora, where modeling errors are more likely to propagate into downstream training stages. The English data is added in the same proportion as the low-resource language data, yielding a 1:1:1 mixture (in number of tokens) of refined low-resource text, refined English, and translated refined English into the low-resource language. This mixture in C4 dataset contains approximately 433M tokens for Chichewa and 745M tokens for Māori.

\begin{table}[!t]
\centering
\caption{\small  \underline{ \bf Dataset mixture used for supervised finetuning} of Chichewa and Māori LLMs.}
\label{tab:chat-data-merged}
\adjustbox{width=\textwidth,center}{
\begin{tabular}{l c c r r r r}
\toprule
\textbf{Dataset} & & \textbf{Language} & \textbf{\# Samples} & \textbf{\# Prompt tokens} & \textbf{\# Response tokens} & \textbf{\# Total tokens} \\
\midrule
\rowcolor{gray!15} \multicolumn{7}{l}{\textbf{Chichewa SFT datasets}} \\
Aya subset$^\star$~\cite{singh-etal-2024-aya}                     & $\rightarrow$ MT & \texttt{nya} & 10{,}884  & 512{,}865   & 1{,}372{,}908   & 1{,}885{,}773 \\
SmolSent/Doc~\cite{caswell2025smol}        &                  & (\texttt{eng}, $\,$ \texttt{nya}) & 7{,}992   & 242{,}220   & 186{,}844      & 429{,}064 \\
SmolTalk2 subset$^\ddagger$~\cite{bakouch2025smollm3} &                  & \texttt{eng} & 350{,}577 & 60{,}325{,}447 & 126{,}499{,}856 & 186{,}825{,}303 \\
SmolTalk2 subset$^\ddagger$~\cite{bakouch2025smollm3} & $\rightarrow$ MT & \texttt{nya} & 350{,}175 & 85{,}530{,}155 & 164{,}008{,}260 & 249{,}538{,}415 \\
\midrule
 \textbf{Total (Chichewa)} & & & 719{,}628 & 146{,}555{,}311 & 292{,}067{,}868 & 438{,}678{,}555 \\
\midrule
\rowcolor{gray!15} \multicolumn{7}{l}{\textbf{Māori SFT datasets}} \\
Aya subset$^\otimes$~\cite{singh-etal-2024-aya}             & $\rightarrow$ MT & \texttt{mri} & 10{,}196  & 559{,}460   & 1{,}392{,}256   & 1{,}951{,}716 \\
Alpaca-Maori-cleaned~\cite{upadhayay2024taco} &              & \texttt{mri} & 41{,}601  & 1{,}181{,}880 & 9{,}128{,}222   & 10{,}310{,}102 \\
SmolTalk2 subset$^\ddagger$~\cite{bakouch2025smollm3} &                  & \texttt{eng} & 350{,}577 & 60{,}325{,}447 & 126{,}499{,}856 & 186{,}825{,}303 \\
SmolTalk2 subset$^\ddagger$~\cite{bakouch2025smollm3} & $\rightarrow$ MT &  \texttt{mri} & 350{,}108 & 96{,}637{,}518 & 184{,}777{,}617 & 281{,}415{,}135 \\
\midrule
\textbf{Total (Māori)} & & & 752{,}482 & 158{,}704{,}305 & 321{,}797{,}951 & 480{,}502{,}256 \\
\bottomrule
\end{tabular}
}
\begin{flushleft}\footnotesize
$^\star$ Aya subsets from five high-resource languages, English, Italian, German, French, and Spanish, were used as sources for translation, in addition to the 688 native Chichewa samples.   
$^\otimes$ Aya subsets from five high-resource languages were used as sources for translation. 
$^\ddagger$ Samples from non-reasoning subsets of SmolTalk2 are drawn in English and machine-translated (MT) into the target language $l$. 
\end{flushleft}
\end{table}

\subsection{Instruction Finetuning Datasets}
\label{annex:SFT_Data}
Table~\ref{tab:chat-data-merged} summarizes the composition of our dataset mixtures for supervised finetuning.
The \AYA Dataset and \AYA Collection~\cite{singh-etal-2024-aya} are among the few large-scale efforts that pair human-annotated multilingual instructions with scaled multilingual expansion via templating and machine translation. The \AYA Dataset provides 204K human-written prompt–completion pairs across 65 languages, including Chichewa with 688 samples, while the \AYA Collection broadens coverage by applying speaker-designed templates and translating selected instruction datasets into 101 languages.  However, to the best of our knowledge, no large-scale, language-dedicated human-curated instruction dataset currently exists for Chichewa or Māori.
To address this gap, we construct two instruction dataset mixtures specifically designed to support finetuning in these languages. Our mixtures integrate machine-translated subsets of \SMOLTALK~\cite{bakouch2025smollm3} with additional translated data from five high-resource languages in the \AYA Dataset~\cite{singh-etal-2024-aya}, extended to both Chichewa and Māori.  For Chichewa, we further incorporate samples from the SmolSent and SmolDoc train-paired datasets~\cite{caswell2025smol}, while for Māori we include cleaned data from the alpaca-maori corpus~\cite{upadhayay2024taco}.  English instruction samples from \SMOLTALK are retained in both mixtures to ensure bilingual consistency and cross-lingual alignment.  These mixtures yield SFT datasets of approximately 438M tokens (146M prompt tokens / 292M response tokens) for Chichewa and 480M tokens (158M prompt tokens / 321M response tokens) for Māori.

\begin{table}[!t]
\centering
\caption{\small \textbf{\underline{Inuktitut--English paired datasets}} for training and testing MT models. News Article and Children Books datasets, labeled as internal, are provided by the Government of Nunavut, Canada. }
\label{tab:mt-datasets}
\adjustbox{width=\textwidth,center}{
\scalebox{0.8}{
\begin{tabular}{lccc}
\toprule
\bf Dataset & \bf \# Training Samples & \bf \# Dev-Test Samples & \bf \# Test Samples \\
\midrule
Nunavut-Hansard (NH 3.0)~\cite{joanis-etal-2020-nunavut} & 1.3 million & 2,658 & 3,573\\
News Articles (Internal)   & 16,428 & -- & 864\\
Children Books (Internal)  & 13,204 & -- & 692 \\
\bottomrule
\end{tabular}
}
}
\end{table}

\subsection{Machine Translation Datasets}
\label{ref:mt-datasets}
For training our MT models, we use three human-translated datasets, summarized in Table~\ref{tab:mt-datasets}. In addition to these datasets, we also include synthetic back-translated data. For English→Inuktitut training, we back-translate 70,000 Inuktitut sentences from \FINEWEB~\cite{penedo2025fineweb2} into English. And for Inuktitut→English model training, we back-translate 497,500 English sentences from various sources, listed in Table~\ref{tab:bt-datasets}.

\begin{table}[!t]
\centering
\caption{\small \underline{\textbf{Composition of the English monolingual dataset}} used for back-translation.}
\label{tab:bt-datasets}
\adjustbox{width=\textwidth,center}{
\scalebox{0.85}{
\begin{tabular}{lcr}
\toprule
\textbf{Data Source Domain} & \textbf{Datasets Included} & \textbf{(Random) Samples} \\
\midrule
Wikipedia \& General Knowledge 
  & WikiMatrix~\cite{schwenk-etal-2021-wikimatrix}, 
    Simple Wikipedia~\cite{simplewiki}, 
    Tatoeba~\cite{tatoeba}
  & 212,500 \\

News \& Commentary 
  & News Commentary~\cite{tiedemann-2012-parallel}, 
    Global Voices~\cite{nguyen-daume-iii-2019-global}, 
    AG News~\cite{zhang2015character}, 
    XSum~\cite{narayan2018don}
  & 195,000 \\

Instructional \& How-To 
  & WikiHow~\cite{koupaee2018wikihow}
  & 45,000 \\

Q\&A and Long-Form Text 
  & GooAQ~\cite{khashabi2021gooaq}, 
    Natural Questions~\cite{kwiatkowski2019natural}, 
    ELI5~\cite{fan2019eli5}, 
    SQuAD~\cite{rajpurkar2016squad}
  & 30,000 \\

Conversational \& Talks 
  & TED Talks~\cite{cettolo2012ted}
  & 15,000 \\
\midrule
\textbf{Total} & & \textbf{497,500} \\
\bottomrule
\end{tabular}
}
}
\end{table}

\section{Evaluation Datasets}
\label{annex:bench}
This section describes the benchmarks used to evaluate our adapted LLMs (\BYOLC and \BYOLM) on Chichewa and Māori, as well as the English benchmarks used to assess their English knowledge performance. All evaluations are conducted with the LM Evaluation Harness (lm-eval)~\cite{eval-harness}, a standard framework widely used in the \ac{llm} community.

\subsection{Base Model Benchmarking Datasets}
\label{annex:base-eval-datasets}
Table~\ref{tab:pre-train-eval} summarizes the benchmarks and experimental settings used to evaluate our base models after continual pretraining (\BYOLC-CPT and \BYOLM-CPT). We report results on fourteen datasets: twelve task-specific benchmarks in both English and the target language, and two English-only benchmarks, GPQA Diamond~\cite{suzgun2022challenging} and BIG-bench Hard (BBH)~\cite{suzgun2022challenging}, which evaluate graduate-level scientific reasoning and complex problem-solving abilities. Among the target-language benchmarks, three (\GLOBALMMLU~\cite{singh2024global}, Belebele~\cite{bandarkar-etal-2024-belebele}, and FLORES-200~\cite{costa2022no}) were translated into Chichewa and Māori by professional human translators, while the remaining benchmarks were translated using Azure's MT system.

\begin{table}[!t]
\caption{\small \underline{ \textbf{Benchmarks used to evaluate our base models after continual pre-training.}} }
\label{tab:pre-train-eval}
\adjustbox{width=\textwidth,center}{
\begin{tabular}{lllcccc}
\toprule
 \textbf{Benchmark} & \textbf{Languages} & \textbf{Category} & \textbf{Task} & \textbf{n-shot} & \textbf{Metric} & \bf Norm \\
\midrule
\rowcolor{gray!15} \multicolumn{7}{l}{\textbf{General \& STEM reasoning}}\\
\addlinespace[2pt]
Global MMLU-Lite$^\star$~\cite{singh2024global} & Chichewa, Māori  & General knowledge/Reasoning & MCQ & 5 & Accuracy & \\
 ARC-Easy$^\ddagger$~\cite{clark2018think} & Chichewa, Māori & Science reasoning (easy) & MCQ & 0 & Accuracy & Char--Len \\
ARC-Hard$^\ddagger$~\cite{clark2018think} & Chichewa, Māori & Science reasoning (hard) & MCQ & 25 & Accuracy  & Char--Len  \\
 MGSM$^\ddagger$~\cite{shi2022language} & Chichewa, Māori & Math reasoning & Generation & 8 & Accuracy (EM) & \\
 BBH~\cite{suzgun2022challenging} & English  & Complex Reasoning & Generation & 3 & Accuracy (EM)  & \\
 GPQA Diamond~\cite{rein2024gpqa} & English  & Graduate Science & MCQ & 5 & Accuracy &    \\ \midrule 
\addlinespace[4pt]
\rowcolor{gray!15} \multicolumn{7}{l}{\textbf{Commonsense \& story}}\\
\addlinespace[2pt]
XCOPA$^\ddagger$~\cite{ponti2020xcopa} & Chichewa, Māori & Causal reasoning & MCQ & 5 & Accuracy & \\
 XStoryCloze$^\ddagger$~\cite{lin2021few} & Chichewa, Māori & Story completion/commonsense & MCQ & 5 & Accuracy & \\
PIQA$^\ddagger$~\cite{bisk2020piqa} & Chichewa, Māori & Physical commonsense & MCQ & 0 & Accuracy & Char--Len  \\
HellaSwag$^\ddagger$~\cite{zellers2019hellaswag} & Chichewa, Māori & Sentence completion & MCQ & 10 &  Accuracy &  Char--Len \\ \midrule 
\addlinespace[4pt]
\rowcolor{gray!15} \multicolumn{7}{l}{\textbf{NLI, coreference, and reading/QA}}\\
\addlinespace[2pt]
XNLI 2.0$^\ddagger$~\cite{upadhyay2023xnli} & Chichewa, Māori & Natural language inference & MCQ & 5 & Accuracy & \\
 XWinograd$^\ddagger$~\cite{tikhonov2021s} & Chichewa, Māori & Coreference resolution & MCQ & 0 & Accuracy & \\
 Belebele$^\star$~\cite{bandarkar-etal-2024-belebele} & Chichewa, Māori & Reading comprehension & MCQ & 1 & Accuracy &  \\
\addlinespace[4pt]
\rowcolor{gray!15} \multicolumn{7}{l}{\textbf{Translation}}\\ 
\addlinespace[2pt]
 FLORES-200$^\star$~\cite{costa2022no} & Chichewa, Māori & Translation & Generation & 1 & BLEU, chrF++  &\\ 
\addlinespace[2pt]
\bottomrule
\end{tabular}
}
\begin{flushleft} \footnotesize
$^\star$Human-expert translation.  $^\ddagger$Machine translation.
EM refers to Exact Match extraction and MCQ to multiple-choice question answering. \\
\end{flushleft}
\end{table}

\subsection{Instruction-Tuned Model Benchmarking Datasets}
\label{annex:it-eval-datasets}
Table~\ref{tab:instruct-eval} lists the benchmarks and experimental settings used to evaluate our chat models after instruction finetuning (\BYOLC-IT, \BYOLM-IT) and model merging (\BYOLC-M, \BYOLM-M).  We evaluate on sixteen benchmarks spanning seven task categories, plus an additional question-answering benchmark, Multi-Wiki-QA~\cite{MultiWikiQA2025}, which is used only for the LLM-as-a-judge evaluation.
Among the sixteen benchmarks, four are English-only and cover general scientific reasoning (GPQA Diamond~\cite{suzgun2022challenging}, BBH~\cite{suzgun2022challenging}), instruction-following (IFEval~\cite{zhou2023instruction}), and code generation (HumanEval~\cite{chen2021evaluating}). The remaining benchmarks are evaluated in both English and the two target languages, Chichewa and Māori. We also perform a multilingual ablation on \GLOBALMMLU~\cite{singh2024global} across 19 languages.
\begin{table}[!t]
\centering
\caption{\small \underline{ \textbf{Benchmarks used to evaluate our instruct-tuned and merged models.}} The chat template is enabled for all benchmarks.}
\label{tab:instruct-eval}
\adjustbox{width=\textwidth,center}{
\begin{tabular}{lllccc}
\toprule
 \textbf{Benchmark} &  \textbf{Languages} & \textbf{Category} & \textbf{Task} & \textbf{n-shot} & \textbf{Metric} \\
\midrule
\rowcolor{gray!15} \multicolumn{6}{l}{\textbf{General \& STEM reasoning}}\\
\addlinespace[2pt]
Global MMLU-Lite$^\star$~\cite{singh2024global} & 19 languages\footnotemark[1] & General knowledge/Reasoning & MCQ (Gen.) & 0 & Accuracy \\ 
ARC Challenge chat$^\ddagger$~\cite{clark2018think} & Chichewa, Māori  & Science reasoning (hard) & MCQ (Gen.) & 0 & Accuracy (EM)   \\
MGSM$^\ddagger$~\cite{shi2022language} & Chichewa, Māori  & Math reasoning & Generation & 0 & Accuracy (EM) \\
BBH~\cite{suzgun2022challenging} & English  & Complex reasoning & Generation  & 0 & Accuracy (EM) \\
GPQA Diamond~\cite{rein2024gpqa} & English  & Graduate Science & MCQ (Gen.) & 0 & Accuracy (EM) \\ \midrule 
\addlinespace[4pt]
\rowcolor{gray!15} \multicolumn{6}{l}{\textbf{Commonsense \& story}}\\
\addlinespace[2pt]
XCOPA$^\ddagger$~\cite{ponti2020xcopa} & Chichewa, Māori & Causal reasoning & MCQ (LH) & 0 & Accuracy  \\
XStoryCloze$^\ddagger$~\cite{lin2021few} & Chichewa, Māori & Story completion/commonsense & MCQ (LH) & 0 & Accuracy \\
PIQA$^\ddagger$~\cite{bisk2020piqa} & Chichewa, Māori & Physical commonsense & MCQ (LH) & 0 & Accuracy   \\
HellaSwag$^\ddagger$~\cite{zellers2019hellaswag} & Chichewa, Māori & Sentence completion & MCQ (LH) & 0 &  Accuracy  \\ \midrule 
\addlinespace[4pt]
\rowcolor{gray!15} \multicolumn{6}{l}{\textbf{Reading comprehension \& QA}}\\
\addlinespace[2pt]
XNLI 2.0$^\ddagger$~\cite{upadhyay2023xnli} & Chichewa, Māori& Natural language inference & MCQ (LH) & 0 & Accuracy  \\
XWinograd$^\ddagger$~\cite{tikhonov2021s} & Chichewa, Māori & Coreference resolution & MCQ (LH) & 0 & Accuracy   \\
Belebele$^\star$~\cite{bandarkar-etal-2024-belebele} & Māori, Chichewa & Reading comprehension & MCQ (LH) & 0 & Accuracy \\
Multi-Wiki-QA$^\otimes$~\cite{MultiWikiQA2025} & Chichewa, Māori & Reading comprehension/QA & Generation & 0 & \acs{llm}-as-judge\footnotemark[2] win–loss  \\ \midrule 
\addlinespace[4pt]
\rowcolor{gray!15} \multicolumn{6}{l}{\textbf{Instruction following}}\\
\addlinespace[2pt]
IFEval\footnotemark[3]~\cite{zhou2023instruction} & English  & Instruction following & Generation & 0 & Accuracy \\  \midrule 
\addlinespace[4pt]
\rowcolor{gray!15} \multicolumn{6}{l}{\textbf{Truthfulness}}\\
\addlinespace[2pt]
TruthfulQA$^\ddagger$~\cite{lin2021truthfulqa} & Chichewa, Māori  & Truthfulness & MCQ (Gen) & 0 & BLEU Accuracy \\ \midrule 
\addlinespace[4pt]
\rowcolor{gray!15} \multicolumn{6}{l}{\textbf{Code generation}}\\
\addlinespace[2pt]
HumanEval~\cite{chen2021evaluating} & English & Code generation & Generation & 0 & Pass@1 \\  \midrule 
\addlinespace[4pt]
\rowcolor{gray!15} \multicolumn{6}{l}{\textbf{Translation}}\\
\addlinespace[2pt]
 FLORES-200$^\star$~\cite{costa2022no} & Chichewa, Māori & Translation & Generation & 0 & BLEU, chrF++ \\
\bottomrule
\end{tabular}
}
\begin{flushleft}\footnotesize
$^\star$ Human-expert translation. $^\ddagger$ Machine translation. $^\otimes$ LLM-generated questions from Wikipedia documents. \\
EM refers to Exact Match extraction, MCQ to multiple-choice question answering, and LH to likelihood-based scoring. \\
\footnotemark[1] Global-MMLU-Lite~\cite{singh2024global} originally includes 16 languages; in this paper we extend it to 19 by adding three human-expert-translated languages: Chichewa, Māori, and Inuktitut. 
\footnotemark[2] The GPT-5-chat model is used as the judge for win–loss comparisons.
\footnotemark[3] prompt\_level\_loose\_acc score is reported. 
\end{flushleft}
\end{table}
\subsection{LLM as Judge Evaluation}
\label{annex:pairwise-eval}

Following prior work on \emph{LLM-as-a-Judge} evaluation~\cite{ustun-etal-2024-aya,rafailov2023direct,dubois2023alpacafarm}, we design a structured prompt template for pairwise preference assessment in Chichewa and Māori. We instantiate this protocol on a question-answering task using 1{,}000 randomly selected samples from the Multi-Wiki-QA dataset~\cite{MultiWikiQA2025}. In each comparison, the judge model receives the context, question, a reference answer, and two candidate responses presented in randomized order. It is instructed to (i) produce a comparative rationale evaluating linguistic correctness, instruction adherence, factual accuracy, semantic comprehension, and grammatical fluency in the target language; (ii) select a preferred response label from \{A, B\}; and (iii) assign individual quality scores (0--5) to each candidate according to these criteria. The full prompt template used for this evaluation is provided in Annex~\ref{subsec:llm-as-judge-template}.  We compare our models against the following competing LLMs: \GEMMA~(4B/12B/27B-IT)~\cite{team2025gemma}, \OSS~(120B)~\cite{openai2025gptoss120bgptoss20bmodel}, \APERTUS~(8B-Instruct-2509)~\cite{hernandez2025apertus}, and GPT-4o. 
GPT-5-chat is used exclusively as the evaluation judge and is not included as a competing model.

\begin{table}[!t]
\centering
\caption{\small\underline{\textbf{Base model results on English}} benchmarks. After continual pretraining of Gemma-3 baseline, our \BYOL models preserve English language performance. Results on Chichewa (\texttt{nya}) and Māori (\texttt{mri}) benchmarks are provided in Tables~\ref{tab:pre-train-resl-nya} and~\ref{tab:pre-train-resl-mri}.}
\label{tab:pre-train-eng}
\adjustbox{width=\textwidth,center}{
\begin{tabular}{l|ccc|ccc|ccc}
\toprule[1.2pt]
\multirow{2}{*}{\textbf{Benchmarks}}  & 
\multicolumn{3}{c|}{\textbf{1B}} & 
\multicolumn{3}{c|}{\textbf{4B}} & 
\multicolumn{3}{c}{\textbf{12B}} \\ 
\cmidrule(lr){2-4} \cmidrule(lr){5-7} \cmidrule(lr){8-10}
&  
\textbf{Gemma-3}~\cite{team2025gemma} & \textbf{\BYOLC} & \textbf{\BYOLM} & \textbf{Gemma-3}~\cite{team2025gemma} & \textbf{\BYOLC} & \textbf{\BYOLM} & \textbf{Gemma-3}~\cite{team2025gemma} & \textbf{\BYOLC} & \textbf{\BYOLM} \\ 
&
\texttt{(PT)} & \texttt{(CPT)} & \texttt{(CPT)} & \texttt{(PT)} & \texttt{(CPT)} & \texttt{(CPT)}  & \texttt{(PT)} & \texttt{(CPT)} & \texttt{(CPT)} \\  
\midrule[1.2pt]
Global MMLU-Lite~\cite{singh2024global} & 25.00 & 23.50 & 23.25 & 66.75 & 65.50 & 64.25 & 76.50 & 76.75 & 76.25 \\
ARC-Easy~\cite{clark2018think} & 71.97 & 68.31 & 68.22 & 81.78 & 77.61 & 75.93 & 87.79 & 86.45 & 85.69 \\
ARC-Hard~\cite{clark2018think} & 39.51 & 40.36 & 39.25 & 58.36 & 58.28 & 59.64 & 67.83 & 68.17 & 68.00 \\
MGSM~\cite{shi2022language} & 3.60 & 1.20 & 2.80 & 46.40 & 52.40 & 50.40 & 76.80 & 80.80 & 82.00 \\
BBH~\cite{suzgun2022challenging} & 28.35 & 26.74 & 27.83 & 38.46 & 37.00 & 37.58 & 52.39 & 51.08 & 52.39 \\
GPQA Diamond~\cite{rein2024gpqa}  & 22.73 & 24.75 & 23.74 & 30.30 & 36.36 & 32.32 & 35.35 & 32.32 & 35.86 \\
\midrule
XCOPA~\cite{ponti2020xcopa}  &  81.00 & 78.00 & 79.00 & 88.00 & 89.00 & 86.00 & 93.00 & 94.00 & 94.00 \\
XStoryCloze~\cite{lin2021few} & 72.20 & 71.87 & 72.20 & 80.08 & 80.87 & 81.07 & 84.18 & 83.59 & 83.98 \\
PIQA~\cite{bisk2020piqa} & 75.03 & 73.39 & 73.56 & 79.71 & 79.05 & 78.94 & 81.83 & 81.72 & 81.34 \\
HellaSwag~\cite{zellers2019hellaswag} & 62.93 & 62.06 & 62.39 & 77.71 & 77.73 & 77.81 & 84.11 & 83.21 & 83.43 \\
\midrule
XNLI 2.0~\cite{upadhyay2023xnli} & 48.31 & 48.76 & 48.39 & 51.00 & 49.88 & 51.45 & 53.98 & 52.45 & 53.61 \\
Winograd~\cite{tikhonov2021s} & 58.80 & 58.33 & 58.25 & 69.46 & 67.72 & 68.11 & 75.22 & 74.03 & 74.35 \\
Belebele~\cite{bandarkar-etal-2024-belebele} & 27.11 & 27.33 & 28.00 & 79.22 & 77.33 & 75.44 & 92.22 & 92.22 & 92.67 \\
\midrule[1.2pt]
\bf Average Score & 47.43 &  46.51 & 46.68 &  65.17 & 65.29 & 64.53 & 73.94 & 73.60 & 74.12 \\
\bottomrule[1.2pt]
\end{tabular}
}
\end{table}

\begin{table}[!t]
\centering
\caption{\small\underline{\textbf{Instruction-Tuned model results on English}} benchmarks. 
Results on Chichewa (\texttt{nya}) and Māori (\texttt{mri}) benchmarks are provided in Tables~\ref{tab:chat-resl-nya} and~\ref{tab:chat-resl-mri}.}
\label{tab:chat-eng}
\adjustbox{width=0.80\textwidth,center}{
\begin{tabular}{l|ccc|ccc}
\toprule
\multicolumn{1}{l|}{\multirow{3}{*}{Benchmarks}}  & 
\multicolumn{3}{c|}{\textbf{4B}} & 
\multicolumn{3}{c}{\textbf{12B}} \\ 
\cmidrule(lr){2-4} \cmidrule(lr){5-7} 
&  
\textbf{Gemma-3}~\cite{team2025gemma} & \textbf{\BYOLC} & \textbf{\BYOLM} & \textbf{Gemma-3}~\cite{team2025gemma} & \textbf{\BYOLC} & \textbf{\BYOLM} \\ 
&
\texttt{(IT)} & \texttt{(M)} & \texttt{(M)} & \texttt{(IT)} & \texttt{(M)} & \texttt{(M)}  \\  
\midrule[1.2pt]
Global MMLU-Lite~\cite{singh2024global}  & 66.21 &  64.47 & 66.30  & 75.15 & 81.71  & 81.00 \\
 ARC-Hard~\cite{clark2018think}   chat & 75.85 &  75.77 & 76.11 & 90.10 & 88.99 & 88.82 \\
 MGSM~\cite{shi2022language} & 47.20 &  57.60 & 60.00  & 58.00 &  75.20 & 76.00 \\
 BBH~\cite{suzgun2022challenging} & 50.44 & 44.16 & 44.40  & 56.27 & 57.58 & 57.40 \\
 GPQA Diamond~\cite{rein2024gpqa}  & 33.84 &  34.85 & 26.26 & 36.36 &  37.37 & 37.37 \\
 \midrule 
XCOPA~\cite{ponti2020xcopa}  &  86.00 &  89.00 & 88.00  & 88.00 & 95.00 & 95.00 \\
XStoryCloze~\cite{lin2021few} & 65.45 & 72.47 & 73.40  & 69.09 & 77.37 & 76.70  \\
PIQA~\cite{bisk2020piqa} & 69.91 &  78.13 &  78.56 & 71.16 & 81.83 & 81.45 \\
HellaSwag~\cite{zellers2019hellaswag} & 47.47 &  71.85 & 70.90 & 55.70 &  77.32  & 77.17 \\
\midrule 
XNLI 2.0~\cite{upadhyay2023xnli} & 43.41 & 54.42 &  53.61 & 50.36 & 56.06  & 54.38 \\
XWinograd~\cite{tikhonov2021s} &61.25 & 67.56 & 67.72 & 65.98 &  74.35  & 74.59 \\
Belebele~\cite{bandarkar-etal-2024-belebele} & 69.67 &  87.11 &  83.22 & 91.33 & 92.67 & 92.56 \\
\midrule
truthfulqa-multi gen~\cite{lin2021truthfulqa} & 42.96 & 47.12 & 49.33 & 49.45 & 49.94 & 46.14 \\ \midrule
IFEval~\cite{zhou2023instruction} & 76.89 &  79.11 &75.97 & 75.23 &  85.77 & 86.51 \\ \midrule
HumanEval~\cite{chen2021evaluating}  & 68.90 & 55.49 & 53.66 & 80.04 & 78.66 & 79.27 \\\midrule[1.2pt]
\bf Average Score & 60.36 & 65.27 & 64.50  & 67.87  & 73.99 & 73.62 \\
\bottomrule
\end{tabular}
}
\end{table}

\section{Performance of BYOL Models on English}
\label{annex:eng-perf-cpt}
Tables~\ref{tab:pre-train-eng} and~\ref{tab:chat-eng} show that our LRL-specific \BYOL models maintain English performance after adaptation and remain comparable to the Gemma-3 baseline.

\section{Ablation Experiment Results}
\label{annex:ablations}

\subsection{RTTBench-Mono Validation}
\label{app:rttbench}
\begin{figure}[h]
\centering
\includegraphics[width=0.8\linewidth]{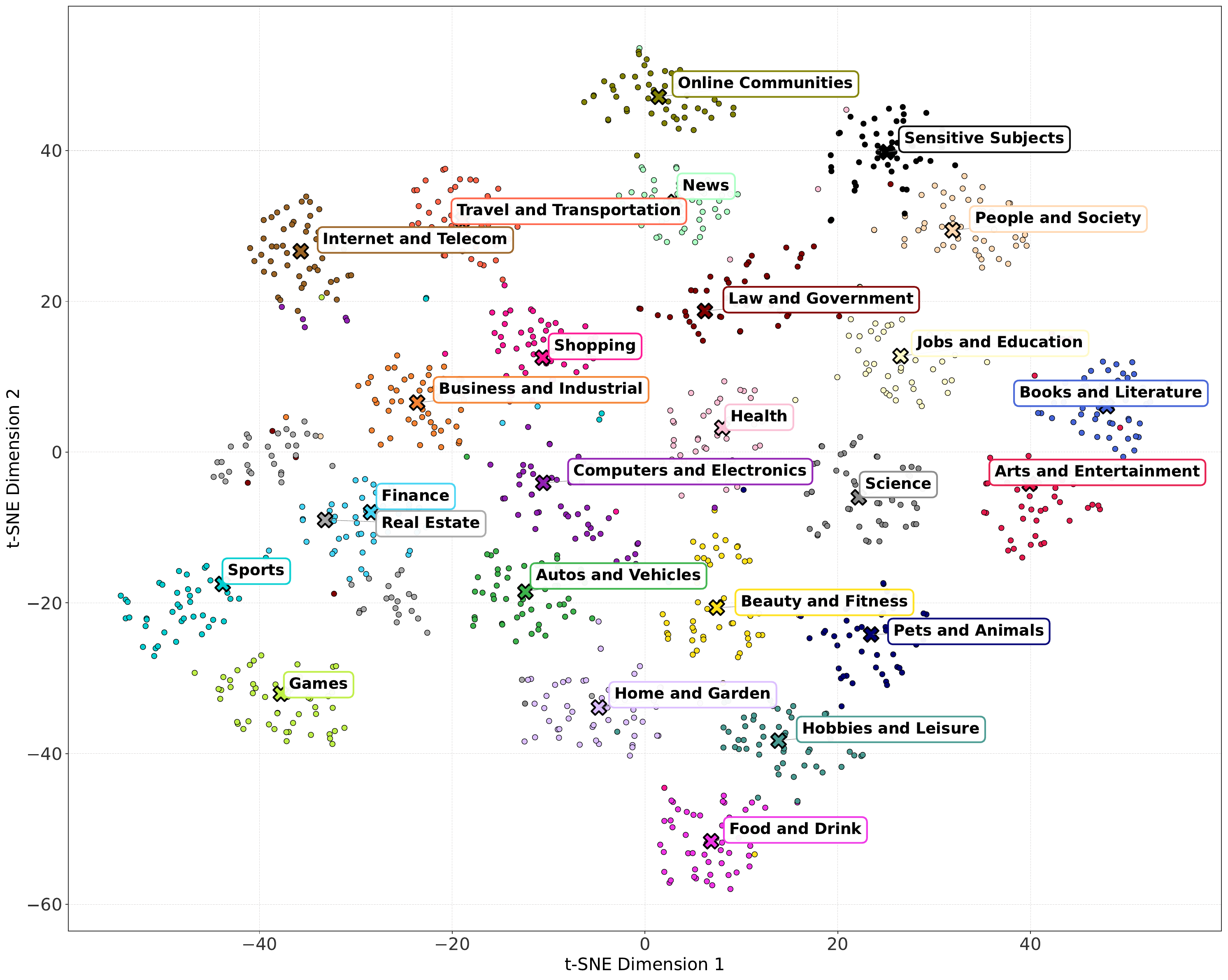}
\vspace{-1em}
\caption{\small \textbf{\underline{t-SNE visualization of RTTBench-Mono embeddings.}} Sentences from all 25 domains form well-separated clusters, indicating strong domain fidelity and minimal cross-domain overlap.}
\label{fig:RTT_bench_clustering}
\end{figure}

The prompt tempelate to construct RTTBench-Mono dataset is provided in ~\ref{app:prompt-rttbench}. To reduce semantic drift between overlapping categories (e.g., {Health} vs.~{Beauty \& Fitness}), each prompt includes the domain definition and a list of confusable categories to avoid during data generation 
We perform dataset quality verification using two complementary approaches. First, we apply NVIDIA’s domain classifier on RTTBench-Mono, which achieves 97.8\% Top-1 accuracy, confirming strong alignment between each sentence and its intended domain. Second, we embed all sentences using \texttt{Azure OpenAI text-embedding-3-large} model and visualize them via t-SNE. Figure~\ref{fig:RTT_bench_clustering} shows that clusters from all 25 domains are well separated, indicating that RTTBench-Mono maintains distinct domains and avoids significant cross-domain overlap.

\subsection{Evaluation of MTs on RTT Benchmarks}
\label{annex:per-domain-mt-results}
\begin{figure}[!t]
    \centering

    \subfloat{%
        \includegraphics[width=0.49\linewidth]{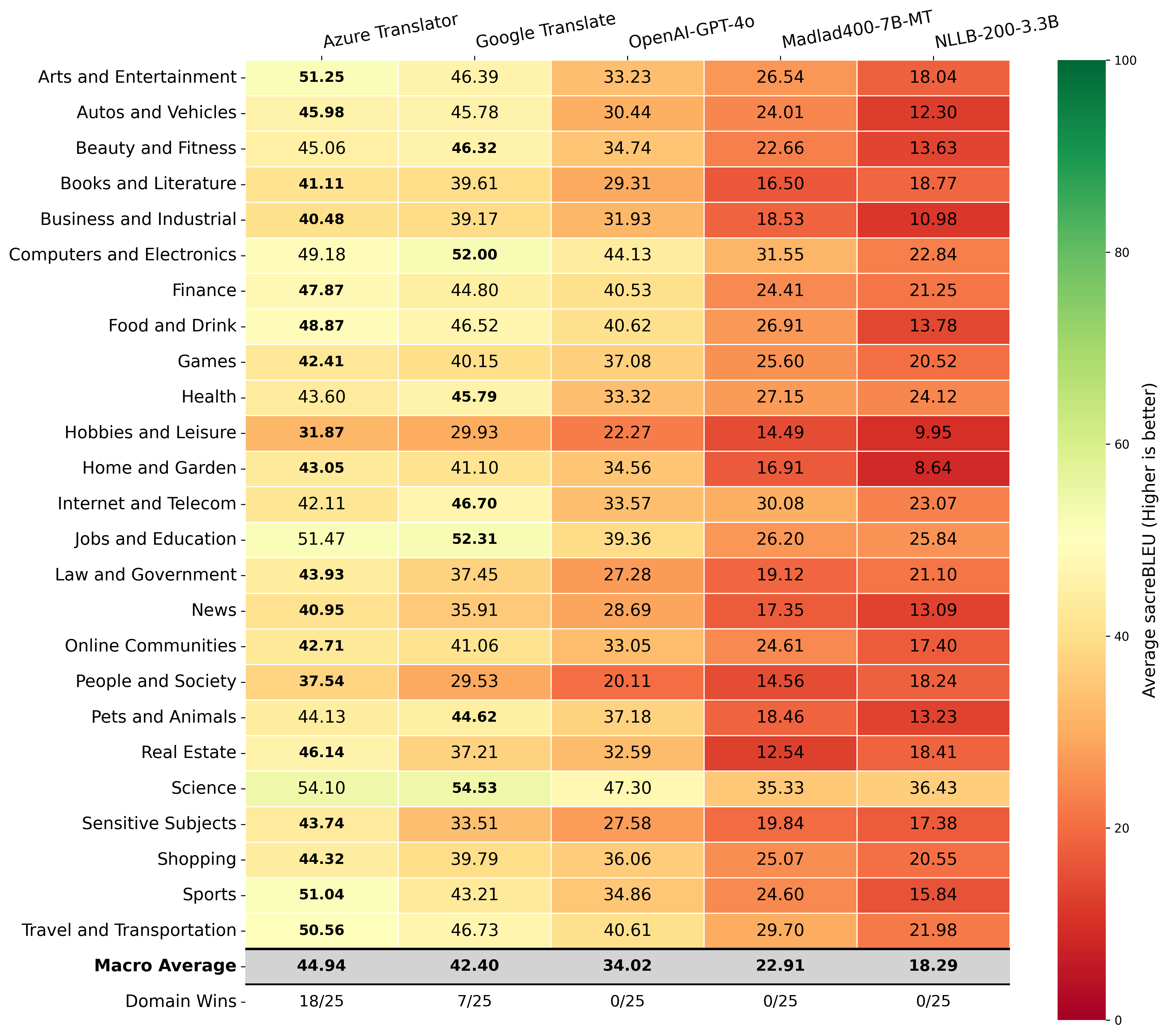}
    }
    \hfill
    \subfloat{%
        \includegraphics[width=0.49\linewidth]{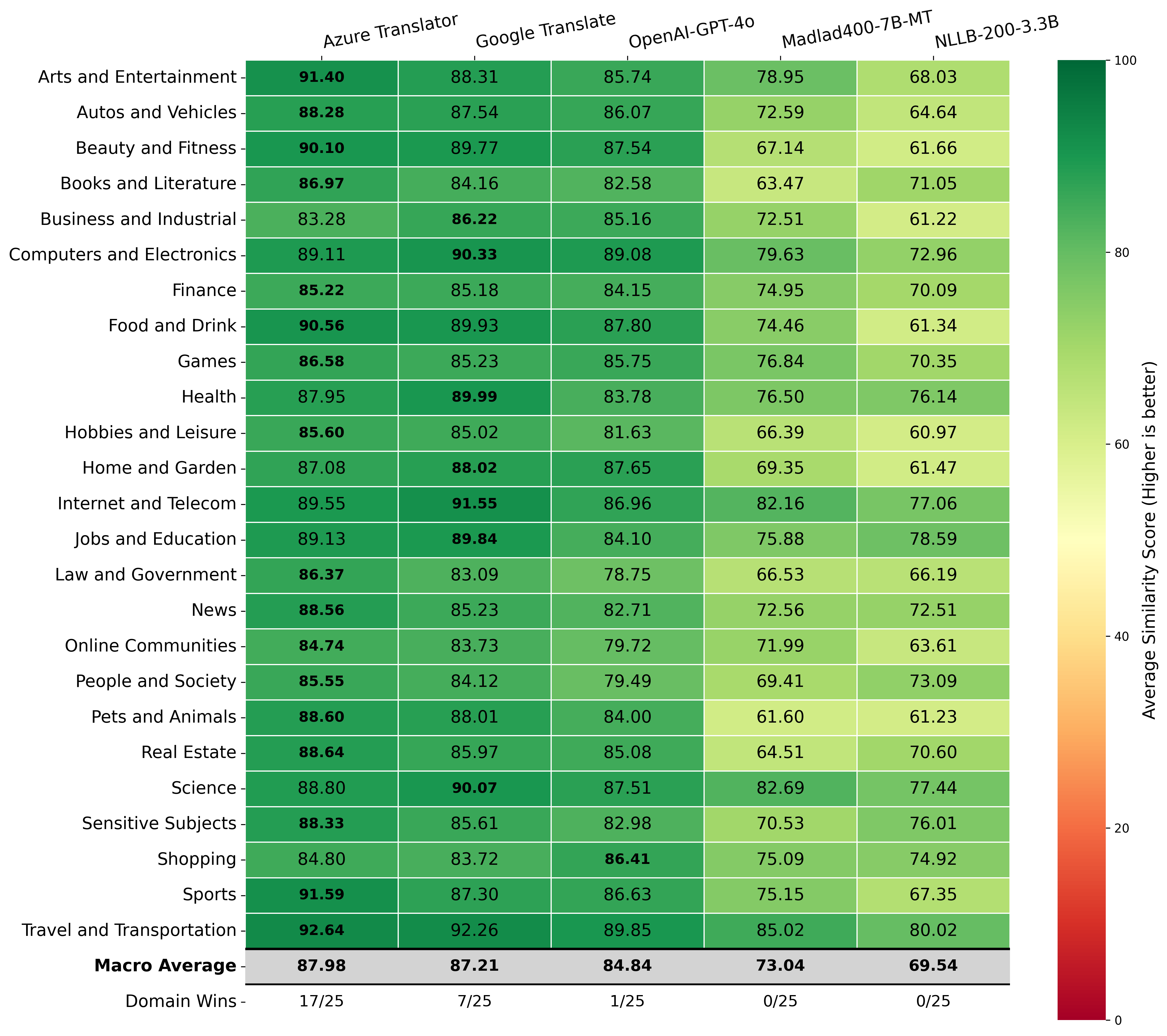}
    }

    \vspace{-0.5em}
    \caption{\small \underline{\textbf{Ablation of per-domain MT evaluation on RTTBench-Mono dataset.}}
    We use the round-trip translation setup (Sec.~\ref{sec:tool-assessment}), where English sentences from 25 domains are translated into Chichewa and back to English, and the reconstructed sentences are compared to the originals.
    \textbf{Left:} per-domain sacreBLEU scores.
    \textbf{Right:} per-domain embedding-similarity scores.
    Azure Translator achieves the highest macro-average and the most domain wins.
    }
    \label{fig:init-assess-mt-per-domain-maps}
\end{figure}

Figure~\ref{fig:init-assess-mt-per-domain-maps} presents a per-domain MT ablation on the RTTBench-Mono dataset using our round-trip translation setup (Sec.~\ref{sec:tool-assessment}), in which English sentences from 25 domains are translated into Chichewa and back to English, and the reconstructed sentences are compared to the originals. The left panel reports per-domain sacreBLEU scores, while the right panel reports per-domain embedding-similarity scores. Across both metrics, Azure Translator achieves the highest macro-average performance and the largest number of domain wins.

\subsection{Impact of CPT Data Mixture}
Table~\ref{tab:pre-train-ablation-sets-full} reports impact of various CPT data mixtures on the performance of \BYOLC.

\subsection{Ablation of LoRA vs. Full-Parameter}
Table~\ref{tab:pre-train-lora-ablation-full} shows results on the LoRA vs.\ full-parameter continual pretraining for \BYOLC (CPT-4B).

\subsection{Impact of Model Merging Weight}
Tables~\ref{tab:lambda-ablation-chi} and~\ref{tab:lambda-ablation-eng} present detailed results on the impact of the weight-merging parameter $\lambda$ across 12 Chichewa benchmarks and 15 English benchmarks, respectively.

\begin{table}[t]
\centering
\captionof{table}{\small \underline{\textbf{Data mixture ablation.}}
Continual pretraining of the  \BYOLC (CPT-4B) model under different data mixtures. For the definitions of the C1--C4 mixtures, see Table~\ref{tab:pretrain_datasets}. \GEMMA is the baseline.}
\label{tab:pre-train-ablation-sets-full}
\adjustbox{width=0.9\textwidth,center}{
\begin{tabular}{llcc|cccccccc}
\toprule[1.2pt]

\multirow{3}{*}{\textbf{Model}} 
& & 
\multicolumn{2}{c}{\multirow{2}{*}{\shortstack{\makecell{\textbf{Gemma-3}~\cite{team2025gemma}\\\texttt{(4B-PT)}}}}} 
& \multicolumn{8}{c}{\textbf{Data Mixture Configurations}} 
\\
\cmidrule(lr){5-12}

&  & & 
& \multicolumn{2}{c}{\textbf{C1}} 
& \multicolumn{2}{c}{\textbf{C2}}  
& \multicolumn{2}{c}{\textbf{C3}}     
& \multicolumn{2}{c}{\textbf{C4}} 
\\  
\cmidrule(lr){3-12}

&  & \texttt{eng} & \texttt{nya} 
& \texttt{eng} & \texttt{nya} 
& \texttt{eng} & \texttt{nya} 
& \texttt{eng} & \texttt{nya} 
& \texttt{eng} & \texttt{nya}  
\\ 

\midrule[1.2pt]

Global MMLU-Lite~\cite{singh2024global} & & 66.75 & 50.75 & 64.00 & 54.00 & 63.25 & 52.50 & 65.75 & 52.50 & 65.50 & 55.25 \\
ARC-Easy~\cite{clark2018think}  & & 81.78 & 30.22 & 80.68 & 42.26 & 79.76 & 41.96 & 79.53 & 42.47 & 77.61 & 48.48 \\
ARC-Hard~\cite{clark2018think} & & 58.36 & 27.13 & 59.04 & 36.35 & 59.81 & 37.63 & 59.90 & 37.88 & 58.28 & 40.61 \\
MGSM~\cite{shi2022language} & & 46.40 & 17.20 & 45.20 & 26.80 & 50.40 & 26.80 & 52.40 & 26.00 & 52.40 & 31.60 \\
BBH~\cite{suzgun2022challenging}  & & 38.46 & - & 37.03 & - & 37.49 & - & 37.29 & - & 37.00 & - \\
GPQA Diamond~\cite{rein2024gpqa} & & 30.30 & - & 30.30 & - & 32.32 & - & 32.32 & - & 36.36 & - \\
\midrule
XCOPA~\cite{ponti2020xcopa}  & & 88.00 & 57.20 & 88.00 & 64.40 & 89.00 & 65.60 & 89.00 & 68.00 & 89.00 & 70.00 \\
XStoryCloze~\cite{lin2021few} & & 80.08 & 54.40 & 80.61 & 63.73 & 80.68 & 64.66 & 80.81 & 64.73 & 80.87 & 65.98 \\
PIQA~\cite{bisk2020piqa}  & & 79.71 & 54.57 & 80.41 & 61.15 & 79.54 & 63.33 & 79.11 & 62.35 & 79.05 & 63.71 \\
HellaSwag~\cite{zellers2019hellaswag}  & & 77.71 & 33.45 & 78.73 & 43.61 & 78.09 & 44.80 & 77.71 & 45.59 & 77.73 & 47.31 \\
\midrule 
XNLI 2.0~\cite{upadhyay2023xnli}  & & 51.00 & 37.82 & 51.61 & 39.58 & 51.53 & 40.92 & 51.16 & 41.10 & 49.88 & 40.32 \\
XWinograd~\cite{tikhonov2021s} & & 69.46 & 54.76 & 68.27 & 66.10 & 69.53 & 68.02 & 67.96 & 66.10 & 67.72 & 68.34 \\
Belebele~\cite{bandarkar-etal-2024-belebele} & & 79.22 & 38.22 & 75.67 & 43.67 & 76.67 & 42.44 & 76.89 & 43.67 & 77.33 & 45.44 \\
\midrule
\multirow{2}{*}{FLORES~\cite{costa2022no} (\texttt{nya}$\rightarrow$\texttt{eng})}
& BLEU & - & 17.28 & - & 23.22 & - & 22.82 & - & 22.93 & - & 23.87 \\
& chrF++ & - & 40.37 & - & 47.28 & - & 46.78 & - & 47.10 & - & 47.95 \\
\multirow{2}{*}{FLORES~\cite{costa2022no} (\texttt{eng}$\rightarrow$\texttt{nya})}
& BLEU & - & 2.24 & - & 11.42 & - & 11.72 & - & 11.86 & - & 12.79 \\
& chrF++ & - & 23.22 & - & 45.11 & - & 47.29 & - & 47.32 & - & 48.66 \\
\midrule[1.2pt]

\bf Average Score & & 65.17 & 39.95 & 64.58 & 48.77 & 65.24 & 49.44 & 65.37 & 49.60 & 65.29 & 51.82 \\

\bottomrule[1.2pt]
\end{tabular}
}
\end{table}

{}
\begin{table}[t]
\centering
\caption{\small \underline{\textbf{Ablation of LoRA vs. full-parameter}} continual pre-training  of \BYOLC (CPT-4B). The accuracy of LoRA model improves with increasing rank but remains below full-parameter tuning (51.82). \GEMMA is the baseline.}
\label{tab:pre-train-lora-ablation-full}
\adjustbox{width=\textwidth,center}{
\begin{tabular}{llcc|cccccccc|cc}
\toprule[1.2pt]

\multirow{3}{*}{\textbf{Model}} 
& & 
\multicolumn{2}{c}{\multirow{2}{*}{\makecell{\textbf{Gemma-3}~\cite{team2025gemma}\\\texttt{(4B-PT)}}}}
& \multicolumn{8}{c}{\textbf{LoRA Rank}}
& \multicolumn{2}{c}{\multirow{2}{*}{\makecell{\textbf{Full-Parameter}\\\textbf{Tuning}}}}
\\
\cmidrule(lr){5-12}

& & &
& \multicolumn{2}{c}{\textbf{64}}
& \multicolumn{2}{c}{\textbf{128}}
& \multicolumn{2}{c}{\textbf{256}}
& \multicolumn{2}{c}{\textbf{512}}
& & 
\\
\cmidrule(lr){3-14}

&  & \texttt{eng} & \texttt{nya}
& \texttt{eng} & \texttt{nya}
& \texttt{eng} & \texttt{nya}
& \texttt{eng} & \texttt{nya}
& \texttt{eng} & \texttt{nya}
& \texttt{eng} & \texttt{nya}
\\

\midrule[1.2pt]


Global MMLU-Lite~\cite{singh2024global} && 
66.75 & 50.75 & 64.25 & 51.25 & 65.25 & 55.25 & 64.50 & 55.25 & 64.75 & 57.00 & 65.50 & 55.25 \\

ARC-Easy~\cite{clark2018think} &&
81.78 & 30.22 &
79.08 & 34.43 & 76.47 & 38.89 & 76.68 & 39.98 & 76.18 & 43.69 &
77.61 & 48.48 \\

ARC-Hard~\cite{clark2018think} &&
58.36 & 27.13 &
59.39 & 31.91 & 58.53 & 33.53 & 58.19 & 34.39 & 57.85 & 36.09 &
58.28 & 40.61 \\

MGSM~\cite{shi2022language} &&
46.40 & 17.20 &
52.00 & 22.40 & 50.40 & 24.80 & 51.60 & 28.40 & 53.60 & 27.60 &
52.40 & 31.60 \\

BBH~\cite{suzgun2022challenging} &&
38.46 & - &
40.45 & - & 40.44 & - & 39.47 & - & 38.06 & - &
37.00 & - \\

GPQA Diamond~\cite{rein2024gpqa} &&
30.30 & - &
27.78 & - & 33.84 & - & 31.31 & - & 30.81 & - &
36.36 & - \\

\midrule

XCOPA~\cite{ponti2020xcopa} &&
88.00 & 57.20 &
88.00 & 60.00 & 87.00 & 64.40 & 87.00 & 64.80 & 87.00 & 66.20 &
89.00 & 70.00 \\

XStoryCloze~\cite{lin2021few} &&
80.08 & 54.40 &
81.14 & 58.17 & 80.21 & 60.03 & 79.88 & 60.09 & 80.41 & 63.47 &
80.87 & 65.98 \\

PIQA~\cite{bisk2020piqa} &&
79.71 & 54.57 &
80.09 & 55.82 & 79.65 & 59.68 & 79.65 & 59.30 & 79.22 & 61.21 &
79.05 & 63.71 \\

HellaSwag~\cite{zellers2019hellaswag} &&
77.71 & 33.45 &
77.16 & 36.32 & 76.40 & 40.92 & 76.26 & 41.46 & 76.31 & 43.79 &
77.73 & 47.31 \\

\midrule

XNLI 2.0~\cite{upadhyay2023xnli} &&
51.00 & 37.82 &
53.33 & 38.42 & 51.61 & 38.96 & 51.53 & 39.78 & 50.60 & 40.60 &
49.88 & 40.32 \\

XWinograd~\cite{tikhonov2021s} &&
69.46 & 54.76 &
69.46 & 60.64 & 69.93 & 65.24 & 70.17 & 66.63 & 69.69 & 68.02 &
67.72 & 68.34 \\

Belebele~\cite{bandarkar-etal-2024-belebele} &&
79.22 & 38.22 &
79.22 & 38.11 & 79.89 & 40.44 & 80.67 & 42.11 & 80.00 & 42.78 &
77.33 & 45.44 \\

\midrule

\multirow{2}{*}{FLORES~\cite{costa2022no} (\texttt{nya}$\rightarrow$\texttt{eng})}
& BLEU & - & 17.28 &
- & 19.76 & - & 21.90 & - & 22.60 & - & 24.22 &
- & 23.87 \\
& chrF++& - & 40.37 &
- & 43.22 & - & 45.91 & - & 46.80 & - & 48.03 &
- & 47.95 \\

\multirow{2}{*}{FLORES~\cite{costa2022no} (\texttt{eng}$\rightarrow$\texttt{nya})}
& BLEU & - & 2.24 &
- & 6.51 & - & 10.13 & - & 10.22 & - & 11.33 &
- & 12.79 \\
& chrF++& - & 23.22 &
- & 35.99 & - & 43.39 & - & 43.74 & - & 46.32 &
- & 48.66 \\

\midrule[1.2pt]

\bf Average Score && 
65.17 & 39.95 &
65.49 & 43.59 & 65.36 & 47.03 & 65.15 & 47.90 & 64.96 & 49.60 &
65.29 & 51.82 \\

\bottomrule[1.2pt]
\end{tabular}
}
\end{table}

{}

\begin{table}[ht]
\centering
\caption{\small {\bf \underline{Effect of model-merging weight $\lambda$ evaluated on Chichewa benchmarks.}} 
$\lambda = 0$ corresponds to the baseline model \GEMMA (4B-IT), and $\lambda = 1$ to our instruction-tuned Chichewa model \BYOLC (4B-IT).}
\label{tab:lambda-ablation-chi}

\adjustbox{width=\textwidth,center}{
\begin{tabular}{llcccccccccccc}
\toprule
\multirow{3}{*}{\textbf{Model}} & & \multicolumn{11}{c}{\textbf{Merging weight parameter ($\lambda$)}}   \\ \cmidrule(lr){3-13} 
&  & 0 & 0.1 & 0.2 & 0.3 & 0.4 & 0.5 & 0.6 & 0.7 & 0.8 & 0.9 & 1   \\ 
\midrule
Global MMLU-Lite~\cite{singh2024global}  & &45.36  & 48.86 & 47.73 & 50.96 & 53.43 & 53.79 & 53.62 & 54.53 & 51.46 & 50.66 & 48.72   \\ 
ARC-Hard chat~\cite{clark2018think}  & & 33.28 & 35.67 & 39.25 & 43.69 & 46.08 & 48.21 & 50.43 & 51.45 & 51.62 & 51.19 & 49.57    \\ 
MGSM~\cite{shi2022language} & & 	11.20 & 22.00 & 21.20 & 27.20 & 28.80 & 26.40 & 30.00 & 30.00 & 27.20 & 19.60 & 16.00   \\ 
 \midrule 
XCOPA~\cite{ponti2020xcopa} &  & 52.20 & 50.60 & 52.00 & 55.40 & 59.40 & 65.40 & 66.40 & 65.80 & 67.60 & 68.80 & 70.80   \\ 
XStoryCloze~\cite{lin2021few} & & 49.31  & 51.56 & 53.47 & 54.60 & 56.12 & 57.71 & 59.23 & 60.82 & 60.62 & 60.42 & 60.03   \\ 
PIQA~\cite{bisk2020piqa}   & & 	52.77& 52.29 & 54.30 & 56.69 & 58.54 & 61.10 & 61.75 & 62.68 & 62.73 & 63.11 & 63.22    \\ 
HellaSwag~\cite{zellers2019hellaswag} & & 	29.10 & 31.25 & 34.15 & 37.29 & 40.52 & 43.30 & 45.32 & 47.08 & 47.54 & 47.62 & 47.21    \\ 
\midrule 
XNLI 2.0~\cite{upadhyay2023xnli}  & & 35.75 & 36.27 & 37.50 & 37.43 & 38.30 & 38.64 & 39.18 & 40.40 & 40.14 & 40.12 & 39.92   \\ 
XWinograd~\cite{tikhonov2021s} & & 52.41 & 51.44 & 54.33 & 56.47 & 60.64 & 61.82 & 66.42 & 68.98 & 68.77 & 70.16 & 71.02   \\ 
Belebele~\cite{bandarkar-etal-2024-belebele}  & &  29.00 & 32.78 & 37.22 & 43.11 & 49.78 & 52.11 & 55.00 & 54.22 & 53.56 & 53.33 & 53.33    \\  \midrule 
\multirow{2}{*}{ FLORES~\cite{costa2022no}  (\texttt{nya}$\rightarrow$\texttt{eng})}  & BLEU  &11.97 & 14.95 & 18.36 & 21.08 & 22.80 & 24.32 & 24.96 & 25.33 & 25.49 & 25.49 & 24.63    \\ 
& chrF++  &35.26 & 39.09 & 42.46 & 45.03 & 47.02 & 48.46 & 49.21 &49.86 & 49.94 & 49.81 & 48.93    \\ 
\multirow{2}{*}{  FLORES~\cite{costa2022no}   (\texttt{eng}$\rightarrow$\texttt{nya}) } & BLEU & \phantom{0}2.80  & \phantom{0}4.51 & \phantom{0}6.49 &\phantom{0}9.03 & 10.68 & 12.39 & 13.31 & 13.83 & 13.99  &14.21 & 13.87   \\ 
& chrF++  & 	25.38 & 31.05 & 36.51 & 41.76 & 45.14 & 47.62 & 48.91 & 49.70 & 49.89 & 50.33 & 50.02    \\  \midrule 
TruthfulQA~\cite{lin2021truthfulqa}  & & 28.76  & 30.23 & 32.07 & 29.13 & 34.27 & 38.80 & 36.11 & 36.96 & 37.33 & 37.09 & 37.45  \\  
\midrule
\bf Average Score & & 36.91 & 39.47 & 41.71 & 44.52 & 47.54 & 49.49 & 50.89 & 51.73 & 51.42 & 50.94 & 50.48  \\ 
\bottomrule
\end{tabular}
}
\end{table}

{}
\begin{table}[h]
\centering
\caption{\small {\bf \underline{Effect of model-merging weight $\lambda$ evaluated on English benchmarks.}} 
$\lambda = 0$ corresponds to the baseline model \GEMMA (4B-IT), and $\lambda = 1$ to our instruction-tuned Chichewa model \BYOLC (4B-IT).}
\label{tab:lambda-ablation-eng}
\adjustbox{width=\textwidth,center}{
\begin{tabular}{llccccccccccc}
\toprule
\multirow{3}{*}{\textbf{Model}}  & \multicolumn{11}{c}{\textbf{Merging weight parameter ($\lambda$)}}   \\ \cmidrule(lr){2-12} 
&   0 & 0.1 & 0.2 & 0.3 & 0.4 & 0.5 & 0.6 & 0.7 & 0.8 & 0.9 & 1       \\ 
\midrule
Global MMLU-Lite~\cite{singh2024global}  & 	 66.21 & 68.23 & 68.58 & 66.75 & 66.80 & 65.43 & 64.47 & 64.55 & 63.60 & 63.45 & 63.06  \\ 
 ARC-Hard  chat~\cite{clark2018think} & 	75.85 & 76.88 & 77.56 & 77.99 & 77.99 & 75.68 & 75.77 & 73.46 & 71.42 & 68.94 & 66.21  \\ 
 MGSM~\cite{shi2022language} & 47.20  & 57.60 & 64.80 & 68.00 &  63.20 & 60.80 & 57.60 & 53.20 & 49.60 & 48.00 & 38.80   \\ 
 BBH~\cite{suzgun2022challenging} & 	50.44 & 51.39 & 50.88 & 49.58 & 48.27 & 45.43 & 44.16 & 42.87 & 42.10 & 40.64 & 38.70    \\ 
 GPQA Diamond~\cite{rein2024gpqa}  & 	33.84 & 35.35 & 34.34 & 33.84 & 33.84 & 35.86 & 34.85 & 32.32 & 32.32 & 34.34 & 34.85   \\ 
 \midrule 
XCOPA~\cite{ponti2020xcopa}  & 	86.00 & 89.00 & 89.00 & 90.00 & 89.00 & 89.00 & 89.00 & 89.00 & 89.00 & 89.00 & 88.00  \\ 
XStoryCloze~\cite{lin2021few}  & 	65.45 & 67.31 & 69.03 & 70.42 & 71.54 & 71.94 & 72.47 & 73.20 & 72.60 & 71.94 & 70.62   \\ 
PIQA~\cite{bisk2020piqa} & 	69.91  &72.14 & 73.56 & 75.35 & 76.93 & 77.42 & 78.13 & 79.11 & 78.78 & 79.05 & 78.45  \\ 
HellaSwag~\cite{zellers2019hellaswag}  & 	 47.47& 51.89 & 57.20 & 62.80 & 67.39 & 69.99 & 71.85 & 72.73 & 72.87 & 72.44 & 71.90  \\ 
\midrule 
XNLI 2.0~\cite{upadhyay2023xnli} & 43.41 &	47.47 & 49.24 & 50.92 & 52.61 & 53.86 & 54.42 & 53.53 & 53.13 & 52.25 & 51.29  \\ 
XWinograd~\cite{tikhonov2021s} & 	61.25 & 61.88 & 65.35 & 67.40 & 66.93 & 67.09 & 67.56 & 67.32 & 68.82 & 68.67 & 67.64    \\ 
Belebele~\cite{bandarkar-etal-2024-belebele} & 	69.67 & 77.78 & 84.11 & 87.11 & 87.22 & 86.89 & 87.11 & 86.00 & 84.89 & 84.11 & 81.67    \\ 
\midrule
TruthfulQA~\cite{lin2021truthfulqa}  & 	 42.96  & 43.08 & 42.72 & 46.63 & 46.27 & 48.23 & 47.12 & 43.70 & 43.94 & 43.45 & 41.74  \\ \midrule 
IFEval~\cite{zhou2023instruction} & 	76.89 & 81.52 & 83.55 & 83.73 & 82.62 & 80.78 & 79.11 & 72.64 & 68.76 & 66.73 & 64.70   \\ \midrule 
Humaneval~\cite{chen2021evaluating}  & 	 68.90 & 68.29 & 66.46 & 65.24 & 64.63 & 59.76 & 55.49 & 53.66 & 51.22 & 45.73 & 43.29  \\ \midrule  
\bf Average Score & 60.36 & 63.32 & 65.09 & 66.38 & 66.35 & 65.88 & 65.27 & 63.82 & 62.87 & 61.92 & 60.06  \\ 
\bottomrule
\end{tabular}
}
\end{table}

{}

\clearpage
\section{Prompts}
\subsection{Prompt for Text Refinement}
\label{app:data-refinement-prompt}
\begin{AIbox}{}
You are an expert content editor and enhancer. You will receive text primarily in \{\texttt{LANGUAGE\_NAME}\} (ISO639-3 code: \{\texttt{LANGUAGE\_CODE}\}).

\textbf{Your task:}
\begin{itemize}
  \item If the text is in  \{\texttt{LANGUAGE\_NAME}\} or mixed with another language (e.g., English), keep and refine it.
  \item If the text is \emph{not} in  \{\texttt{LANGUAGE\_NAME}\} at all, remove it completely.
\end{itemize}

\textbf{For each input item:}
\begin{itemize}
  \item Rewrite the text for clarity, flow, and readability while preserving meaning.
  \item Fix grammar, spelling, and punctuation.
  \item Reorganize ideas logically to improve coherence.
  \item Replace repetitive or awkward wording with smoother alternatives, but do \textbf{not} shorten the overall text.
  \item Enrich the text with on-topic elaboration, nuance, or re-expression so the final output is equal to or longer than the input.
  \item Target length: $100$--$140$\% of original. \textbf{Absolute rule: never shorter than the input.}
  \item Default tone: clear, engaging, and respectful.
  \item Do not add unrelated facts or change intent.
  \item Do not alter or rewrite any direct quotations from religious scriptures (e.g., Bible, Qur’an, Hadith, Torah, Vedas, etc.). Preserve them exactly as written.
\end{itemize}

\textbf{Rules:}
\begin{itemize}
  \item Expansion is required: if you reduce wording in one place, expand in another place on the same topic.
  \item Keep all additions factually aligned with the input.
  \item Respect sensitive language; maintain a respectful tone.
  \item Remove toxic or unsafe text.
\end{itemize}
\end{AIbox}

\subsection{Prompt for Sentence Alignment in Extreme-Low-Resource Languages}
\label{app:text-alignment-prompt}
\begin{AIbox}{}
You are a bilingual text alignment assistant. 
Given two ordered lists of sentences (Set A = source, Set B = target):

\begin{enumerate}
  \item Align them monotonically, contiguous, non-overlapping.
  \item Fix formatting issues (merge broken lines, normalize spacing/punctuation, ensure each aligned pair is a clean sentence or short paragraph).
  \item Output ONLY valid JSON Lines (JSONL) format, one alignment per line:
\end{enumerate}

\medskip
\texttt{\{"source": ["<cleaned source text>"], "target": ["<cleaned target text>"]\}}

\medskip
\textbf{Rules:}
\begin{itemize}
  \item Prefer 1--1 sentence alignment; allow 1--N, M--1, or M--N if needed.
  \item Copy text exactly after cleaning, no translation.
  \item Skip pairs where source and target are in the same language or where text overlap is greater than $\sim$70\%. Do not output these at all.
  \item Return ONLY the JSONL output, no other text or explanations.
\end{itemize}
\end{AIbox}

\subsection{Prompt for LLM-Based Enhancement of English Text for Back-Translation}
\label{app:BT-english-text-filtering}

\begin{AIbox}{}
You are an expert English copy editor preparing sentences for machine translator training.

\textbf{For each input item:}
\begin{itemize}
  \item If the text is not natural-language English (e.g., another language, code, LaTeX, complex math) $\rightarrow$ skip.
  \item If the text is meaningless $\rightarrow$ skip.
  \item Fix spelling, grammar, agreement, casing, whitespace, and hyphenation. Keep named entities intact.
  \item You may freely rephrase to improve clarity and flow, staying close to the main theme and content.
  \item If text exceeds 30 words $\rightarrow$ rewrite and summarize within 30 words.
  \item Maintain sentence case and end with proper punctuation.
  \item Produce clean, easy-to-understand English sentences suitable for translation.
\end{itemize}
\end{AIbox}

\subsection{Prompt for LLM-Based Post-Editing of MT Outputs}
\label{app:llm-postedit-prompt}
\begin{AIbox}{}
You are a translation post-editor. The \texttt{\{source\_text}\} is in \texttt{\{source\_lang\}}, and the \texttt{\{translated\_text}\} is in \texttt{\{target\_lang\}}.

The translation may contain minor errors or wrong word choices. Your task is to correct only those errors with minimal edits. 

Do not change sentence structure, phrasing, or style unnecessarily - preserve the original translation as much as possible. 

Do not try to make factual corrections, only fix language issues.

Output only the corrected \texttt{\{target\_lang\}} sentence without any quotes, explanations, or additional formatting.
\end{AIbox}

\subsection{Prompt to Generate RTTBench-Mono}
\label{app:prompt-rttbench}
\begin{AIbox}{}
You are a domain-aware writing assistant generating English sentences that will later be used to benchmark machine-translation systems. Your task is to generate sentences that are diverse in style and complexity, natural, and highly specific to the \texttt{\{domain\}} domain. Each sentence must be self-contained and unambiguously belong to this domain.

\medskip
\textbf{Domain Context and Disambiguation}
\begin{itemize}
  \item \textbf{Target Domain:} \texttt{\{domain\}}
  \item \textbf{Domain Description:} \texttt{\{domain\_description\}}
  \item \textbf{Confusable Domains to Avoid:} \texttt{\{', '.join(confusable\_domains)\}}
\end{itemize}

\textbf{Instructions}
\begin{enumerate}
  \item Generate exactly \texttt{\{count\}} independent sentences that are clearly and specifically about \texttt{\{domain\}}.
  \item Each sentence must be unambiguously about \texttt{\{domain\}} and should feel out of place or less relevant in these related domains: \texttt{\{', '.join(confusable\_domains)\}}. This is the most important rule.
  \item Return the sentences as a plain numbered list (1., 2., etc.). No extra commentary.
  \item Every sentence must be unique, self-contained, safe, and obviously about \texttt{\{domain\}}.
  \item \textbf{Strictly enforce sentence-length quotas.} Your final output of \texttt{\{count\}} sentences \textbf{must} be composed of an exact number of sentences from each length band, as specified below:
    \begin{itemize}
      \item Short (\texttt{SENTENCE\_LENGTH\_BANDS[0]} words): exactly \texttt{\{quotas[0]\}} sentences.
      \item Medium (\texttt{SENTENCE\_LENGTH\_BANDS[1]} words): exactly \texttt{\{quotas[1]\}} sentences.
      \item Long (\texttt{SENTENCE\_LENGTH\_BANDS[2]} words): exactly \texttt{\{quotas[2]\}} sentences.
    \end{itemize}
    You must verify your own output to ensure this distribution is perfectly met.
  \item \textbf{Vary the tone, register, and complexity.} The set of \texttt{\{count\}} sentences should include everyday/accessible style, informal with some domain-specific jargon, formal/literary, and technical/professional language. Do \emph{not} follow a fixed pattern; the mix should look organic.
  \item Cover different linguistic dimensions across your sentences:
    \begin{itemize}
      \item sentence type: declarative, interrogative, imperative, exclamatory, factual, reasoning-based, comparative, causal, hypothetical, counterfactual, indirect speech;
      \item voice: active and passive;
      \item tense/aspect: past, present, future, perfect, conditional;
      \item terminology: mix common domain vocabulary with more specialized jargon or acronyms;
      \item entities and numerics: names, dates, currencies, units, measurements;
      \item figurative language where appropriate;
      \item occasional ambiguity (e.g., ‘bank’, ‘rock’);
      \item co-reference and pronouns (he, she, they, it) — but not in every sentence.
    \end{itemize}
  \item \textbf{Final formatting rules:}
    \begin{itemize}
      \item If a sentence needs quotation marks, use single quotes (').
      \item Do not add external commentary or restate these instructions in the output.
    \end{itemize}
\end{enumerate}
\end{AIbox}

\subsection{Prompt Template for {LLM-as-a-Judge} Evaluation}
\label{subsec:llm-as-judge-template}
\begin{AIbox}{}
\textbf{System Preamble}

You are an expert evaluator of language-model outputs in \texttt{\{LANGUAGE\_NAME}\}.
Your task is to compare two candidate answers to the same question in \texttt{\{LANGUAGE\_NAME}\} and select the better one.
\end{AIbox}

\begin{AIbox}{}
\textbf{Prompt Template}

Evaluation criteria:

\begin{tabular}{@{}V{0.7cm}V{13cm}@{}}
1) & Linguistic correctness: the answer must be primarily in [LANGUAGE\_NAME] and use appropriate vocabulary and style. \\
2) & Instruction following: the answer should fully address the question and follow all given instructions. \\
3) & Factual accuracy and semantic comprehension with respect to the context and reference answer. \\
4) & Grammar and fluency in \texttt{\{LANGUAGE\_NAME}\}. \\
\end{tabular}

\vspace{6pt}

\begin{tabular}{@{}V{3cm}V{10.8cm}@{}}
Context: & \texttt{\{CONTEXT}\} \\
Question: & \texttt{\{QUESTION}\} \\
Reference answer: & \texttt{\{ANSWER}\}\\
\end{tabular}

\vspace{6pt}

\begin{tabular}{@{}V{3cm}V{10.8cm}@{}}
Answer (A): &  \texttt{\{COMPLETION A}\} \\
Answer (B): &  \texttt{\{COMPLETION B}\}\\
\end{tabular}

\vspace{6pt}

Evaluate both answers according to the criteria above and provide:

\begin{tabular}{@{}V{0.7cm}V{13cm}@{}}
1) & A comparison in English explaining which answer is better and why. \\
2) & Your preference: Answer (A) or Answer (B). \\
3) & A rating for each answer on a 0--5 scale based on the evaluation criteria (0 = unusable, 5 = excellent). \\
\end{tabular}

\vspace{6pt}

Format your response exactly as follows (no additional text):\\[2pt]
\begingroup\ttfamily
Comparison: <your comparison>\\
Preferred: <"Answer (A)" or "Answer (B)">\\
Rating output A: [0-5]\\
Rating output B: [0-5]%
\endgroup

\end{AIbox}

\end{document}